\documentclass[Afour, times]{sagej}      
   
\usepackage[table]{xcolor}
\usepackage[colorlinks,bookmarksopen,bookmarksnumbered]{hyperref}
\usepackage{moreverb,url}
\usepackage{algorithm}
\usepackage{multicol}
\usepackage{multirow}
\usepackage{algpseudocode}
\usepackage{color}
\usepackage{graphicx}
\usepackage{bm}
\usepackage{amsmath}
\usepackage{amssymb}
\usepackage{tabularx}
\usepackage{subcaption}
\usepackage{apacite}
\usepackage{siunitx}

\usepackage{prettyref}
\newrefformat{fig}{Figure~\ref{#1}}
\newrefformat{sec}{Section~\ref{#1}}
\newrefformat{tab}{Table~\ref{#1}}
\newrefformat{alg}{Algorithm~\ref{#1}}
\newrefformat{eq}{Equation~\ref{#1}}

\newcommand{\minus}{\scalebox{0.75}[1.0]{$-$}}
\DeclareMathOperator*{\argmin}{\arg\!\min}

\newcommand\BibTeX{{\rmfamily B\kern-.05em \textsc{i\kern-.025em b}\kern-.08em
T\kern-.1667em\lower.7ex\hbox{E}\kern-.125emX}}

\def\volumeyear{2018}

\begin{document}

\title{Fully Distributed Multi-Robot Collision Avoidance via Deep Reinforcement Learning for Safe and Efficient Navigation in Complex Scenarios}

\author{Tingxiang Fan\affilnum{1*}, Pinxin Long\affilnum{2*}, Wenxi Liu\affilnum{3} and Jia Pan\affilnum{1}}

\affiliation{* These authors contributed equally. \\
\affilnum{1}Department of Mechanical and Biomedical Engineering, City University 
of Hong Kong, China \\
\affilnum{2}Metoak Technology (Beijing) CO., LTD, China \\
\affilnum{3}College of Mathematics and Computer Science, Fuzhou University, China}

\corrauth{Jia Pan, City University of Hong Kong}
\email{jiapan@cityu.edu.hk}

\begin{abstract}
Developing a safe and efficient collision avoidance policy for multiple robots is challenging in the decentralized scenarios where each robot generates its paths with the limited observation of other robots' states and intentions. Prior distributed multi-robot collision avoidance systems often require frequent inter-robot communication or agent-level features to plan a local collision-free action, which is not robust and computationally prohibitive. In addition, the performance of these methods is not comparable to their centralized counterparts in practice. 

In this paper, we present a decentralized sensor-level collision avoidance policy for multi-robot systems, which shows promising results in practical applications. In particular, our policy directly maps raw sensor measurements to an agent's steering commands in terms of the movement velocity. As a first step toward reducing the performance gap between decentralized and centralized methods, we present a multi-scenario multi-stage training framework to learn an optimal policy. The policy is trained over a large number of robots in rich, complex environments simultaneously using a policy gradient based reinforcement learning algorithm. The learning algorithm is also integrated into a hybrid control framework to further improve the policy's robustness and effectiveness.

We validate the learned sensor-level collision avoidance policy in a variety of simulated and real-world scenarios with thorough performance evaluations
for large-scale multi-robot systems. The generalization of the learned policy is verified in a set of unseen scenarios
including the navigation of a group of heterogeneous robots and a large-scale scenario with $100$ robots. Although the policy is trained using simulation data only, we have successfully deployed it on physical robots with shapes and dynamics characteristics that are different from the simulated agents, in order to demonstrate the controller's robustness against the sim-to-real modeling error.
Finally, we show that the collision-avoidance policy learned from multi-robot navigation tasks provides an excellent solution to the safe and effective autonomous navigation for a single robot working in a dense real human crowd. Our learned policy enables a robot to make effective progress in a crowd without getting stuck. More importantly, the policy has been successfully deployed on different types of physical robot platforms without tedious parameter tuning.
Videos are available at~\url{https://sites.google.com/view/hybridmrca}.
\end{abstract}

\maketitle

\section{Introduction}
\label{sec:intro}
Multi-robot navigation has recently gained much interest in robotics and artificial intelligence and has many applications including multi-robot search and rescue, navigation through human crowds, and autonomous warehouse. One of the major challenges for multi-robot navigation is to develop a safe and robust collision avoidance policy such that each robot can navigate from its starting position to its desired goal safely and efficiently. More importantly, a sophisticated collision-avoidance skill is also a prerequisite for robots to accomplish many complex tasks, including multi-robot collaboration, tracking, and navigation through a dense crowd.

Some of prior works, known as \textit{centralized methods}, such as~\cite{schwartz1983piano,yu2016optimal,tang2018hold}, assume that the actions of all agents are determined by a central server aware of comprehensive knowledge about all agents' intents (e.g., initial states and goals) and their workspace (e.g., a 2D grid map). These centralized approaches generate collision avoidance actions by planning optimal paths for all robots simultaneously. They can guarantee safety, completeness, and approximate optimality but are difficult to scale to large systems with many robots due to several main limitations. First, the centralized control and scheduling become computationally prohibitive as the number of robots increases.
Second, these methods require a reliable synchronized communication between the central server and all robots, which is either uneconomical or not feasible for large-scale systems. Third, the centralized system is vulnerable to failures or disturbances of the central server, communication between robots, or motors and sensors mounted on any individual robot. Furthermore, these centralized methods are inapplicable when multiple robots are deployed in an unknown and unstructured environment, e.g., in a warehouse with human co-workers.

\begin{figure*}[t] 
\centering
\includegraphics[width=6.3in]{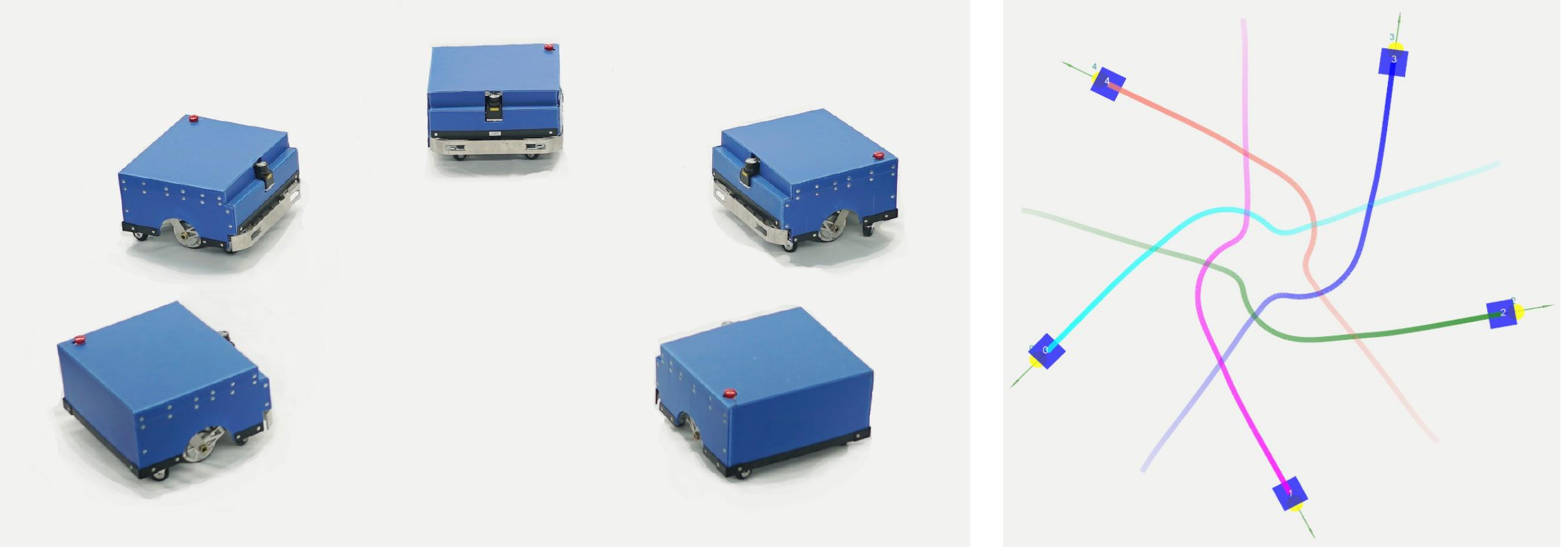}\\
\caption{Direct deployment of our policy trained in simulation to physical robots without parameter fine tuning. The physical robots (left) are square-shaped and twice the size of the disc-shaped robots for which we trained the policy in the simulation. The physical robot trajectories (right) in the \emph{Circle} scenario generated from our learned navigation policy are smooth and safe. Here we use different colors to distinguish trajectories for different robots and use the color transparency to indicate the timing along a trajectory sequence. 
This experiment verifies the excellent generalization of our learned policy. }
\label{fig:first}
\end{figure*}

Compared with centralized methods, some existing works propose \textit{agent-level decentralized collision avoidance} policies, where each agent independently makes its decision by taking into account the observable states (e.g., shapes, velocities, and positions) of other agents.
Most agent-level policies are based on the velocity obstacle (VO) framework~\cite{ Berg:ORCA:2011, snape2011hybrid, hennes2012multi, claes2012collision, bareiss2015generalized}, which infers local collision-free motion efficiently for multiple agents in cluttered workspaces. However, these methods have several severe limitations which prevent them from being widely used in real scenarios. 
First, the simulation-based works~\cite{van2008reciprocal,Berg:ORCA:2011} assume that each agent has perfect sensing about the surrounding environment, but this assumption does not hold in real-world scenarios due to omnipresent sensing uncertainty. 
To moderate the limitation of perfect sensing, some previous approaches use a global positioning system to track the positions and velocities of all robots~\cite{snape2011hybrid, bareiss2015generalized}. However, such external infrastructure is too expensive to scale to large multi-robot systems.
Some other methods design inter-agent communication protocols for sharing position and velocity information among nearby agents~\cite{claes2012collision, hennes2012multi, godoy2016implicit}. However, the communication systems introduce additional difficulties such as  delay or blocking of communication signals due to obstacle occlusions.
The second limitation is that the VO-based policies are controlled by multiple tunable parameters that are sensitive to scenario settings and thus must be carefully set to accomplish satisfactory multi-robot motion. 
Finally, the timing performance of previous decentralized methods for navigation is significantly inferior to their centralized counterparts in practical applications.

Inspired by VO-based approaches, ~\cite{chen2017decentralized} trained an agent-level collision avoidance policy using deep reinforcement learning (DRL), which learns a two-agent value function that explicitly maps an agent's own state and its neighbors' states to a collision-free action, whereas it still demands the perfect sensing. In their later work~\cite{chen2017socially}, multiple sensors are deployed to perform tasks of segmentation, recognition, and tracking in order to estimate the states of nearby agents and moving obstacles. However, this complex pipeline not only requires expensive online computation but also makes the whole system less robust to the perception uncertainty.

In this paper, we focus on 
\textit{sensor-level decentralized collision avoidance} policies that directly map the raw sensor data to desired collision-free steering commands. Compared with agent-level policies, our approach requires neither perfect sensing for neighboring agents and obstacles nor tedious offline parameter-tuning for adapting to different scenarios. We have successfully deployed the learned policy to a large number of heterogeneous robots to achieve high-quality, large-scale multi-robot collaboration in both simulation and physical settings. Our method also endows a robot with high mobility to navigate safely and efficiently through a dense crowd with moving pedestrians.

Sensor-level collision avoidance policies are often modeled using deep neural networks (DNNs) and trained using supervised learning on a large dataset~\cite{long2017deep,pfeiffer2017perception}. 
However, there are several limitations to learning policies under supervision.
First, it demands a large amount of training data that should cover different interaction situations for multiple robots. Second, the expert trajectories in datasets are not guaranteed to be optimal in interaction scenarios, which makes training difficult to converge to a robust solution. Third, it is difficult to hand-design a proper loss function for training robust collision avoidance policies. 
To overcome the above drawbacks of supervised learning, we propose a multi-scenario multi-stage deep reinforcement learning framework to learn the optimal collision avoidance policy using the policy gradient method. The policy is trained in simulation for a large-scale multi-robot system in a set of complex scenarios. The learned policy outperforms the existing agent- and sensor-level approaches in term of navigation speed and safety. It can be deployed to physical robots directly and smoothly without tedious parameter tuning.

Our collision avoidance policy is trained in simulation rather in the real world because it is difficult to generate training data safely and effectively for physical robots while the performance of robot learning heavily depends on the quality and quantity of the collected training data. In particular, to learn a sophisticated collision avoidance policy, robots need to undergo thousands of millions of collisions with the static environments and the moving pedestrians in order to fully explore different reaction and interaction behaviors. Such data collection and policy update procedure would be expensive, time-consuming and dangerous if being implemented in the real world. As a result, we follow the recent trend of sim-to-real methods~\cite{rusu2016sim,sadeghi2016cad2rl,tobin2017domain,peng2017sim}, by first training a control policy in virtual simulation and then deploying the learned model to the real robot directly. However, it is non-trivial to transfer learned policies from simulation to the real world and may suffer a significant loss in performance, because the real world is messy and contains an infinite number of novel scenarios that may not be encountered during training. Fortunately, our policy uses an input space that is robust to the sim-to-real gap, and therefore we are able to use simulation to greatly accelerate our robot learning process. 

Another challenge for learning-based collision avoidance is the lack of guarantee for the safety and completeness of the learned policy, which is considered as one of the main concerns when using deep learning in robotics. Inspired by the hybrid control framework~\cite{egerstedt2002hybrid}, we alleviate this difficulty by combining the learning-based policy with traditional control. In particular, the traditional control takes charge of relatively simple scenarios or emergent situations, while the learned policies will handle more complex scenarios by generating more sophisticated behaviors bounded in a limited action space. In this way, we can further improve the performance of the collision avoidance policy in terms of safety without sacrificing efficiency.

\noindent \textbf{Contributions:} Our main contributions in this paper are:
\begin{itemize}
	\item We develop a fully decentralized multi-robot collision avoidance framework, where each robot makes navigation decisions independently without any communication with others. The navigation policy has the ranging data collected from a robot's on-board sensors as the input and outputs a control velocity. The policy is optimized offline via a novel multi-scenario multi-stage reinforcement learning algorithm and trained with a robust policy gradient method. The learned policy can be executed online on each robot using a relatively cheap on-board computer, e.g., an NUC or a Movidius neural compute stick. 
	\item We further combine the deep-learned policy with traditional control approaches to achieve a robust hybrid navigation policy. The hybrid policy is able to find time-efficient and collision-free paths for a large-scale nonholonomic robot system and it can be safely generalized to unseen simulated and real-world scenarios. Its performance is also much better than previous decentralized methods, and can serve as a first step toward reducing the gap between centralized and decentralized navigation policies.
	\item Our decentralized policy is trained in simulation but does not suffer from the sim-to-real gap and can be directly deployed to physical robots without tedious fine-tuning. 
	\item We deploy the policy in a warehouse scenario to control multiple robots in a decentralized manner. Compared to previous solutions to the multi-robot warehouse, our system requires neither a pre-constructured bar-code infrastructure nor a map. It also alleviates the demands for high-band and low-latency synchronized communication that is considered as a main bottleneck when scaling a multi-robot system. In particular, we use an Ultra-Wide-Band (UWB) system to provide a rough guidance to the robots about their goals, and all robots navigate toward their goals independently by using our hybrid policy to resolve the collisions during the navigation. Hence, our system can be easily extend to hundreds or more robots.
	\item We deploy the policy directly on a physical robot to achieve smooth and effective navigation through a dense crowd with moving pedestrians. The system's excellent mobility outperforms state-of-the-art solutions. Moreover, the policy can be easily deployed to different robotic platforms with different shapes and dynamics, and still can provide satisfactory navigation performance.
\end{itemize}

The rest of this paper is organized as follows. \prettyref{sec:related} provides a brief review about related works. In \prettyref{sec:prob}, we introduce the notations and formulate our problem. In \prettyref{sec:approach}, we discuss the details about how to use reinforcement learning to train a high-quality multi-robot collision avoidance policy, and how to combine the learned policy with traditional optimal control schemes to achieve a more robust hybrid control policy. Finally, we highlight our experimental results in both the simulation scenarios (\prettyref{sec:sim_exp}) and the real-world scenarios (\prettyref{sec:real_exp}).

This paper is an extension of our previous conference paper~\cite{long2017towards}.

\begin{figure}
\centering
\includegraphics[width=1\linewidth]{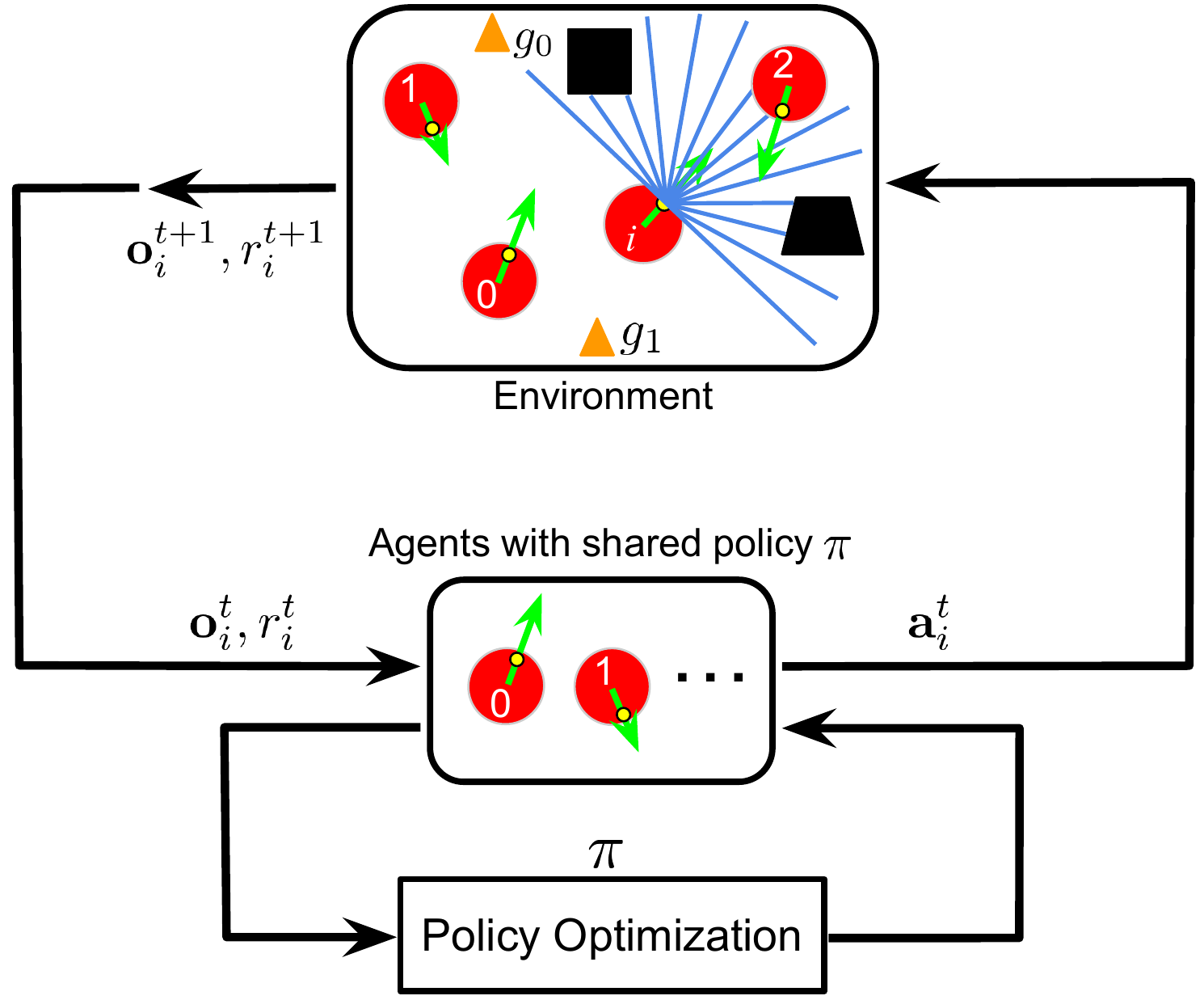} 
\caption{An overview of our approach. At each time-step $t$, each robot receives its observation $\mathbf{o}^t_i$ and reward $r^t_i$ from the environment, and generates an action $a_i^t$ following the policy $\pi$. The policy $\pi$ is shared across all robots and updated by a policy gradient based reinforcement learning algorithm.}
\label{fig:overview}
\end{figure}

\begin{figure*}[t]
\centering
\includegraphics[width=0.8\textwidth]{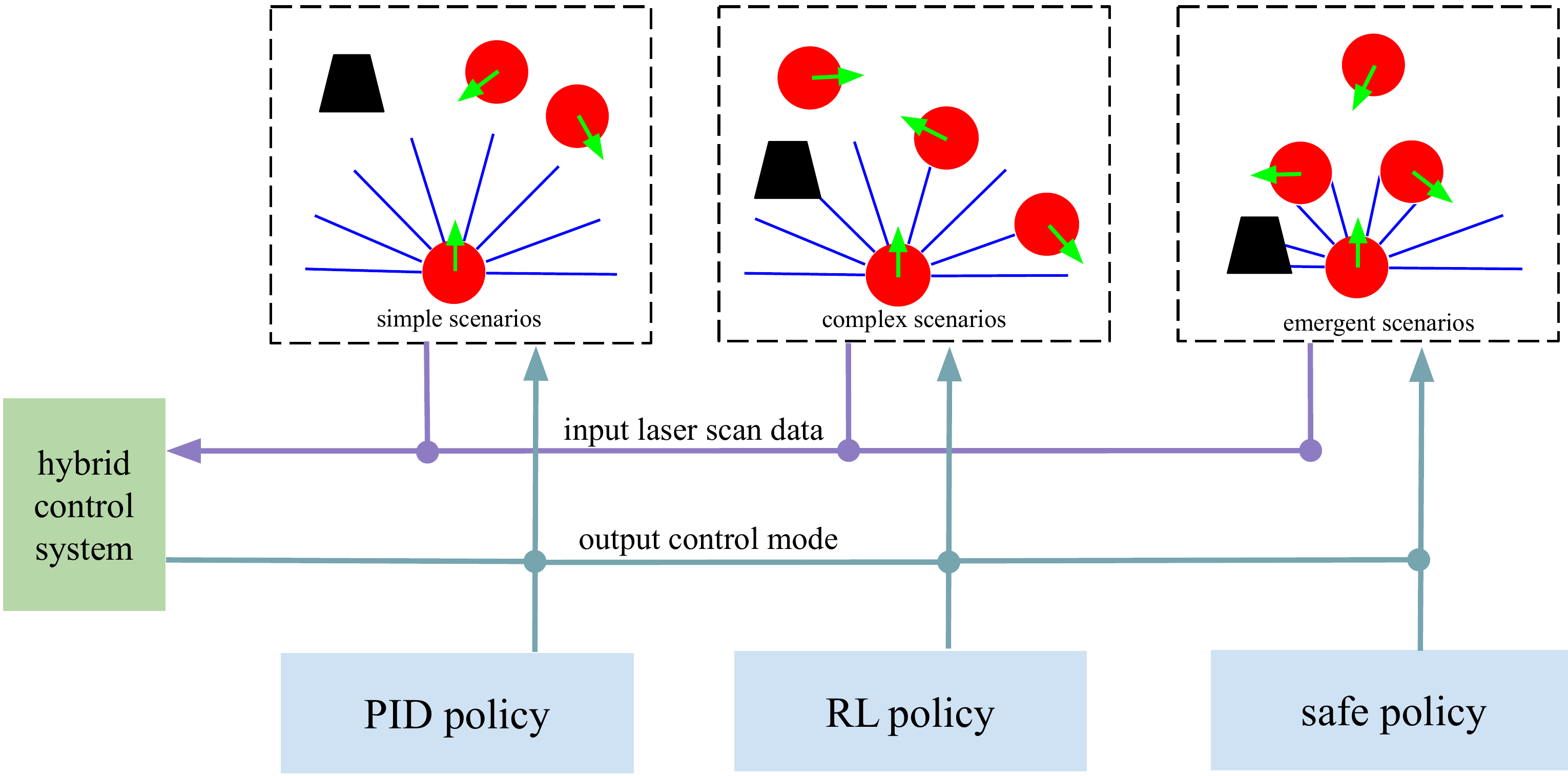} 
\caption{An overview of the hybrid control architecture. According to the sensor measurement, a robot will classify its surrounding environment into three categories: the simple scenario, the complex scenario, and the emergent scenario. The robot will choose different controllers for different scenarios, in particular, the PID policy for simple scenarios, the reinforcement learned policy for complex scenarios, and the safe policy for emergent scenarios. In this way, the robot can make a good balance between finishing its tasks efficiently and avoiding obstacles safely. For more details, please refer to \prettyref{sec:hybrid}.
}
\label{fig:hybrid_ctrl}
\end{figure*}
\section{Related works}
\label{sec:related}

In this section, we first briefly survey the related works on multi-robot navigation, including the centralized and decentralized approaches. Next, we summarize recent literatures about how to solve the multi-robot navigation problem in a data-driven manner with machine learning techniques. We further discuss the sim-to-real gap of the learning-based navigation control, and then summarize the recent progress in resolving this issue, including a short introduction to the hybrid control. Finally, we briefly survey the deep reinforcement learning and its application in reactive control. 

\subsection{Centralized multi-robot navigation}
Over the past decade, the centralized multi-robot motion planning methods have been widely applied in many applications, such as task allocation~\cite{stone1999task,chen2011resource}, formation control~\cite{balch1998behavior,de2008dynamic,sun2009synchronization,chen2010leader}, and object transportation~\cite{michael2011cooperative,hu2011automatic,alonso2017multi,turpin2014capt}. 
The centralized methods assume that each robot can have access to the complete information (e.g., velocity and desired goal) of other robots via a global communication system for motion planning. In this way, they can guarantee a safe, optimal and complete solution for navigation~\cite{luna2011efficient,sharon2015conflict,yu2016optimal,tang2018hold}. However, these centralized approaches are difficult to be deployed in large-scale real-world systems due to several reasons. 
First, the centralized control and scheduling become computationally prohibitive as the number of robots increases.
Second, these methods require a reliable synchronized communication between the central server and all robots, which is either uneconomical or not feasible for large-scale systems. Third, the centralized system is vulnerable to failures or disturbances of the central server, the communication between robots, or motors and sensors of any individual robot. Furthermore, these centralized methods are inapplicable when multiple robots are deployed in an unknown and unstructured environment, e.g., in a warehouse with human co-workers. Due to these limitations, in this paper, we present a decentralized method to accomplish large-scale multi-robot navigation.

\subsection{Decentralized multi-robot navigation}
Compared with the centralized methods, the decentralized methods are extensively studied mainly for the collision avoidance task in multi-agent systems.
As one representative approach, the Optimal Reciprocal Collision Avoidance (ORCA) framework~\cite{Berg:ORCA:2011} has been widely used in crowd simulation and multi-agent systems. In particular, ORCA provides a sufficient condition for multiple agents to avoid collisions with others in a short time horizon, and can easily be scaled to large systems with many agents. ORCA and its variants~\cite{snape2011hybrid,bareiss2015generalized} used heuristics to construct a parametric collision avoidance model, which are tedious to be tuned for satisfactory performance. Besides, these methods are difficult to adapt to real-world scenarios with ubiquitous uncertainties, because they assume that each robot has perfect sensing about the surrounding agents' positions, velocities, and shapes. To alleviate the demand for perfect sensing, communication protocols have been introduced by~\cite{hennes2012multi,claes2012collision,godoy2016implicit} to share agents' state information, including positions and velocities among the group. However, introducing the communication network hurts the robustness and flexibility of multi-robot system. To alleviate this problem,~\cite{zhou2017fast} planned the robots' paths by computing each robot's buffered Voronoi cell. Unlike previous work that needs both the position and velocity knowledge about adjacent robots, this method only assumes each robot knowing the position of surrounding agents, though it still uses a motion capture system as the global localization infrastructure.

In addition, the original formulation of ORCA is based on holonomic robots, while the robots  in the real-world scenarios are often non-holonomic.
In order to deploy ORCA on the most common differential drive robots, several methods have been proposed to deal with kinematics of non-holonomic robots. ORCA-DD~\cite{snape2010smooth} doubles the effective radius of each agent to ensure collision-free and smooth paths for robots under differential constraints. But it causes problems for the navigation in narrow passages and unstructured environments. 
NH-ORCA~\cite{alonso2013optimal} enabled a differential drive robot to follow a desired speed command within an error of $\varepsilon$. It only needs to slightly increase the effective radius of a robot according to the value of $\varepsilon$ and thus outperforms ORCA-DD in collision avoidance performance. 

\subsection{Learning-based collision avoidance policy}
The decentralized polices discussed above are based on some first-principle rules for collision avoidance in general situations, and thus cannot achieve the optimal performance for a specific task or scenario, such as navigation in the crowded pedestrian scenario. To solve this problem, learning-based collision avoidance techniques try to optimize a parameterized policy using the data collected from a specific task. Most recent work focus on the task of single robot navigation in environments with static obstacles. 
Many approaches adopted the supervised learning paradigm to train a collision avoidance policy by mapping sensor input to motion commands.~\cite{muller2006off} presented a vision-based static obstacle avoidance system. They trained a $6$-layer convolutional network which maps raw input images to steering angles.~\cite{zhang2016deep} exploited a deep reinforcement learning algorithm based on the successor features~\cite{Barreto:2017:SFT} to transfer the depth information learned in previously mastered navigation tasks to new problem instances. ~\cite{sergeant2015multimodal} proposed a mobile robot control system based on multimodal deep autoencoders.~\cite{ross2013learning} trained a discrete controller for a small quadrotor helicopter with imitation learning techniques to accomplish the task of collision avoidance using a single monocular camera. In their formulation, only discrete movements (left/right) need to be learned.
Note that all of the aforementioned approaches only take into account the static obstacles and require manually collecting training data in a wide variety of environments. 

To deal with unstructured environments, one data-driven motion planner is presented by~\cite{pfeiffer2017perception}. They trained a model that maps laser range findings and target positions to motion commands using expert demonstrations generated by the ROS navigation package. This model can navigate a robot through an unseen environment and successfully react to sudden changes. Nonetheless, similar to other supervised learning methods, the performance of the learned policy is severely constrained by the quality of the training sets. 
To overcome this limitation,~\cite{tai2017virtual} proposed a map-less motion planner trained through a deep reinforcement learning method.~\cite{kahn2017uncertainty} presented an uncertainty-aware model-based reinforcement learning algorithm to estimate the probability of collision in an unknown environment. However, these test environments are relatively simple and structured, and the learned planner is difficult to generalize to complex scenarios with dynamic obstacles and other proactive agents. 

To address highly dynamic unstructured environments, some decentralized multi-robot navigation approaches are proposed recently. ~\cite{godoy2016moving} introduced Bayesian inference approach to predict surrounding dynamic obstacles and to compute collision-free command through the ORCA framework~\cite{van2011reciprocal}. ~\cite{chen2017decentralized,chen2017socially,everett2018motion} proposed multi-robot collision avoidance policies by deep reinforcement learning, which requires deploying multiple sensors to estimate the states of nearby agents and moving obstacles. However, their complicated pipeline not only demands expensive online computation but also makes the whole system less robust to the perception uncertainty.

\subsection{Sim-to-real gap and hybrid control}
In order to learn a sophisticated control policy with reinforcement learning, robots need to interact with the environment for a long period to accumulate knowledge about the consequences of different actions. Collecting such interaction data in real world is expensive, time-consuming, and sometimes infeasible due to safety issues. As a result, most reinforcement learning studies are conducted in simulation. However, a policy learned in simulation may be problematic when directly adapting to real world applications, known as the sim-to-real gap. Many recent researches attempt to address this problem. ~\cite{rusu2016sim} proposed a progressive nets architecture that assembles a set of neural networks trained for different tasks to build a new network that is able to bridge the simulation and the real world. ~\cite{sadeghi2016cad2rl} leveraged diversified rendering parameters in simulation to accomplish collision avoidance in real world without the input of any real images. ~\cite{tobin2017domain} introduced domain randomization to train a robust and transferable policy for manipulation tasks.

In this paper, we utilize the hybrid control framework to alleviate the sim-to-real problem. Hybrid control architecture has been widely used for robot navigation problems in the past two decades~\cite{egerstedt2002hybrid,adouane2009hybrid}, which designs a high-level control strategy to manage a set of low-level control rules. Hybrid control has also been heavily leveraged in multi-robot systems~\cite{alur1999formal,quottrup2004multi}. For distributed control, ~\cite{shucker2007switching} proposed switching rules to deal with challenging cases where collisions or noises cannot be handled appropriately. ~\cite{gillula2011applications} introduced a combination of hybrid decomposition and reachability analysis to design a decentralized collision avoidance algorithm for multiple aerial vehicles. Different from prior methods, we use hybrid control to improve the transferability and robustness of the learned control policies. In particular, we use traditional control policies to deal with situations with large sim-to-real gap (i.e., the situations cannot be handled well by learned policy alone), and use a high level heuristic to switch between the learned policy and traditional control policies.

\begin{figure*}
\centering 
\includegraphics[width=0.95\linewidth]{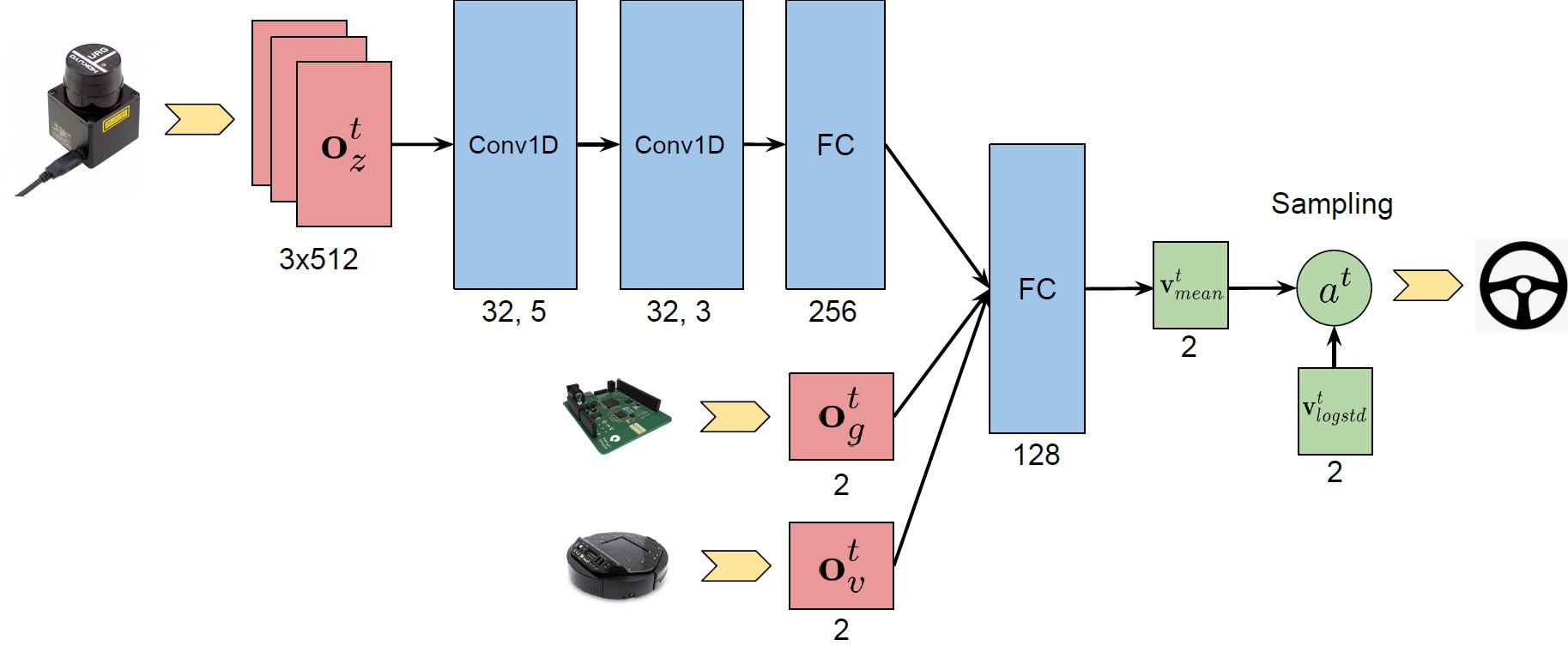} 
\caption{The architecture of the policy network. The input of the network includes the scan measurements $\mathbf{o}^t_z$, relative goal position $\mathbf{o}^t_g$ and current velocity $\mathbf{o}^t_v$. It outputs the mean of velocity $\mathbf{v}^t\textsubscript{mean}$. The final action $\mathbf a^t$ is sampled from a Gaussian distribution with a mean $\mathbf{v}^t\textsubscript{mean}$ and a separated log standard deviation vector $\mathbf{v}^t\textsubscript{logstd}$.} 
\label{fig:model} 
\end{figure*}

\subsection{Deep reinforcement learning and reactive control}
Deep learning methods have been successfully used in a wide variety of applications, including computer vision~\cite{krizhevsky2012imagenet,he2016deep} and nature language processing~\cite{graves2013speech,amodei2016deep}. 
These successes indicate that deep neural networks are good at extracting hierarchal features from complex and high-dimensional data and making high-quality decisions. These advantages make deep neural networks a powerful tool for robotic applications, which also need to deal with complex raw input data captured from onboard sensors. 
Recently, there is an increasing amount of research that appeals for deep neural networks to solve robotic perception and control problems.~\cite{muller2006off} developed a vision-based obstacle avoidance system for a mobile robot by training a $6$-layer convolutional network that maps raw input images to steering angles.~\cite{levine2016end} presented an end-to-end framework which learns control policies mapping raw image observations to torques at the robot’s motors. Deep neural networks have also been integrated with model predictive control to control robotic systems~\cite{lenzdeepMPC2015,zhang:2016:LDC}.~\cite{OndruskaAAAI2016,OndruskaRSS2016} combined recurrent neural networks with convolutional operations to learn a direct mapping from raw 2D laser data to the unoccluded state of the entire scene; the mapping is then used for object tracking. In this paper, we trained a multi-layer neural network to produce the navigation velocity for each robot according to observations.

\section{Problem Formulation}
\label{sec:prob}
The multi-robot collision avoidance problem is formulated primarily in the context of a nonholonomic differential drive robot moving on the Euclidean plane and avoiding collision with obstacles and other fellow-robots. 

To tackle this problem, we first assume all robots in the environment are homogeneous.
Specifically, all of $N$ robots are modeled as discs with the same radius $R$, and the multi-robot collision avoidance is formulated as a partially observable decision making problem.
At each timestep $t$, the $i$-th robot $(1 \leq i \leq N)$ has access to an observation $\mathbf o_{i}^t$ which only provides partial knowledge about the world and then computes a collision-free steering command $\mathbf a_{i}^t$ that drives itself toward the goal $\mathbf g_{i}$ from the current position $\mathbf p_{i}^t$.

In our formulation, the observation $\mathbf o_{i}^t$ is drawn from a probability distribution w.r.t. the latent system state $\mathbf s_{i}^t$, i.e., $\mathbf o_{i}^t \sim \mathcal{O}(\mathbf s_{i}^t)$ and it only provides partial information about the state $\mathbf s_{i}^t$. 
In particular, the $i$-th robot cannot access other robots' states and intents, which is in accord with the real world situation. 
Compared with the perfect sensing assumption applied in prior methods, e.g.~\cite{chen2017decentralized,chen2017socially,claes2012collision,hennes2012multi,van2008reciprocal,Berg:ORCA:2011}, our partial observation assumption makes our proposed approach more applicable and robust in real world applications.
The observation vector of each robot is divided into three parts: $\mathbf o^t = [\mathbf{o}_{z}^t, \mathbf{o}_{g}^t, \mathbf{o}_{v}^t]$ (here we ignore the robot ID $i$ for legibility), where $\mathbf{o}_{z}^t$ denotes the raw 2D laser measurements about its surrounding environment, $\mathbf{o}_{g}^t$ stands for its relative goal position (i.e. the coordinate of the goal in the robot's local polar coordinate frame), and $\mathbf{o}_{v}^t$ refers to its current velocity. 
Given the partial observation $\mathbf o^t$, each robot \textit{independently} computes an action or a steering command, $\mathbf{a}^t$, sampled from a stochastic policy $\pi$ shared by all robots: 
\begin{equation}
\label{eq:stpolicy}
  \mathbf{a}^t \sim \pi_{\theta}(\mathbf{a}^t \mid \mathbf{o}^t),
\end{equation}
where ${\theta}$ denotes the policy parameters. 
The computed action $\mathbf{a}^t$ is a vector representing the velocity $\mathbf{v}^t$ driving the robot to approach its goal while avoiding collisions with other robots and obstacles $\mathbf{B}_k$ $(0 \leq k \leq M)$ within the time horizon $\Delta t$ until the next observation $\mathbf o^{t+1}$ is received. Hence, each robot sequentially makes decision until it reaches the goal.
Given the sequential decisions consisting of observations and actions (velocities) $(\mathbf{o}_i^t, \mathbf{v}_i^t)_{t=0:t^g_i}$, the trajectory of the $i$-th robot, $l_{i}$, can be computed, starting from the position $\mathbf{p}_i^{t=0}$ to its desired goal $\mathbf{p}_i^{t=t_i^g} \equiv \mathbf{g}_i$, where $t_i^g$ is the traveled time. 

To wrap up the above formulation, we define $\mathbb{L}$ as the set of trajectories for all robots, which are subject to the robot's kinematic (e.g., non-holonomic) constraints, i.e.:  
\begin{equation}
\begin{aligned}
\mathbb{L} ={} & \{ l_i, i = 1,..., N \mid \\
        & \mathbf{v}_i^t \sim \pi_{\theta}(\mathbf{a}_i^t \mid \mathbf{o}_i^t), \\
        & \mathbf{p}_i^t = \mathbf{p}_i^{t-1} + \Delta t \cdot \mathbf{v}_i^{t},  \\
        & \forall j \in [1, N], j\neq i: \| \mathbf{p}_i^t - \mathbf{p}_j^t \| > 2R \\
        & \land \forall k \in [1, M]: \| \mathbf{p}_i^t - \mathbf{B}_k \| > R   \\
        & \land \| \mathbf{v}_i^t \| \leq v_i^{\mathrm{max}} \}.
\end{aligned}
\end{equation}

To find an optimal policy shared by all robots, we adopt an objective by minimizing the expectation of the mean arrival time of all robots in the same scenario, which is defined as:
\begin{align}
\label{eq:obj}
\argmin_{\pi_{\theta}}~~\mathbb{E}[\frac{1}{N}\sum_{i=1}^N t_i^g|{\pi}_{\theta}],
\end{align}
where $t_i^g$ is the traveled time of the trajectory $l_i$ in $\mathbb{L}$ controlled by the shared policy $\pi_{\theta}$.

The average arrival time will also be used as an important metric to evaluate the learned policy in \prettyref{sec:sim_exp}.
We solve this optimization problem through a policy gradient based reinforcement learning method, which bounds the policy parameter updates to a trust region to ensure stability.

\section{Approach}
\label{sec:approach}
We begin this section by introducing the key ingredients of our reinforcement learning framework. Next, we describe the details about the architecture of the collision avoidance policy based on a deep neural network. Then we discuss the training protocols used to optimize the policy and to bridge the sim-to-real gap. Finally, we introduce the hybrid control framework (as shown in \prettyref{fig:hybrid_ctrl}), which combines the learning-based policy with traditional rule-based control methods to improve the controller's safety, effectiveness, and transferability.

\subsection{Reinforcement learning setup}
\label{sec:setup}
The partially observable sequential decision making problem defined in \prettyref{sec:prob} can be formulated as a Partially Observable Markov Decision Process (POMDP) solved with reinforcement learning. Formally, a POMDP can be described as a $6$-tuple $(\mathcal{S}, \mathcal{A}, \mathcal{P}, \mathcal{R}, \Omega, \mathcal{O})$, where $\mathcal{S}$ is the state space, $\mathcal{A}$ is the action space, $\mathcal{P}$ is the state-transition model, $\mathcal{R}$ is the reward function, $\Omega$ is the observation space ($\mathbf o \in \Omega$), and $\mathcal{O}$ is the observation probability distribution given the system state ($\mathbf o \sim \mathcal{O}(\mathbf s)$). In our formulation, each robot only has access to the observation sampled from the underlying system states. Furthermore, since each robot plans its motions in a fully decentralized manner, a multi-robot state-transition model $\mathcal{P}$ determined by the robots' kinematics and dynamics is not needed. Below we describe the details of the observation space, the action space, and the reward function. 

\subsubsection{\textbf{Observation space}} 
\label{sec:obs}
As mentioned in \prettyref{sec:prob}, the observation $\mathbf{o}^t$ consists of three parts: the readings of the 2D laser range finder $\mathbf{o}_z^t$, the relative goal position $\mathbf{o}_g^t$, and the robot's current velocity $\mathbf{o}_v^t$. Specifically, $\mathbf{o}_z^t$ includes the measurements of the last three consecutive frames from a 180-degree laser scanner which provides 512 distance values per scanning (i.e., $\mathbf{o}_z^t \in \mathbb{R}^{3 \times 512}$) with a maximum range of 4 meters. In practice, the scanner is mounted on the forepart of the robot instead of in the center (as shown in the left image in \prettyref{fig:first}) to obtain a large un-occluded view. The relative goal position $\mathbf{o}_g^t$ is a 2D vector representing the goal in polar coordinate (distance and angle) with respect to the robot's current position. The observed velocity $\mathbf{o}_v^t$ includes the current translational and rotational velocity of the differential-driven robot. The observations are normalized by subtracting the mean and dividing by the standard deviation using the statistics aggregated over the course of the entire training.  
\subsubsection{\textbf{Action space}} 
The action space is a set of permissible velocities in continuous space. The action of differential robot includes the translational and rotational velocity, i.e., $\mathbf{a}^{t} = [v^{t}, w^{t}]$. In this work, considering the real robot's kinematics and the real world applications, we set the range of the translational velocity $v \in (0.0, 1.0)$ and the rotational velocity in $w \in (-1.0, 1.0)$. Note that backward moving (i.e., $v < 0.0$) is not allowed since the laser range finder can not cover the back region of the robot. 
\subsubsection{\textbf{Reward design}} 
Our objective is to avoid collisions during the navigation and to minimize the mean arrival time of all robots. A reward function is designed to guide a team of robots to achieve this objective: 
\begin{equation}
\label{eq:reward}
r_i^t = (^gr)_i^t + (^cr)_i^t + (^wr)_i^t. 
\end{equation}
The reward $r$ received by robot $i$ at time-step $t$ is a sum of three terms, $^gr$, $^cr$, and $^wr$. In particular, the robot is awarded by $(^gr)_i^t$ for reaching its goal:
\begin{equation} 
(^gr)_i^t = 
\begin{cases}
  r\textsubscript{arrival}  & \text{if } \| \mathbf{p}_i^t - \mathbf{g}_i \| < 0.1 \\
  \omega_g(\|\mathbf{p}_i^{t-1}- \mathbf{g}_i \| - \|\mathbf{p}_i^{t}- \mathbf{g}_i \|) & \text{otherwise}. \\
\end{cases}
\end{equation}
When the robot collides with other robots or obstacles in the environment, it is penalized by $(^cr)_i^t$: 
\begin{equation}
(^cr)_i^t =  
\begin{cases}
r\textsubscript{collision} & \quad \text{if } \| \mathbf{p}_i^t - \mathbf{p}_j^t \| < 2R \\ 
              & \quad \text{or } \|\mathbf{p}_i^t - \mathbf{B}_k\| < R \\
0 & \quad \text{otherwise}.  \\
\end{cases}
\end{equation}
To encourage the robot to move smoothly, a small penalty $(^wr)_i^t$ is introduced to punish the large rotational velocities: 
\begin{equation}  
(^wr)_i^t = \omega_{w}|w_i^t|  \quad \quad \text{if }  |w_i^t| > 0.7 . 
\end{equation}
We set $r\textsubscript{arrival}=15$, $\omega_g=2.5$, $r\textsubscript{collision}=\minus 15$ and $\omega_w=\minus 0.1$ in the training procedure.

\subsection{Network architecture}
\label{sec:model}

Given the input (observation $\mathbf{o}_i^t$) and the output (action $\mathbf{v}_i^t$), now we elaborate the policy network mapping $\mathbf{o}_i^t$ to $\mathbf{v}_i^t$. 

We design a $4$-hidden-layer neural network as a non-linear function approximation to the policy $\pi_{\theta}$. Its architecture is shown in \prettyref{fig:model}. 
We employ the first three hidden layers to process the laser measurements $\mathbf{o}_z^t$ effectively. The first hidden layer convolves $32$ one-dimensional filters with kernel size $=5$, stride $=2$ over the three input scans and applies ReLU nonlinearities~\cite{nair2010rectified}. The second hidden layer convolves $32$ one-dimensional filters with kernel size $=3$, stride $=2$, again followed by ReLU nonlinearities. The third hidden layer is a fully-connected layer with $256$ rectifier units. The output of the third layer is concatenated with the other two inputs ($\mathbf{o}_g^t$ and $\mathbf{o}_v^t$), and then are fed into the last hidden layer, a fully-connected layer with $128$ rectifier units. The output layer is a fully-connected layer with two different activations: a sigmoid function, which is used to constrain the mean of translational velocity $v^t$ within $(0.0, 1.0)$, and a hyperbolic tangent function ($\tanh$), which constrains the mean of rotational velocity $w^t$ within $(\minus 1.0, 1.0)$.   

As a summary, the neural network maps the input observation vector $\mathbf{o}^t$ to a vector $\mathbf{v}^t\textsubscript{mean}$. 
The final action $\mathbf{a}^t$ is sampled from a Gaussian distribution $\mathcal{N}(\mathbf{v}^t\textsubscript{mean}, \mathbf{v}\textsubscript{logstd}^t)$ that is used to model the stochastic policy formulated in \prettyref{eq:stpolicy},
where $\mathbf{v}\textsubscript{logstd}^t$ is a separate set of parameters referring to a log standard deviation which will be updated only during the training process. The entire input-output architecture is as shown in \prettyref{fig:model}.

\subsection{Multi-scenario multi-stage training}
In order to learn a robust policy, we present a multi-stage training scheme in varied scenarios. 

\subsubsection{\textbf{Training algorithm}} 
\label{sec:train}
Although deep reinforcement learning algorithms have been successfully applied in mobile robot motion planning, they have mainly focused on a discrete action space~\cite{zhu2017target,zhang2016deep} or on small-scale problems~\cite{tai2017virtual,chen2017decentralized,chen2017socially,kahn2017uncertainty}. Large-scale, distributed policy optimization over multiple robots and the corresponding algorithms have been less studied for mobile robot applications. 

Here we focus on learning a robust collision avoidance policy for navigating a large number of robots in complex scenarios (e.g., mazes) with static obstacles. To accomplish this, we deploy and extend a recently proposed robust policy gradient algorithm, the Proximal Policy Optimization (PPO)~\cite{schulman2017proximal,heess2017emergence,trpo}, in our multi-robot system. Our approach adapts the \textit{centralized learning, decentralized execution} paradigm. In particular, the policy is learned with experiences collected by all robots simultaneously. Each robot receives its own $\mathbf{o}^t_i$ at each time step $t$ and executes the action generated from the shared policy $\pi_{\theta}$. 

As summarized in \prettyref{alg:ppo}, which is adapted from~\cite{schulman2017proximal,heess2017emergence}, the training process alternates between sampling trajectories by executing the policy in parallel, and updating the policy with the sampled data. During data collection, each robot exploits the shared policy to generate trajectories until the maximum time period $T_{\max}$ is reached and a batch of trajectory data is sampled. Then the sampled trajectories are used to construct the surrogate loss $L\textsuperscript{PPO}(\theta)$ which is optimized with the Adam optimizer~\cite{kingma2014adam} for $E_{\pi}$ epochs under the Kullback-Leiber (KL) divergence constraint. As a baseline for estimating the advantage $\hat{A}_i^t$, the state-value function $V_{\phi}(s_i^t)$ is approximated with a neural network with parameters $\phi$ using the sampled trajectories. The network structure of $V_{\phi}$ is the same as that of the policy network $\pi_{\theta}$, except that it has only one linear activation unit in its last layer. Besides, we adopt $L_2$-Loss $L^V(\phi)$ for $V_{\phi}$, and optimize it with the Adam optimizer for $E_{V}$ epochs as well.
We update $\pi_{\theta}$ and $V_{\phi}$ independently and their parameters are not shared since we have found that using two separate networks could provide better performance in practice. 

This parallel PPO algorithm can be easily scaled to a large-scale multi-robot system with hundreds of robots in a decentralized fashion, since each robot in the team serves as an independent worker collecting data. The decentralized execution not only dramatically reduces the sampling time cost, also makes the algorithm suitable for training a large number of robots in various scenarios. In this way, our network can quickly converge to a solution with good generalization performance. 

\begin{algorithm}
\caption{PPO for multiple robots}
\label{alg:ppo}
\begin{algorithmic}[1]
\State Initialize policy network $\pi_{\theta}$ and value function $V_{\phi}(s_t)$, and set hyper-parameters as shown in \prettyref{tab:parameters}. 
\For {$\text{iteration} = 1, 2,..., $}
  \State // \textit{Collect data in parallel}
  \For{$\text{robot } i = 1, 2, ... N$}
  \State Run policy $\pi_{\theta}$ for $T_i$ time-steps, collecting $\{ \mathbf{o}_i^t, r_i^t, \mathbf{a}_i^t \} $, where $t \in [0, T_i]$
  \State Estimate advantages using GAE~\cite{schulman2015high} $\hat{A}_i^t = {\sum}_{l=0}^{T_i} ({\gamma \lambda})^l {\delta}_{i}^{t} $, where ${\delta}_{i}^{t} = r_i^t + \gamma V_{\phi}(s_i^{t+1}) - V_{\phi}(s_i^{t})$
   \State {\bf break}, if $\sum _{i=1}^N T_i > T_{\max}$
  \EndFor
  \State $\pi\textsubscript{old} \gets \pi_{\theta} $ 
  \State // \textit{Update policy}
  \For{$j = 1,..., E_{\pi}$}
    \State $ L\textsuperscript{PPO}(\theta) = \sum _{t=1}^{T_{\max}} \frac{\pi_{\theta}(a_i^t \mid o_i^t)}{\pi_{old}(a_i^t \mid o_i^t)} \hat{A}_i^t - \beta \mathrm{KL}[\pi\textsubscript{old} \mid \pi_{\theta}]+ \xi \max(0, \mathrm{KL}[\pi\textsubscript{old} \mid \pi_{\theta}] - 2\mathrm{KL}\textsubscript{target})^2 $ 
    \If{$\mathrm{KL}[\pi\textsubscript{old} \mid \pi_{\theta}] > 4\mathrm{KL}\textsubscript{target}$}
    	\State {\bf break} and continue with next iteration $i+1$
    \EndIf{}
    \State Update $\theta$ with $lr_{\theta}$ by Adam~\cite{kingma2014adam} w.r.t. $L\textsuperscript{PPO}(\theta)$ 
  \EndFor
  \State // \textit{Update value function}
  \For{$k = 1,..., E_{V}$}
    \State $ L^{V}(\phi) = -\sum _{i=1}^N \sum _{t=1}^{T_{i}}(\sum _{t'>t}\gamma ^{t'-t}r_i^{t'} - V_{\phi}(s_i^t))^2$
    \State Update $\phi$ with $lr_{\phi}$ by Adam w.r.t. $L^{V}(\phi)$ 
  \EndFor
  \State // \textit{Adapt KL penalty coefficient}
  \If{$\mathrm{KL}[\pi\textsubscript{old} \mid \pi_{\theta}] > \beta\textsubscript{high} \mathrm{KL}\textsubscript{target}$}
  	\State $\beta \gets \alpha \beta$
  \ElsIf{$\mathrm{KL}[\pi\textsubscript{old} \mid \pi_{\theta}] < \beta\textsubscript{low} \mathrm{KL}\textsubscript{target}$}
  	\State $\beta \gets \beta / \alpha$
  \EndIf
\EndFor
\end{algorithmic}
\end{algorithm}

\subsubsection{\textbf{Training scenarios}} 
\begin{figure}[t] 
\centering
\includegraphics[width=1\linewidth]{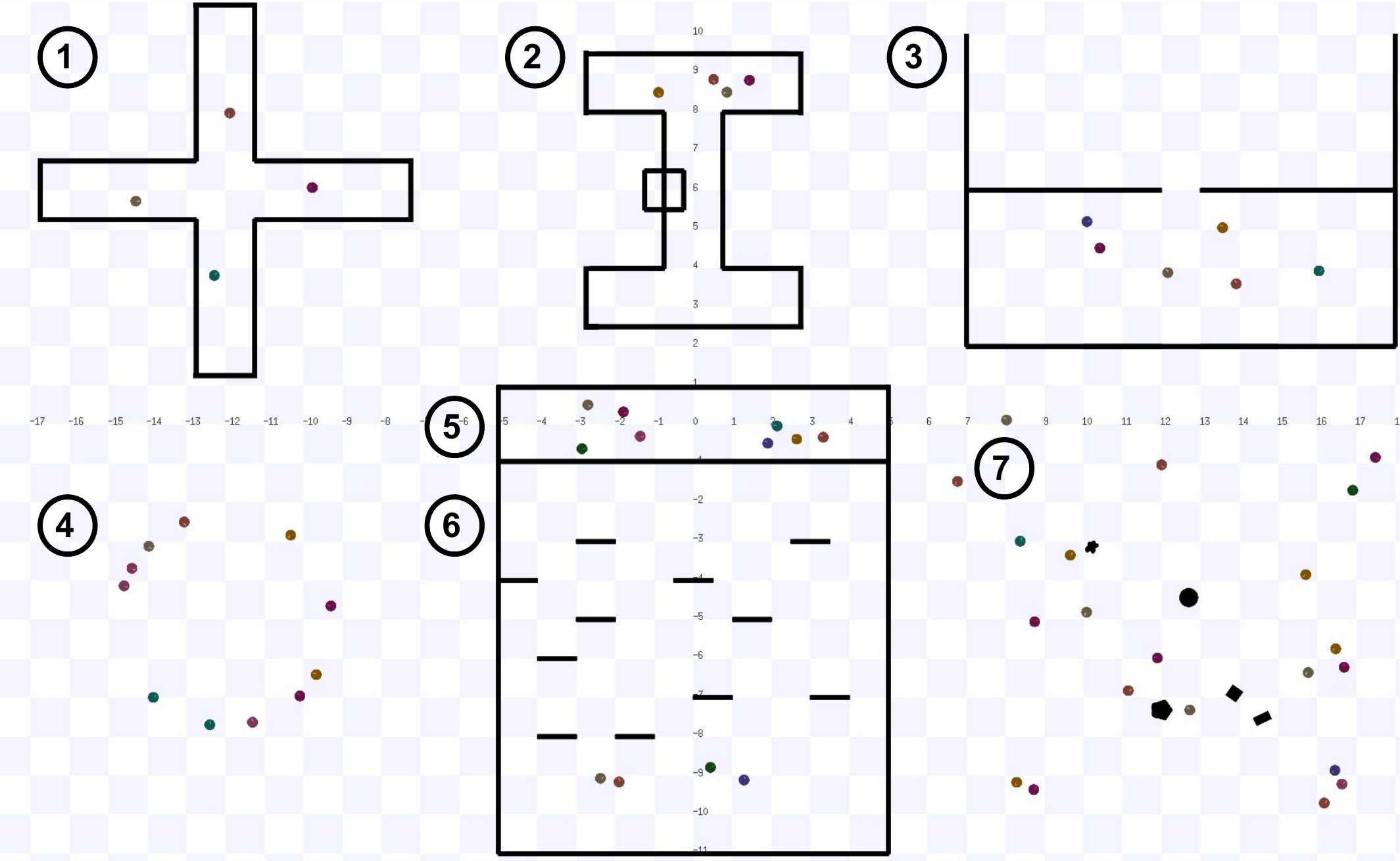}
\caption{Scenarios used to train the collision avoidance policy. All robots are modeled as a disc with the same radius. Obstacles are shown in black.}
\label{fig:scene}
\end{figure}

To achieve a generalized model that can handle different complex real world scenarios, we empirically build up multiple scenarios with a variety of obstacles using the Stage mobile robot simulator\footnote{\url{http://rtv.github.io/Stage/}} (as shown in \prettyref{fig:scene}) and move all robots concurrently. For Scenario $1$, $2$, $3$, $5$, and $6$ shown in \prettyref{fig:scene} (where the black solid lines are obstacles), we first select reasonable starting and arrival areas from the available workspace, then randomly sample the start and goal positions for each robot in the corresponding areas. In Scenario $4$, robots are randomly initialized in a circle with varied radii. Since each robot only has a limited-range observation about its surrounding environment, these robots cannot make a farsighted decision to avoid congestion. As for Scenario $7$, we generate random positions for both robots and obstacles (as marked by the black areas) at the beginning of each episode; and the goals of robots are also randomly selected. In general, these rich, complex training scenarios enable robots to explore their high-dimensional observation space and can improve the quality, robustness, and generalization of the learned policy. 

\subsubsection{\textbf{Training stages}} 
\begin{figure} 
\centering
\includegraphics[width=1\linewidth]{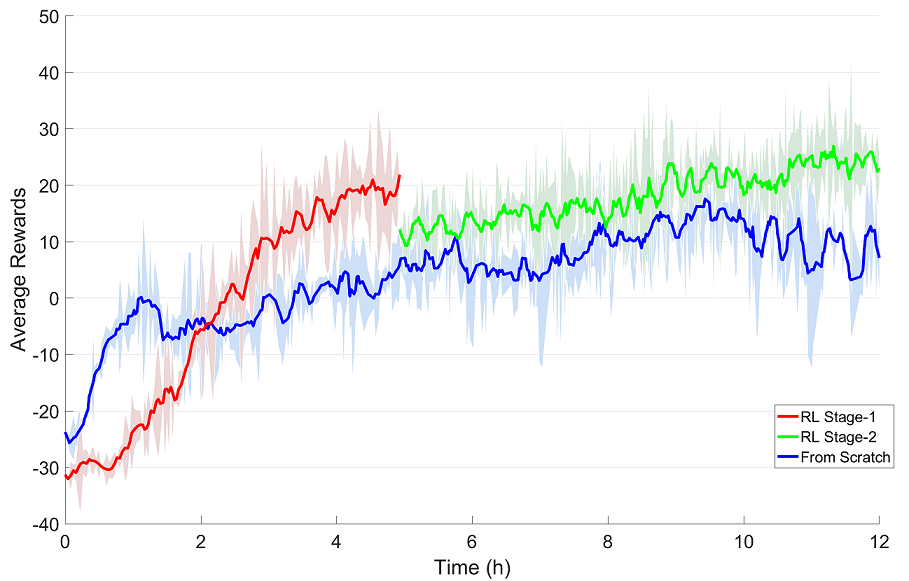}
\caption{Comparison of the average rewards shown in wall time for training in two stages (red line for the \textit{RL Stage-1 policy} and green line for the \textit{RL Stage-2 policy}) and for training from scratch (blue line). Multi-stage training scheme can achieve satisfactory performance within a shorter period of time.}
\label{fig:reward}
\end{figure}

Although training on multiple environments simultaneously brings robust performance over different test cases (as discussed in \prettyref{sec:gen}), it makes the training process harder. Inspired by the curriculum learning paradigm~\cite{bengio2009curriculum}, we propose a two-stage training process, which accelerates the policy to converge to a satisfactory solution, and gets higher rewards than the policy trained from scratch with the same number of epoch (as shown in \prettyref{fig:reward}). In the first stage, we only train $20$ robots on the random scenarios (i.e., Scenario 7 in \prettyref{fig:scene}) without any obstacles. This allows to find a reasonable solution quickly on relatively simple collision avoidance tasks. Once the robots achieve stable performance, we stop the first stage and save the trained policy. The policy will continue to be updated in the second stage, where the number of robots increases to $58$. The policy network is trained on the richer and more complex scenarios shown in \prettyref{fig:scene}. In our experiments discussed in \prettyref{sec:sim_exp}, we call the policy generated after the first stage as the \textit{RL Stage-1 Policy}, and the policy generated after the second stage as the \textit{RL Stage-2 Policy}. 

\subsection{Hybrid control architecture}
\label{sec:hybrid}
Our reinforcement learned policy generalizes well in complex scenarios in both simulation and real world. Unfortunately, it still cannot produce perfect behaviors all the time.
There exist some trivial cases that the learned policy cannot provide high quality collision avoidance decisions. For instance, as a robot runs towards its goal through a wide-open space without other agents, the robot may approach the goal in a curved trajectory rather than in a straight line. We have also observed that a robot may wander around its goal rather than directly moving toward the goal, even though the robot is already in the close proximity of the target. In addition, the learned collision avoidance policy is still more vulnerable in real world than in simulation in terms of a higher collision rate. 

These imperfect behaviors are due to several properties of the learning scheme. First, it is difficult to collect training data from some special situations, e.g., the chance to sample a straight line trajectory in our training scenarios is close to zero since the policy is modeled as a conditional probability distribution. Second, in a high-dimensional continuous action space, the probability to learn low-dimensional optimal action subspaces, e.g., turnaround in place or moving in a straight line, is also extremely low. Third, the sensor noises added during the training procedure also increase the difficulty of learning a deterministic optimal control. Finally, the sim-to-real gap may fail the collision avoidance policy in marginal cases. For instance, a robot that learned its collision avoidance capability with simulated crowds may difficult to avoid real-world pedestrians. Such sim-to-real gap is caused by the difference between the simulation and the real world, including a robot's shape, size, dynamics, and the physical properties of the workspace.

However, the situations which are challenging for the learned policy could be simple for traditional control. For example, we can use a PID controller to move a robot in a straight line or to quickly push a robot toward its goal in a wide-open space without any obstacles. Hence, we present a hybrid control framework (as shown in~\prettyref{fig:hybrid_ctrl}) that combines the learned policy and several traditional control policies to generate an effective composite policy. In particular, we first classify the scenario faced by a robot into three categories, and then design separate control sub-policies for each category. During the execution (\prettyref{alg:hybrid}), the robot heuristically switches among different sub-policies. 

\subsubsection{\textbf{Scenario classification}}
According to a robot's sensor measurement about the surrounding environment, we classify the scenarios faced by the robot into three categories: the simple scenarios, the complex scenarios, and the emergent scenarios, by using the safe radius $r\textsubscript{safe}$ and the risk radius $r\textsubscript{risk}$ as classification parameters. The classification rule is as follows. When the measured distances from the robot to all nearby obstacles are larger than $r\textsubscript{safe}$ (i.e., $\min \{\mathbf o_z\} > r\textsubscript{safe}$ and thus the robot is still far away from obstacles) or when the distance between the robot and target position is smaller than the distance to the nearest obstacle (i.e., $\mathbf o_g < \min \{\mathbf o_z\}$ and thus the robot may go straight toward the target), we classify this scenario as a simple scenario where the robot can focus on approaching the goal without worrying too much about collision avoidance.
When the robot's distance to an obstacle is smaller than $r\textsubscript{risk}$, we consider the situation as an emergent scenario because the robot is too close to the obstacle and thus needs to be cautious and conservative when making movement decisions. 
All other scenarios are classified as complex scenarios where the robot's distance to the surrounding obstacles is neither too small nor too large and the robot needs to make a good balance between approaching the goal and avoiding collisions.

In this work, $r\textsubscript{safe}$ and $r\textsubscript{risk}$ are two hyper-parameters and thus the switching rule among different scenarios is manually designed. 
Ideally, we hope to use deep reinforcement learning to learn these switch parameters automatically from data, similar to the meta-learning scheme~\cite{frans2017meta}. However, in practice we find that it is difficult to guarantee the deep reinforcement learning to recognize the emergent scenarios reliably. As a result, in this work, we only use the $r\textsubscript{safe}$ and $r\textsubscript{risk}$ as hyper-parameters to robustly distinguish different scenarios, and leave the meta-learning as the future work.

\begin{figure*}[!htb] 
\centering
\includegraphics[width=0.8\linewidth]{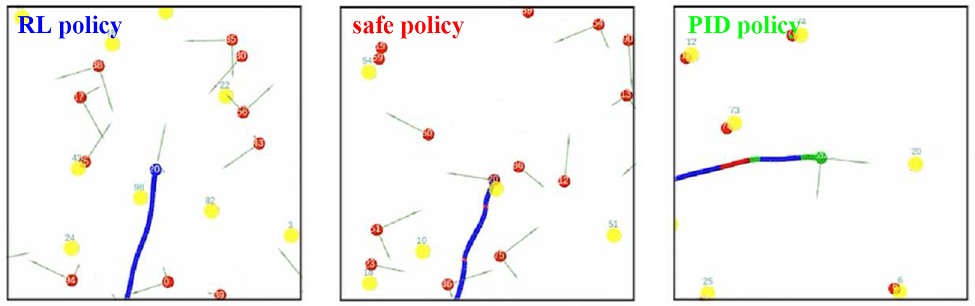}
\caption{Illustration of the three sub-policies in the hybrid control framework and their corresponding situations. We use different colors to indicate the different sub-policy taken at a given point on the trajectory. The blue color means that the robot is taking the RL policy $\pi\textsubscript{RL}$ to balance the navigation efficiency and safety in a situation that is neither too crowd nor too open. The red color indicates that the robot is taking the safe policy $\pi\textsubscript{safe}$ to deal with the obstacles that are very close. The green color means that the robot is in an open space and is taking PID control $\pi\textsubscript{PID}$ to approach its target quickly. Here the red points are the agents' current positions and the yellow points are the agents' goals. }
\label{fig:hybrid_scenarios}
\end{figure*}

\subsubsection{\textbf{Sub-policies}}
Each of the three scenarios discussed above will be handled by one separate sub-policy. For the simple scenario, we will apply the PID control law $\pi\textsubscript{PID}$, which provides an almost optimal solution to move the robot toward its target and guarantees stability. For the emergent scenario, we apply the conservative safe policy $\pi\textsubscript{safe}$ as shown in \prettyref{alg:safe} to generate a conservative movement by leveraging the DRL trained policy but scaling the neural network's input and restricting the output of the robot.
For the complex scenario, we use the reinforcement learned policy $\pi\textsubscript{RL}$ directly, which uses multi-scenario multi-stage reinforcement learning framework to achieve the state-of-the-art navigation performance. An intuitive illustration of the three sub-policies and their corresponding situations are provided in \prettyref{fig:hybrid_scenarios}.

\begin{algorithm}
\caption{Hybrid controller for multi-robot collision avoidance}
\label{alg:hybrid}
\begin{algorithmic}[1]
 \renewcommand{\algorithmicrequire}{\textbf{Input:}}
 \renewcommand{\algorithmicensure}{\textbf{Output:}}
 \Require the 2D laser measurements $\mathbf o^t_z$ and the distance between the robot's position to its goal $\mathbf o^t_g$
  \Ensure the robot steering command $\mathbf a^t$
\State Configure three sub-policies: the PID control policy $\pi\textsubscript{PID}$, the RL trained policy $\pi\textsubscript{RL}$ and the conservative safety policy $\pi\textsubscript{safe}$; set the safety radius $r\textsubscript{safe}$ and the emergent radius $r\textsubscript{risk}$

\If{$\min\{\mathbf o^t_z\} > r\textsubscript{safe}$ or $\min\{\mathbf o^t_z\} > \mathbf o^t_g$}
	\State  $\mathbf a^t \gets \pi\textsubscript{PID}$
\ElsIf{$\min\{\mathbf o^t_z\} \leq r\textsubscript{risk}$}
	\State $\mathbf a^t \gets \pi\textsubscript{safe}$
\Else{}
	\State $\mathbf a^t \gets \pi\textsubscript{RL}$
\EndIf{}
\State \Return $\mathbf a^t$
\end{algorithmic}
\end{algorithm}

\begin{algorithm}
\caption{Conservative safety policy}
\label{alg:safe}
\begin{algorithmic}[1]
 \renewcommand{\algorithmicrequire}{\textbf{Input:}}
 \renewcommand{\algorithmicensure}{\textbf{Output:}}
\Require the observation $\mathbf{o}^t$ stated in \prettyref{sec:obs}.
\Ensure the robot steering command $\mathbf a^t$
\State Set the scale parameter $\mathbf{p}_{scale}$ and the maximal moving velocity $\mathbf v_{\max}$ 
\If{$\mathbf o^t_v > \mathbf v_{\max}$}
\State $\mathbf a^t \gets (0, 0)$
\Else{}
\State $\hat{\mathbf o}^t_z = \mathbf o^t_z / \mathbf{p}_{scale}$
\State $\mathbf a^t \gets \pi\textsubscript{DRL}(\hat{\mathbf o}^t_z, \mathbf o^t_g, \mathbf o^t_v)$
\State $\mathbf a^t \gets \text{clip}(\mathbf a^t, -\mathbf v_{\max}, \mathbf v_{\max})$
\EndIf{}
\State \Return $\mathbf a^t$
\end{algorithmic}
\end{algorithm}

\subsubsection{\textbf{Hybrid control behaviors}}
\begin{figure}[!htb]
\begin{subfigure}[b]{0.49\linewidth}
\includegraphics[trim=90 20 40 0, clip, width=1\linewidth]{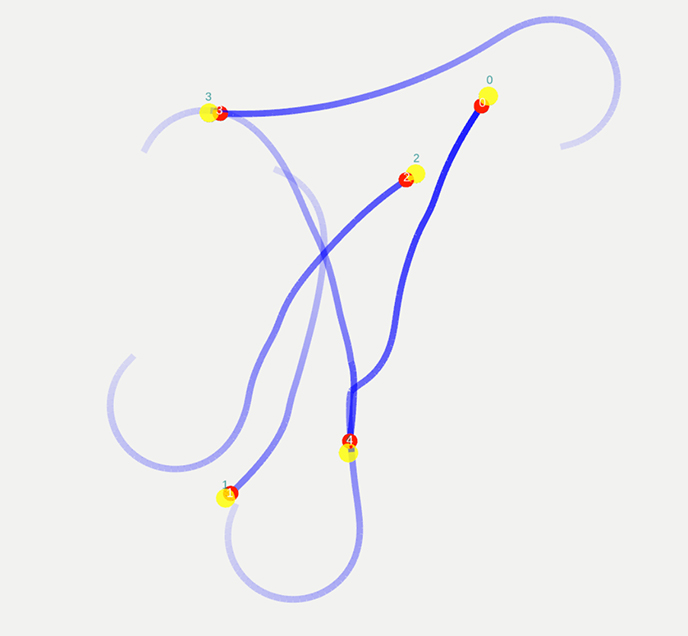}
\caption{Random scene RL}
\label{fig:random_5_rl_traj}
\end{subfigure}
\begin{subfigure}[b]{0.49\linewidth}
\includegraphics[trim=90 20 40 0, clip, width=1\linewidth]{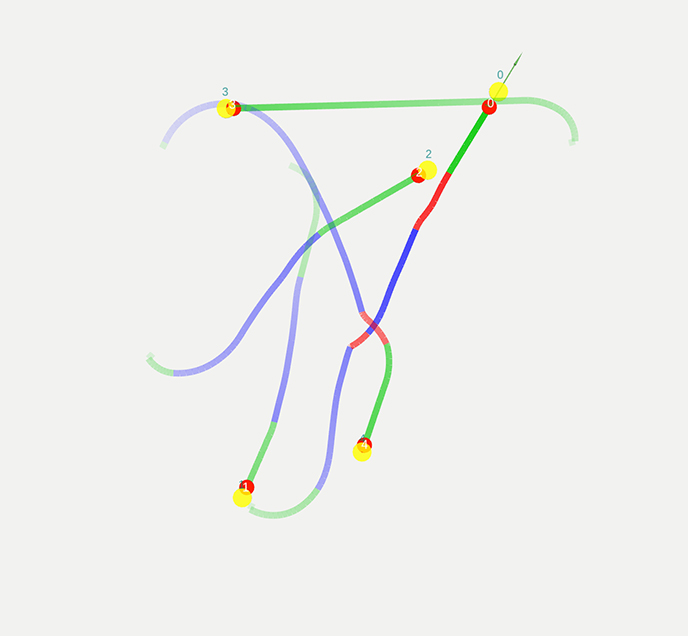}
\caption{Random scene Hybrid-RL}
\label{fig:random_5_hybrid_traj}
\end{subfigure}
\begin{subfigure}[b]{0.49\linewidth}
\includegraphics[trim=80 40 120 40, clip, width=1\linewidth]{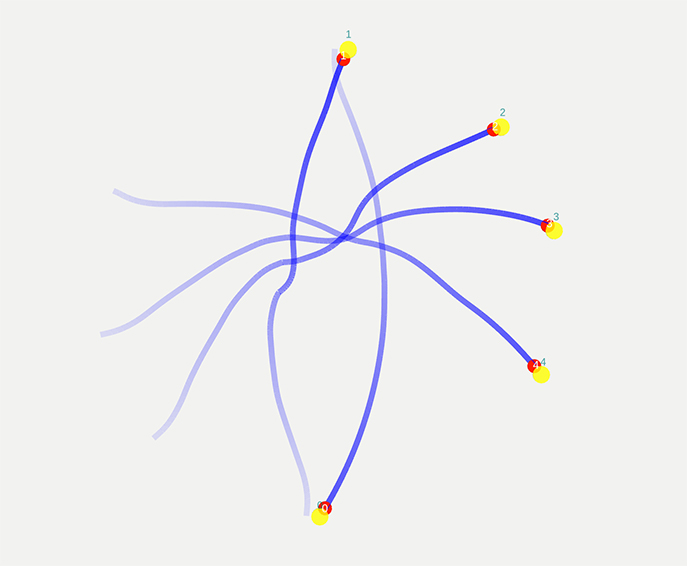}
\caption{Circle scene RL}
\label{fig:random_5_circle_traj}
\end{subfigure}
\begin{subfigure}[b]{0.49\linewidth}
\includegraphics[trim=80 40 120 40, clip, width=1\linewidth]{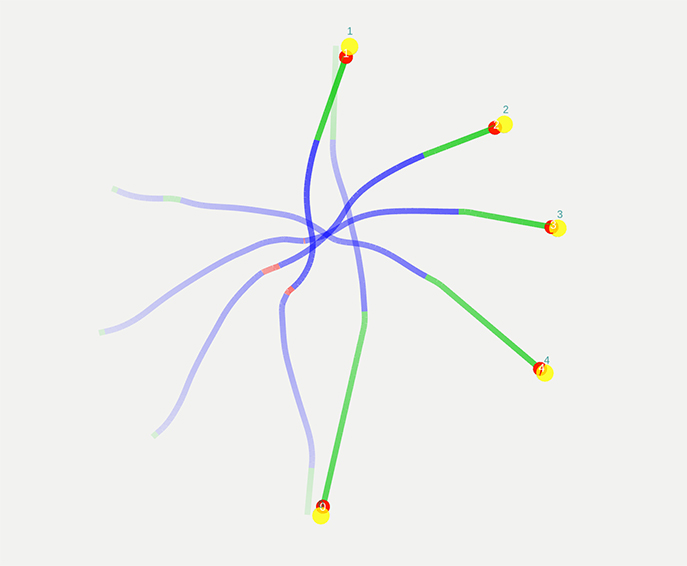}
\caption{Circle scene Hybrid-RL}
\label{fig:random_circle_hybrid_traj}
\end{subfigure}
\caption{Comparison of trajectories generated by RL policy and Hybrid-RL policy for a group of $5$ agents in the random scene (top row) and in the circle scene (bottom row). In the random scene, the agents are starting from random positions and moving toward random goals. (a) shows the trajectories generated by the RL policy and (b) shows the trajectories generated by the Hybrid-RL policy. In the circle scene, the agents are starting from positions on the circle and moving toward antipodal positions. (c) shows the trajectories generated by the RL policy and (d) shows the trajectories generated by the Hybrid-RL policy. In (b) and (d), we use three colors to distinguish different sub-policies chosen by each robot at run-time: red for the safe policy  $\pi\textsubscript{safe}$, green for the PID policy $\pi\textsubscript{PID}$, and blue for the learned policy $\pi\textsubscript{RL}$. In all figures, the color transparency is used to indicate the timing along a trajectory.}
\label{fig:hybrid_rl_traj}
\end{figure}

In \prettyref{fig:hybrid_rl_traj}, we compare different behaviors of the reinforcement learning policies with or without using the hybrid control, which are denoted as RL policy and Hybrid-RL policy, respectively. In both benchmarks, we can observe that Hybrid-RL will switch to $\pi\textsubscript{PID}$ in the open space, which helps the robot to go toward the goal in the most efficient way, rather than taking the curved paths as the RL policy. In cases when the collision risk is high, the agent will switch to $\pi\textsubscript{safe}$ to guarantee safety. For other cases, the agents will enjoy the flexibility of $\pi\textsubscript{RL}$. The switch among sub-policies allows robots to fully leverage the available space by taking trajectories with short lengths and small curvatures. More experimental results verifying the effectiveness of the hybrid control architecture are also available latter in \prettyref{sec:sim_exp}.

\section{Simulation Experiments}
\label{sec:sim_exp}
In this section, we first describe the setup details for the policy training, including the simulation setup, the hyper-parameters for the deep reinforcement learning algorithm and the training time. Next, we provide a thorough evaluation of the performance of our hybrid multi-scenario multi-stage deep reinforcement learned policy by comparing with other approaches. In particular, we compare four approaches in the experiments:
\begin{itemize}
\item{\textit{NH-ORCA policy}} is the state-of-the-art rule-based approach proposed by~\cite{alonso2013optimal,hennes2012multi,claes2012collision}
\item{\textit{SL policy}} is trained with
the supervised learning approach proposed by~\cite{long2017deep}
\item{\textit{RL policy}} is the original multi-scenario multi-stage reinforcement learned policy without the hybrid architecture, as proposed by~\cite{long2017towards}
\item{\textit{Hybrid-RL policy}} is our proposed method in this paper, which augments the RL policy with the hybrid control.
\end{itemize}
For the \textit{RL policy}, in some experiments, we analyze its two stages -- the \textit{RL Stage-1 policy} and \textit{RL Stage-2 policy} -- separately, in order to better understand its behavior. 

The comparison of these policies is performed from different perspectives, including the generalization to unseen scenarios and tasks, the navigation efficiency, and the robustness to the agent density and to the robots' inaccurate shape and dynamics. These experiments demonstrate the superiority of our proposed methods over previous methods.

\subsection{Training setup}
The training of our policy is implemented in TensorFlow and the large-scale multi-robot system with the 2D laser scanner measurements is simulated in the Stage simulator. The simulation time step is set as \SI{0.1}{s} and the robot's control frequency is \SI{10}{Hz}. 
We train the multi-robot collision avoidance policy on a computer with one i7-7700 CPU and one Nvidia GTX 1080 GPU. The offline training takes $12$ hours to run about $600$ iterations in \prettyref{alg:ppo} so that the policy converges in all the training scenarios. The hyper-parameters in \prettyref{alg:ppo} are summarized in \prettyref{tab:parameters}. Specifically, the learning rate $lr_{\theta}$ of the policy network is set to $5e\minus5$ in the first training stage, and is then reduced to $2e\minus5$ in the second training stage. The hyper-parameters for the hybrid control are summarized in \prettyref{tab:hybrid_parameters}.

The online execution of the learned policy for both simulation and the real-world robots runs in real-time. In simulation, it takes \SI{3}{ms} on CPU or \SI{1.3}{ms} on GPU for the policy network to compute new actions for $10$ robots. Deployed on a physical robot with one Nvidia Jetson TX1, the policy network takes about \SI{5}{ms} to generate online robot control commands.

\begin{table}
\rowcolors{1}{gray!25}{}
	\begin{tabularx}{0.48\textwidth}{m{1.09in}l}
 	\hline
  	\textbf{Parameter} & \textbf{Value}  \\
   	\hline
   	\text{$\lambda$ in line 6} & 0.95  \\
  	\text{$\gamma$ in line 6 and 20} & 0.99 \\
   	\text{$T_{\max}$ in line 7} & 8000 \\
   	\text{$E_{\phi}$ in line 11} & 20 \\
   	\text{$\beta$ in line 12} & 1.0 \\
   	\text{$\mathrm{KL}\textsubscript{target}$ in line 12} & $15e\minus 4$ \\
   	\text{$\xi$ in line 12} & 50.0 \\
   	\text{$lr_{\theta}$ in line 16} & $5e\minus 5$ ($1$st stage), $2e\minus 5$ ($2$nd stage) \\
   	\text{$E_V$ in line 19} & 10 \\
   	\text{$lr_{\phi}$ in line 21} & $1e\minus 3$ \\ 
   	\text{$\beta\textsubscript{high}$ in line 24} & 2.0 \\
   \text{$\alpha$ in line 24 and 27} & 1.5 \\  
   \text{$\beta\textsubscript{low}$ in line 26} & 0.5 \\
   \hline
 \end{tabularx}
\caption{Hyper-parameters of our training algorithm described in \prettyref{alg:ppo}. }
\label{tab:parameters}
\end{table}

\begin{table}
\rowcolors{1}{gray!25}{}
 \begin{tabularx}{0.48\textwidth}{m{1.09in}l}
   \hline
  \textbf{Parameter} & \textbf{Value}  \\
   \hline
   \text{$r\textsubscript{safe}$} & 0.1 \quad \quad \quad \quad \quad \quad \quad \quad \quad \quad \quad \quad \quad   \\  
   \text{$r\textsubscript{risk}$} & 0.8 \\
   \text{$\mathbf{p}\textsubscript{scale}$} & 1.25 \\
   \text{$\mathbf{v}_{\max}$} & 0.5 \\
   \hline
 \end{tabularx}
\caption{Hyper-parameters of our hybrid control system described in \prettyref{alg:hybrid} and \prettyref{alg:safe}.}
\label{tab:hybrid_parameters}
\end{table}

\subsection{Generalization capability}
\label{sec:gen}
A notable benefit of the multi-scenario training is the excellent generalization capability of our learned policy after the two-stage training. Thus, we first demonstrate the generalization of the RL policy with a series of experiments. 
\subsubsection{\textbf{Non-cooperative robots}} As mentioned in \prettyref{sec:prob}, our policy is trained on a team of robots, in which all robots share the same collision avoidance strategy. Non-cooperative robots are not introduced in the entire training process. However, as shown in \prettyref{fig:dynamic}, our learned policy can directly generalize to avoid non-cooperative agents, i.e., the rectangle-shaped robots which travel in straight lines with a fixed speed. 

\subsubsection{\textbf{Heterogeneous robots}} 
\begin{figure*}[t]
\centering
\begin{minipage}[b]{0.47\textwidth}
\centering
\includegraphics[width=.75\linewidth]{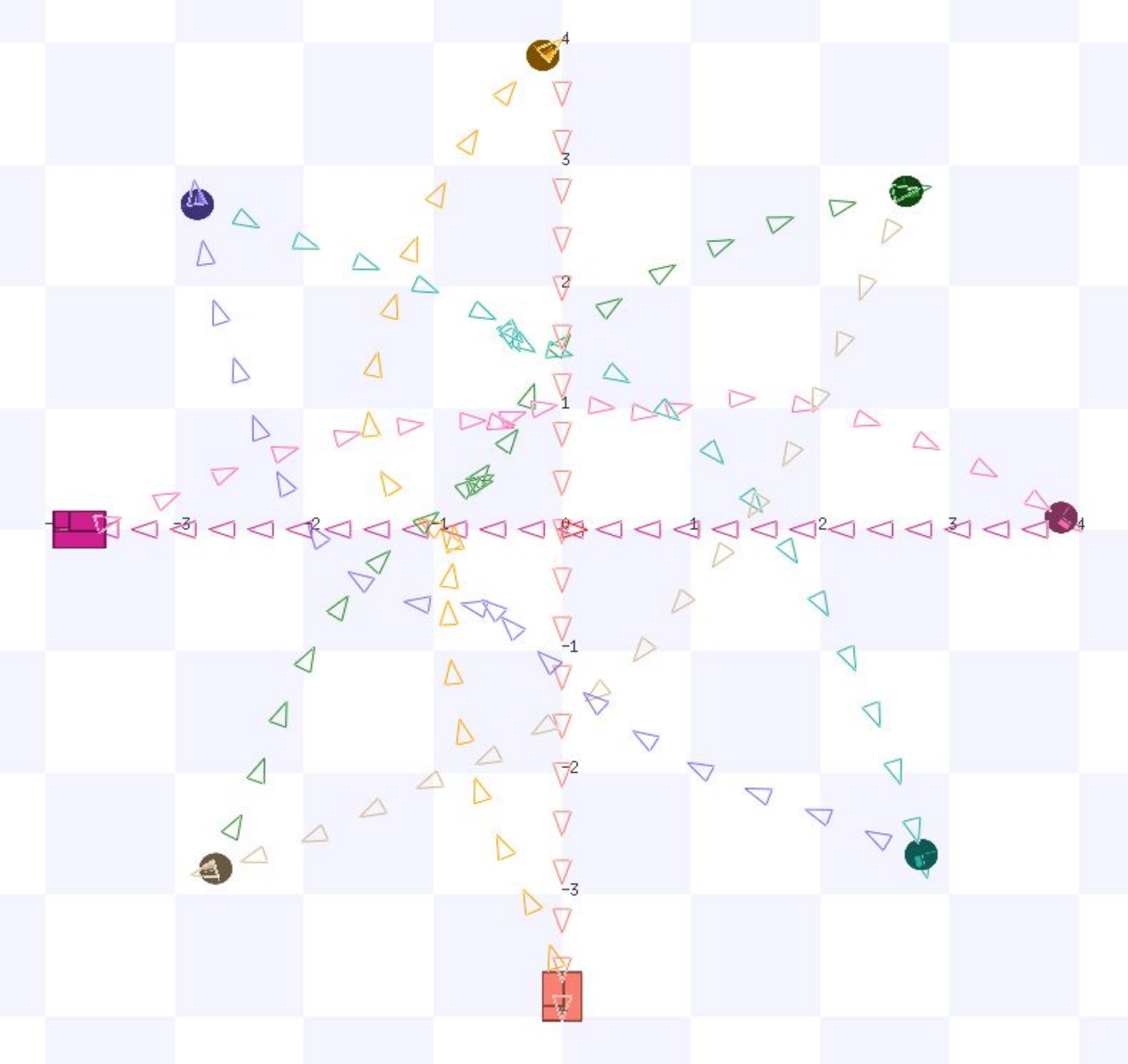}
\caption{Six disc-shaped robots controlled by the learned policy interact with two non-cooperative robots with the rectangle shape. The non-cooperative robots are traveling in straight lines with fixed high speed.}
\label{fig:dynamic}
\end{minipage}
\qquad
\begin{minipage}[b]{0.47\textwidth}
\centering
\includegraphics[width=.75\linewidth]{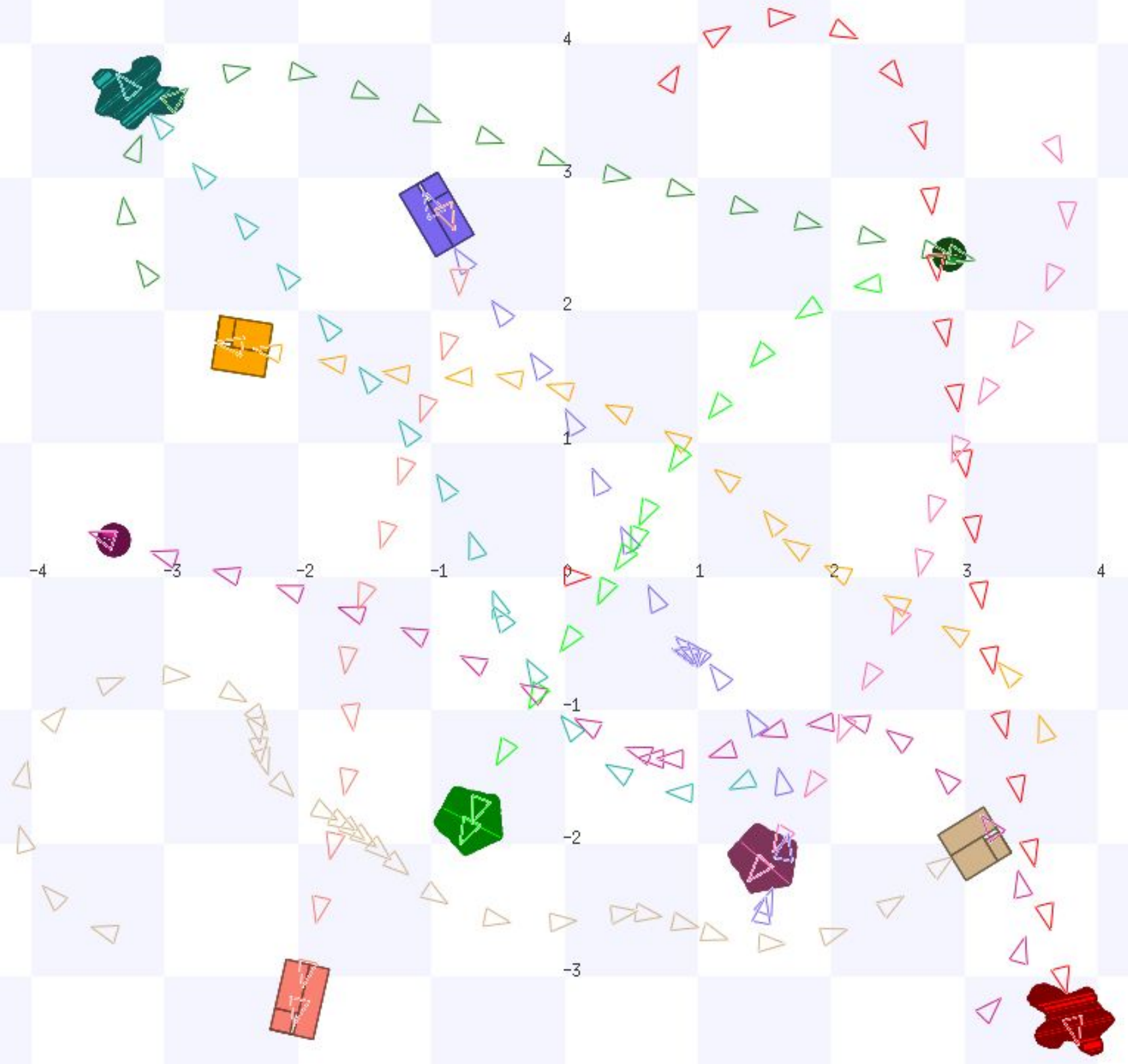}
\caption{Illustration of the collision-free and smooth trajectories of ten heterogeneous robots with different shapes and sizes. All robots adopt the same navigation policy which is trained by the prototype disc-shaped robots.}
\label{fig:hete}
\end{minipage}
\end{figure*}

Our policy is trained on robots with the same disc shape and a fixed radius. \prettyref{fig:hete} demonstrates that the learned policy can efficiently navigate a group of heterogeneous robots with different sizes and shapes to reach their goals without any collisions. 

\subsubsection{\textbf{Large-scale scenarios}} 
\begin{figure*}[t] 
\centering
\captionsetup[subfigure]{justification=centering}
\centering
\begin{subfigure}[b]{0.45\textwidth}
\includegraphics[width=.9\linewidth]{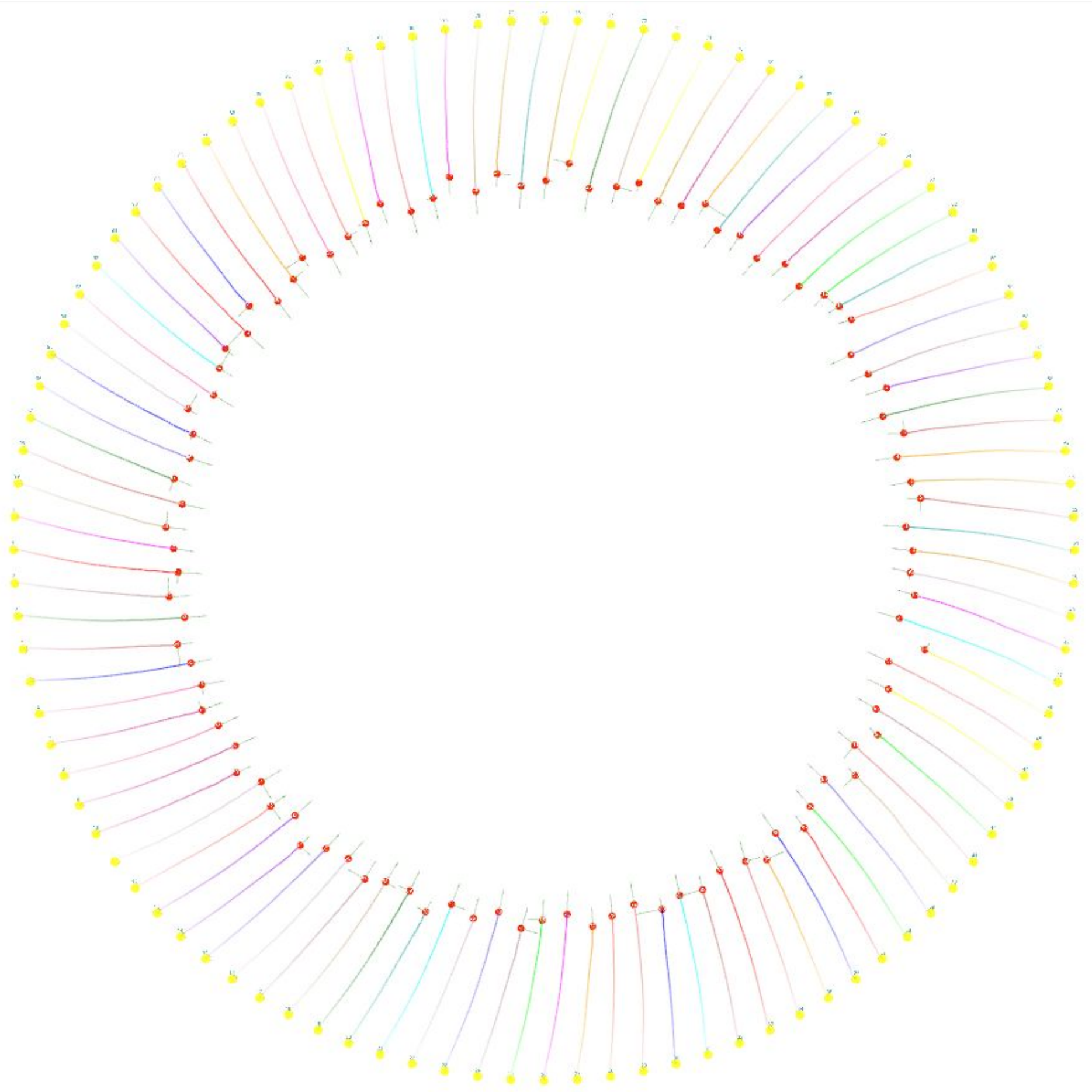}
\caption{$100$ robots (in red) are moving to their goal positions (in yellow).}
\label{fig:100_start}
\end{subfigure}
\qquad
\begin{subfigure}[b]{0.45\textwidth}
\includegraphics[width=.9\linewidth]{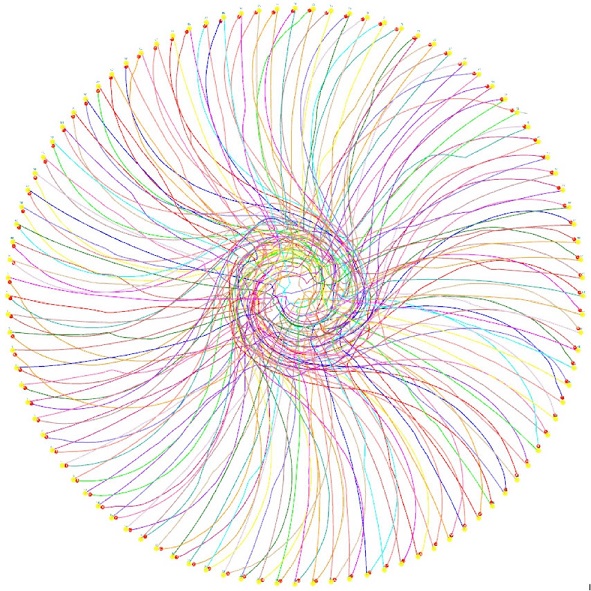}
\caption{The final trajectories of $100$ robots arriving at their goal positions using the RL policy.}
\label{fig:100_end}
\end{subfigure}
\begin{subfigure}[b]{0.45\textwidth}
\includegraphics[width=.9\linewidth]{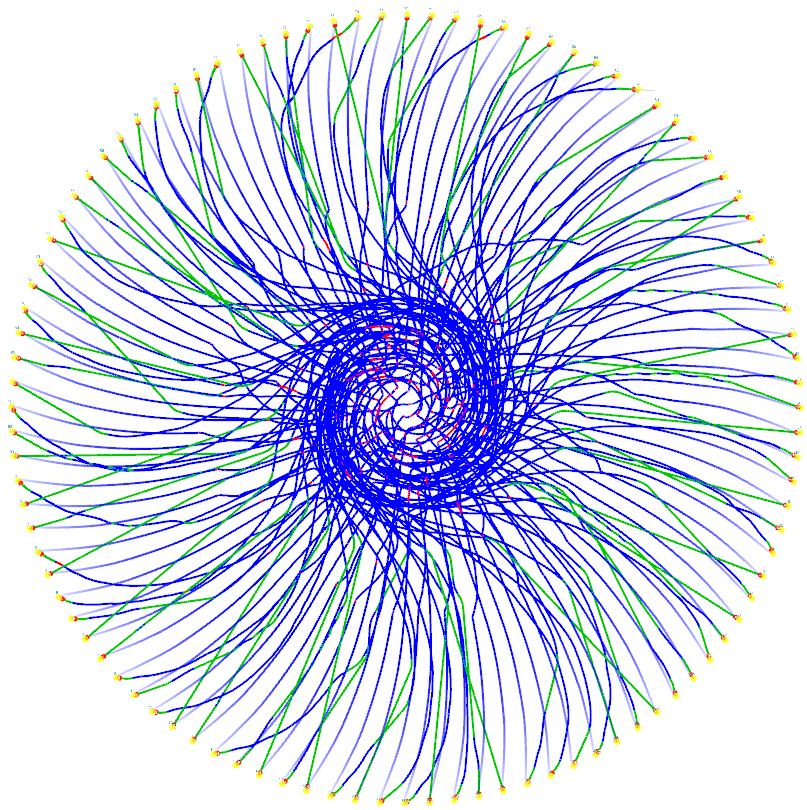}
\caption{The final trajectories of $100$ robots arriving at their goal positions using the Hybrid-RL policy.}
\label{fig:hybrid100_end}
\end{subfigure}
\caption{Simulation of $100$ robots which are uniformly placed around a circle in the beginning and are trying to move through the center of a circle toward the antipodal positions. The red points are the robots' current positions. The yellow points are the robots' goals on the circle. In (b), we use different colors to distinguish trajectories of different robots and use the color transparency to indicate the timing along a trajectory sequence. In (c), we use three colors to distinguish the different sub-polices chosen by each robot in run-time: red for the safe policy $\pi\textsubscript{safe}$, green for the PID policy $\pi\textsubscript{PID}$, and blue for the learned policy $\pi\textsubscript{RL}$. 
Please also refer to the video for the detailed comparison between the trajectories of RL policy and Hybrid-RL policy.}
\label{fig:100_robots}
\end{figure*}

\begin{figure*}[h] 
\centering
\captionsetup[subfigure]{justification=centering}
\begin{subfigure}[b]{0.46\textwidth}
\includegraphics[width=.9\linewidth]{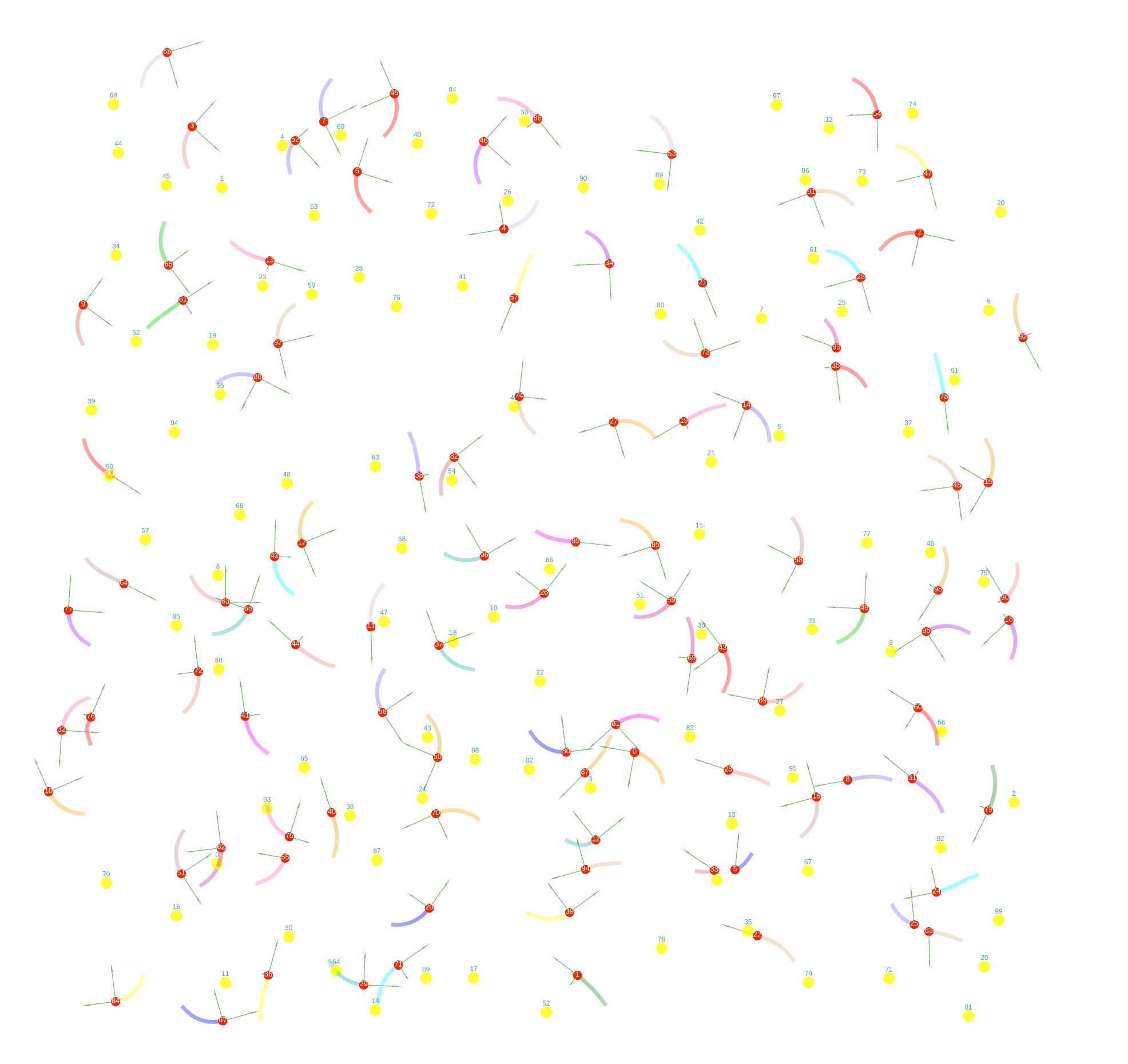}
\caption{$100$ robots (in red) move from their random start positions to their own random goals (in yellow).}
\label{fig:random_start_100}
\end{subfigure}
\qquad
\begin{subfigure}[b]{0.46\textwidth}
\includegraphics[width=.9\linewidth]{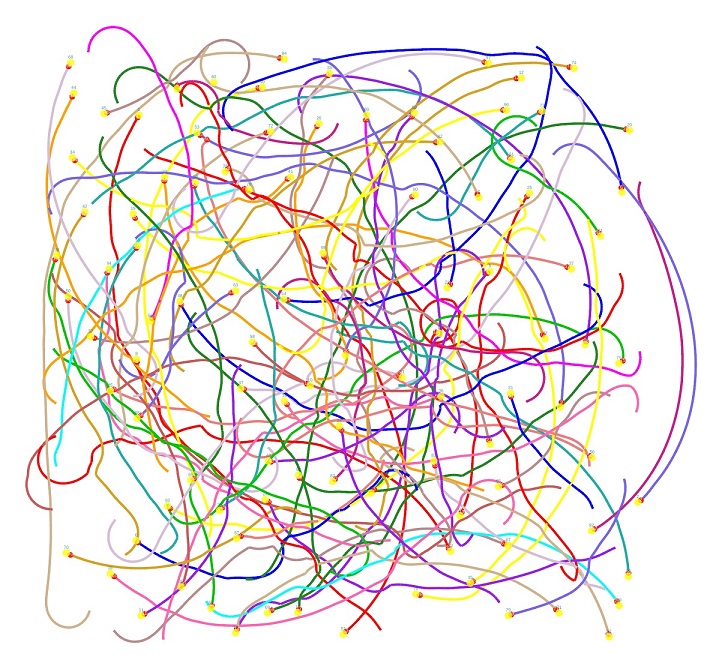}
\caption{The final trajectories of $100$ robots arriving at their goal positions using Hybrid-RL policy. }
\label{fig:random_rl_100}
\end{subfigure}
\begin{subfigure}[b]{0.46\textwidth}
\includegraphics[width=.9\linewidth]{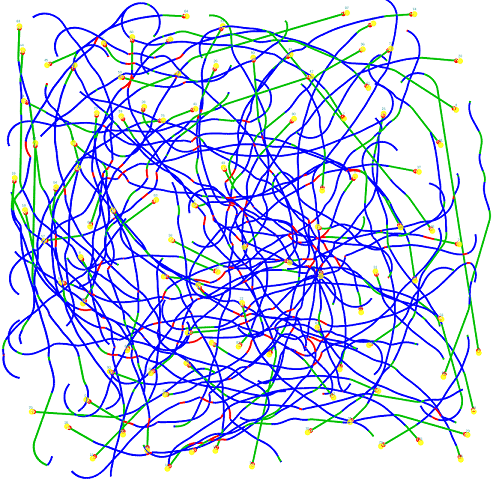}
\caption{Illustration of the sub-polices chosen by the robots in run-time using Hybrid-RL policy. }
\label{fig:random_hybrid_100}
\end{subfigure}
\caption{Simulation of $100$ robots that move from their random start positions to their own random goals. The red points are the robots' current positions and the yellow points are the robots' destinations. In (b), we use different colors to distinguish the trajectories of different robots and use the color transparency to indicate the propagation of each trajectory. In (c), we use three colors to distinguish the different sub-polices chosen by each robot in run-time: red for the safe policy $\pi\textsubscript{safe}$, green for the PID policy $\pi\textsubscript{PID}$, and blue for the learned policy $\pi\textsubscript{RL}$. 
Please also refer to the video for the detailed comparison between RL policy and Hybrid-RL policy.}
\label{fig:100_random}
\end{figure*}

To evaluate the performance of our method on large-scale scenarios, we simulate $100$ robots in a large circle moving to antipodal positions as shown in \prettyref{fig:100_end}. This simulation experiment serves as a pressure test for our learned collision avoidance policy, since the robots may run into congestion easily
due to their limited sensing about the environment. The result shows that our learned policy generalize well to large-scale environments without any fine-tuning. In addition, we also simulate $100$ robots in the random scenario as shown in \prettyref{fig:random_rl_100}. All these robots are able to arrive at their destinations without any collisions.

Note that the good generalization capability of the RL policy is also preserved in the Hybrid-RL policy, since all decisions in the complex and emergent situations are made by the RL policy. The generalization of the Hybrid-RL policy is also verified by the pressure test of $100$ robots, as shown in \prettyref{fig:hybrid100_end} and \prettyref{fig:random_hybrid_100}. In both scenarios, most robots adopt the $\pi\textsubscript{PID}$ policy near the start positions and goals (shown as green lines in \prettyref{fig:hybrid100_end} and \prettyref{fig:random_hybrid_100}) due to the considerable distance to its neighbors. They then switch to the $\pi\textsubscript{RL}$ or $\pi\textsubscript{safe}$ policy when they encounter other robots (shown as blue or red lines in \prettyref{fig:hybrid100_end} and \prettyref{fig:random_hybrid_100}).

\subsection{Efficiency evaluation}
\label{sec:efficiency}
In this section, we evaluate the performance of the proposed hybrid collision avoidance policy in terms of the navigation efficiency.
In particular, we first present multiple performance metrics for evaluating the navigation efficiency, and then compare different polices. 

\subsubsection{\textbf{Performance metrics}}
To compare the performance of our approach with other methods, we use the following metrics:
\renewcommand{\labelitemi}{\textbullet}
\begin{itemize}
\item \textit{Success rate} is the ratio of the number of robots reaching their goals in a certain time limit without any collisions to the total number of robots. 
\item \textit{Extra time $\bar{t}_e$} measures the difference between the average travel time of all robots and the lower bound of robots' travel time which is computed as the average travel time of going straight toward goals at the maximum speed without collision avoidance~\cite{godoy2016implicit,chen2017decentralized}.
\item \textit{Extra distance $\bar{d}_e$} measures the difference between the average traveled trajectory length of all robots and the lower bound of robots' traveled distance which is computed as the average traveled distance for robots following the shortest paths toward their goals.
\item \textit{Average speed $\bar{v}$} measures the average speed of all robots during the navigation.
\end{itemize}
Note that in the definitions of the \textit{extra time} $\bar{t}_e$ and \textit{extra distance} $\bar{d}_e$, we use the lower-bound baseline to alleviate the biases brought by the different number of agents and different distances to goals. Hence, we can provide a fair comparison among different methods, and this is important especially for the random scenarios.

Since all four policies in comparison (the NH-ORCA policy, the SL policy, the RL policy, and the Hybrid-RL policy) contain some randomness, we run each method $50$ times for each test scenario, and then report the mean and variance for each metric.

\subsubsection{\textbf{Circle scenarios}}
We first compare our Hybrid-RL policy with other methods on several circle scenarios with different number of robots. These scenarios are similar to the Scenario $4$ in \prettyref{fig:scene}, except that the initial positions of these robots are uniformly along the circle.
We test in seven circle scenarios with 4, 6, 8, 10, 12, 15 and 20 agents each. Accordingly, we adjust the circle radius in each scenario to make sure that its average agent density (i.e. the number of agents divided by the area of the circle) is similar (around $0.2~\text{agent}$\SI{}{/m^2}).  

We use the open-source NH-ORCA implementation from~\cite{hennes2012multi,claes2012collision} in the evaluation. Since the agents executing the NH-ORCA policy cannot make decisions based on the raw sensor measurements, we need to share the ground truth positions and velocities for all robots in the simulation, such information is not available for the learning-based approaches.
The SL policy learned a neural network model the same as the Hybrid-RL policy (as described in \prettyref{sec:model}),  and it is trained in a supervised mode on about $800,000$ samples using the same training strategy as~\cite{long2017deep,pfeiffer2017perception}.

The comparison results are summarized in \prettyref{tab:1circle}.
Compared to the NH-ORCA policy, the reinforcement learned policies (the RL policy and the Hybrid-RL policy) have significant improvements in terms of the success rate, while the SL policy has the lowest success rate. For the largest test scenario with $20$ robots, the Hybrid-RL policy still achieves $100$\% success rate while the RL policy has a non-zero failure rate.
In addition, both the RL policy and the Hybrid-RL policy are superior to the NH-ORCA policy in the average extra time and the travel speed, which means that the robots executing the RL-based policies can provide a higher throughput. And the Hybrid-RL policy performs better than the RL policy according to these two metrics. 

\begin{figure*}
\centering
\captionsetup[subfigure]{justification=centering}
\begin{subfigure}{0.32\textwidth}
\includegraphics[trim=15 5 20 15, clip, width=1.0\linewidth]{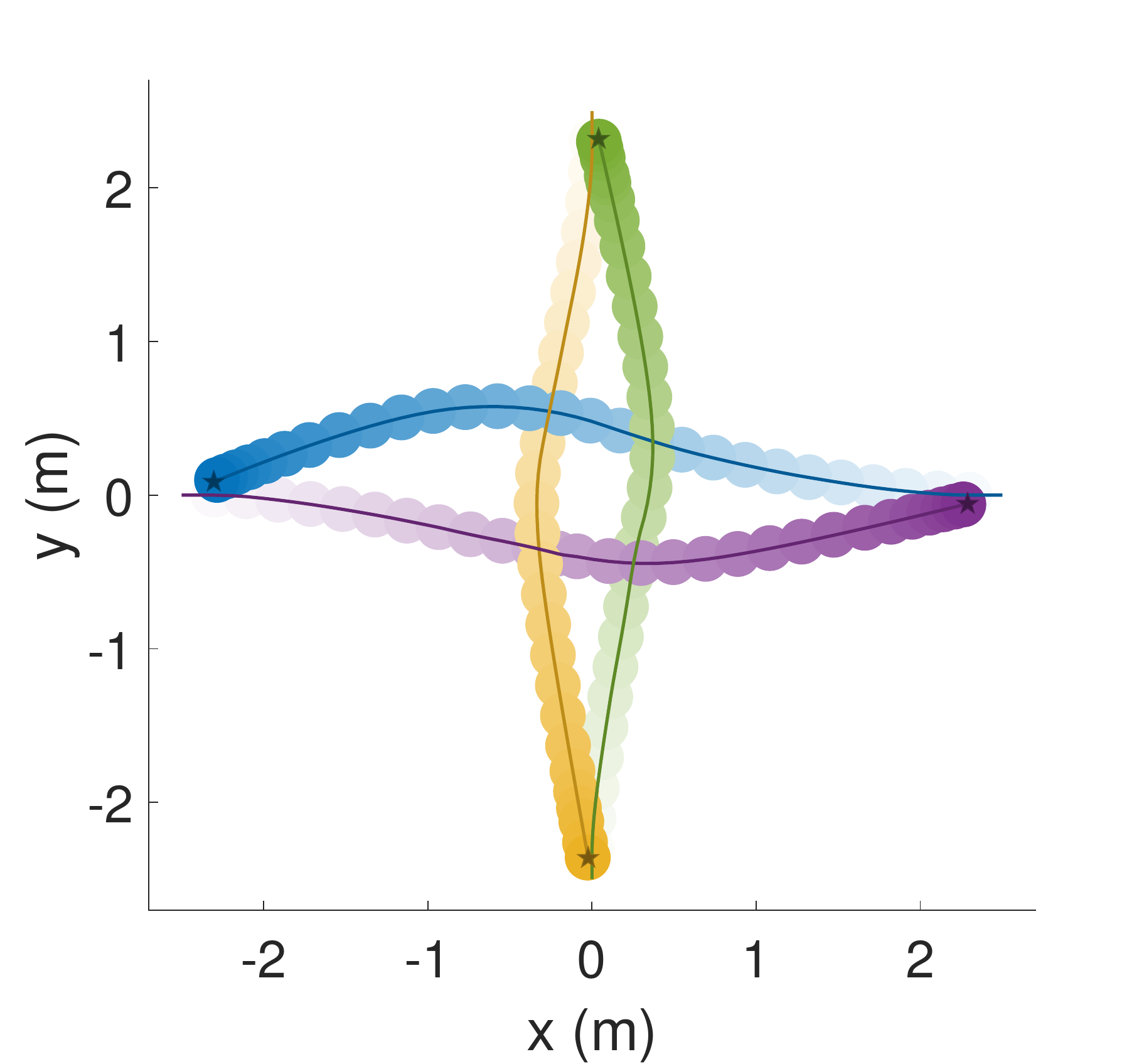}
\label{fig:orca_4_traj}
\end{subfigure}
\begin{subfigure}{0.32\textwidth}
\includegraphics[trim=15 5 20 15, clip, width=1.0\linewidth]{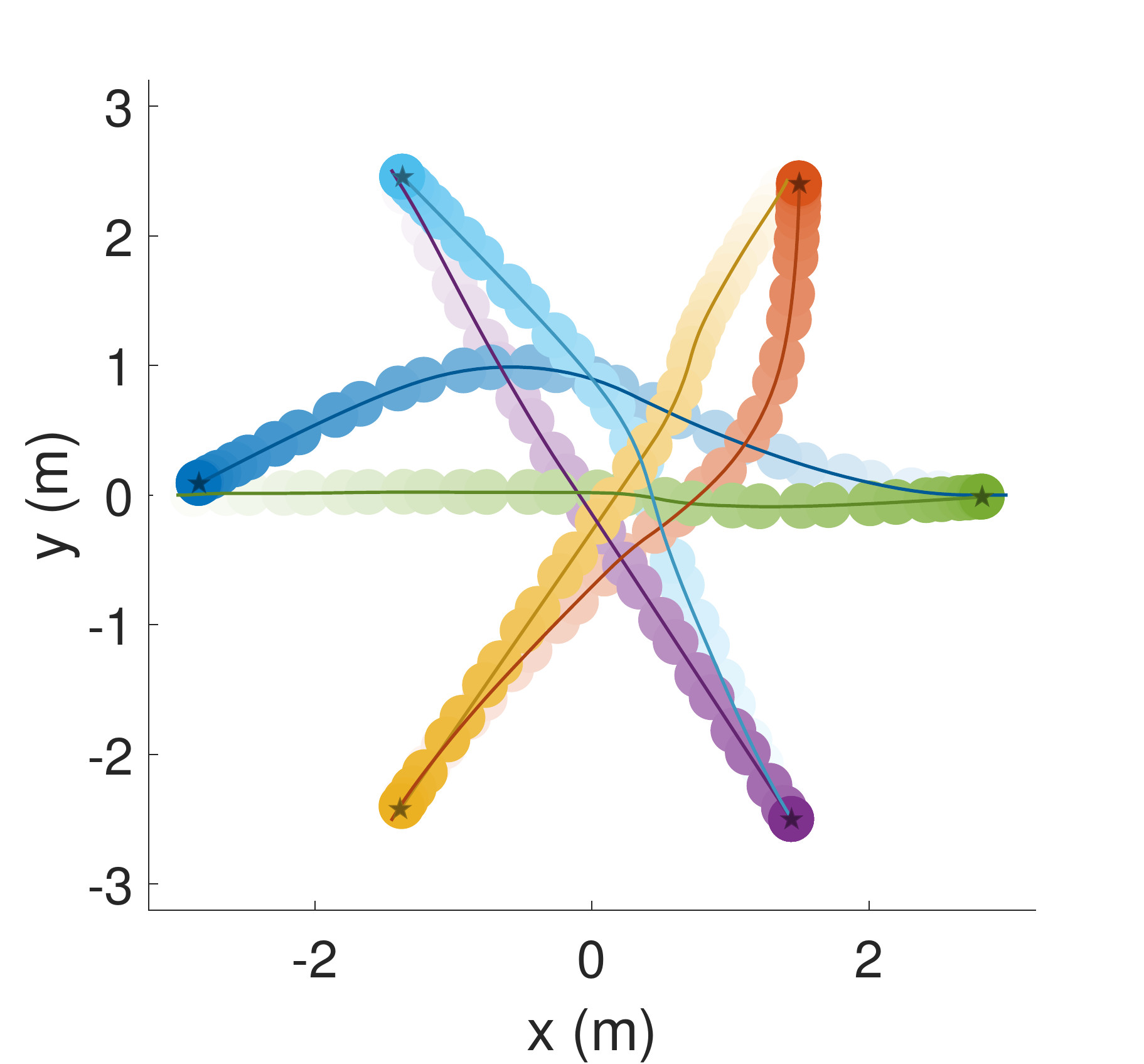}
\label{fig:orca_6_traj}
\end{subfigure} 
\begin{subfigure}{0.32\textwidth}
\includegraphics[trim=15 0 20 0, clip, width=1.0\linewidth]{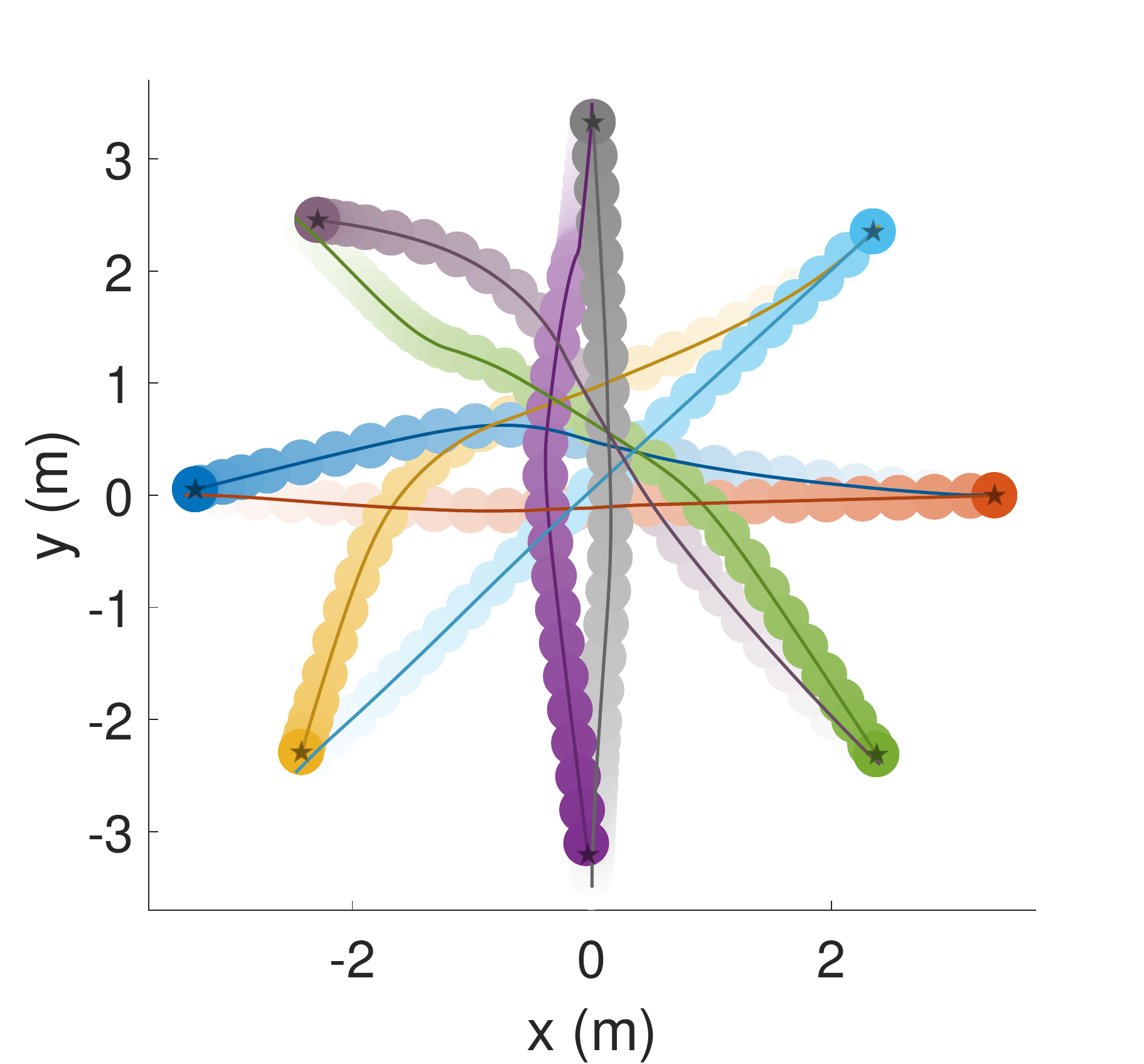}
\label{fig:orca_8_traj}
\end{subfigure}
\begin{subfigure}{0.32\textwidth}
\includegraphics[trim=15 5 20 15, clip, width=1.0\linewidth]{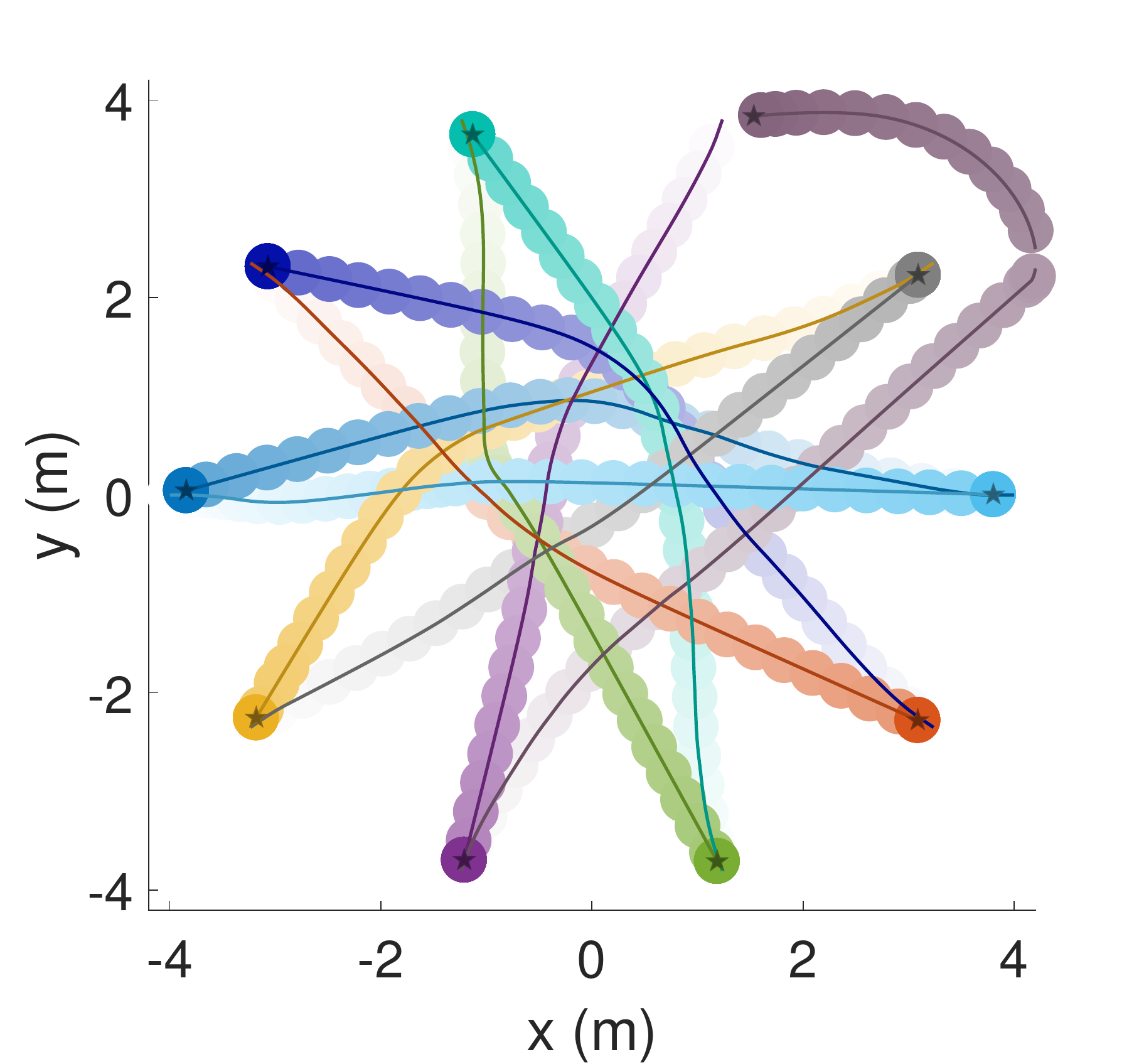}
\label{fig:orca_10_traj}
\end{subfigure}
\begin{subfigure}{0.32\textwidth}
\includegraphics[trim=15 5 20 15, clip, width=1.0\linewidth]{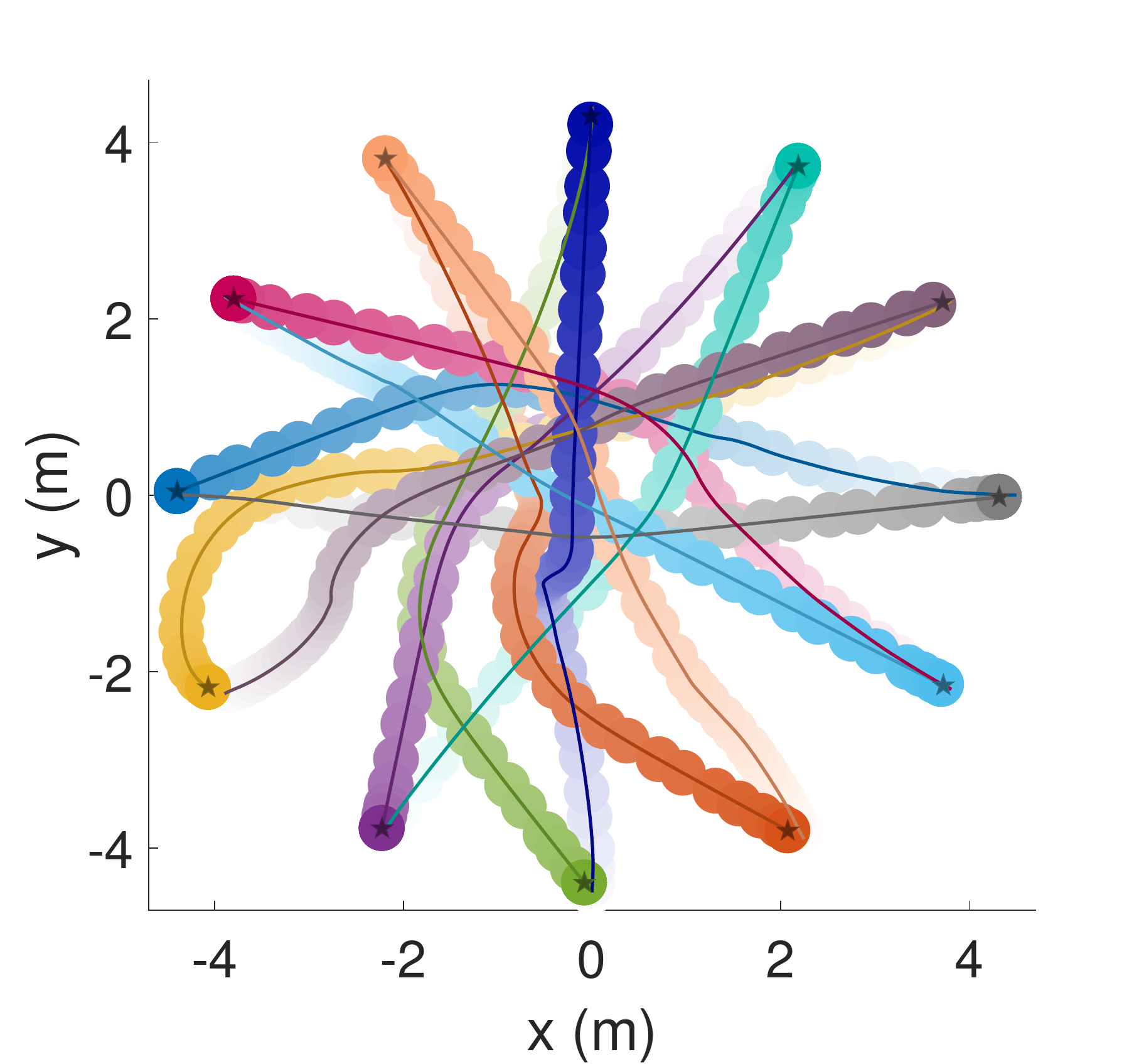}
\label{fig:orca_12_traj}
\end{subfigure}
\begin{subfigure}{0.32\textwidth}
\includegraphics[trim=15 5 20 15, clip, width=1.0\linewidth]{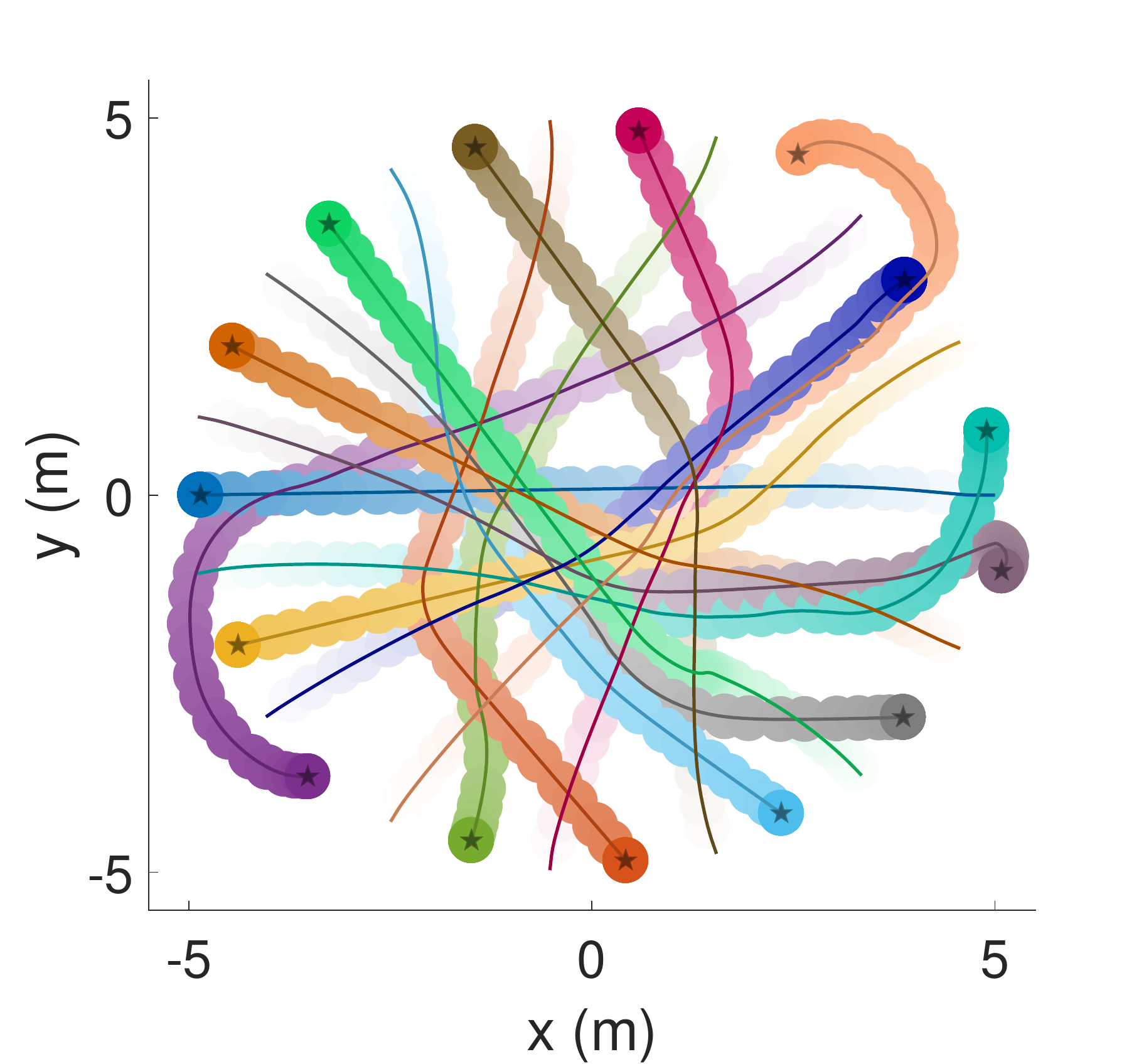}
\label{fig:orca_15_traj}
\end{subfigure}
\vspace*{-0.15in}
\caption{Trajectories of agents executing the NH-ORCA policy in circle scenarios with $4$, $6$, $8$, $10$, $12$ and $15$ agents respectively. We use different colors to represent trajectories of different robots and use the color transparency to indicate the temporal state along each trajectory.}
\label{fig:orca_circle}

\vspace*{0.25in}

\centering
\captionsetup[subfigure]{justification=centering}
\begin{subfigure}{0.32\textwidth}
\includegraphics[trim=15 5 20 15, clip, width=1.0\linewidth]{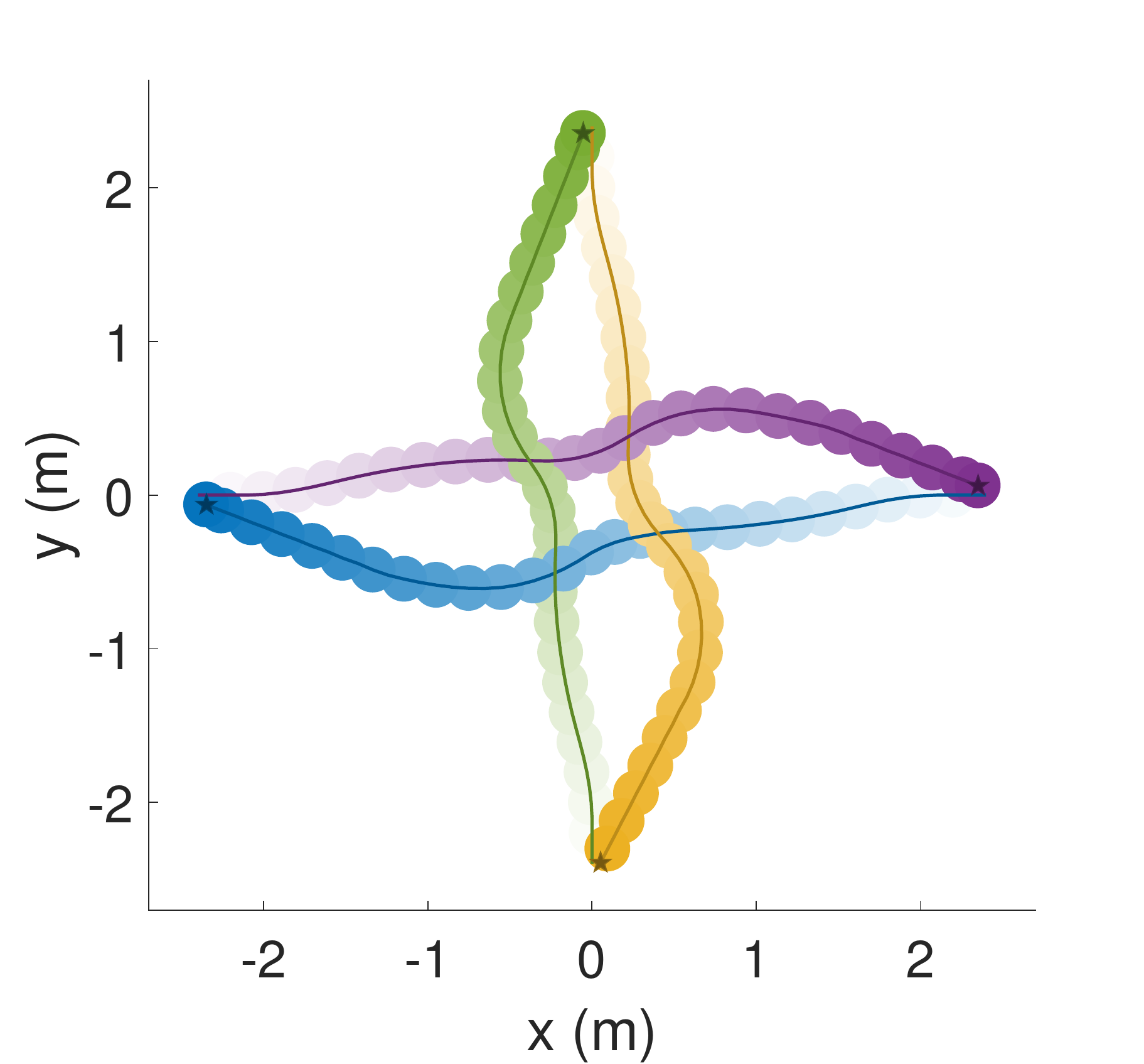}
\label{fig:rl_4_traj}
\end{subfigure}
\begin{subfigure}{0.32\textwidth}
\includegraphics[trim=15 5 20 15, clip, width=1.0\linewidth]{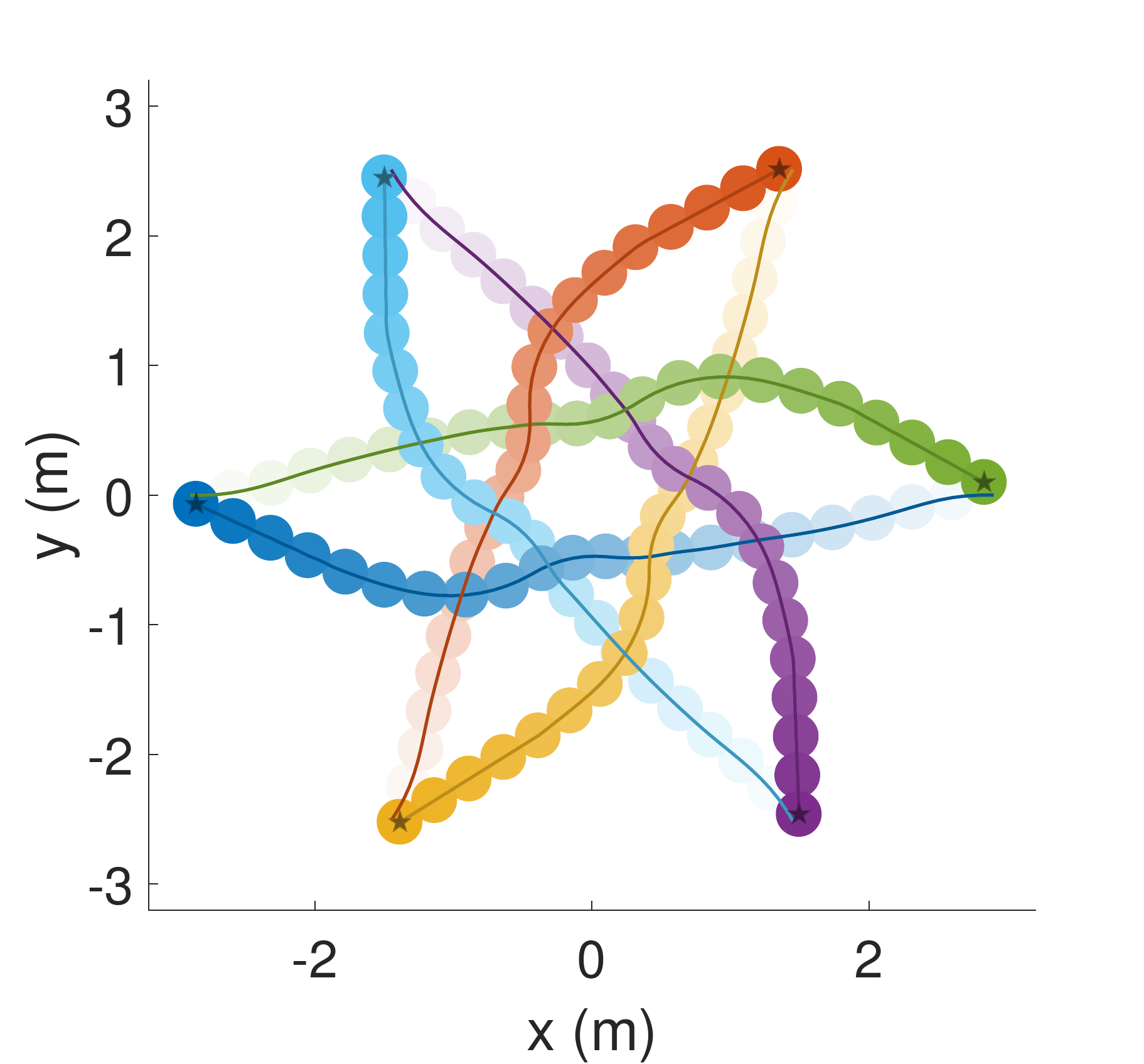}
\label{fig:rl_6_traj}
\end{subfigure} 
\begin{subfigure}{0.32\textwidth}
\includegraphics[trim=15 5 20 15, clip, width=1.0\linewidth]{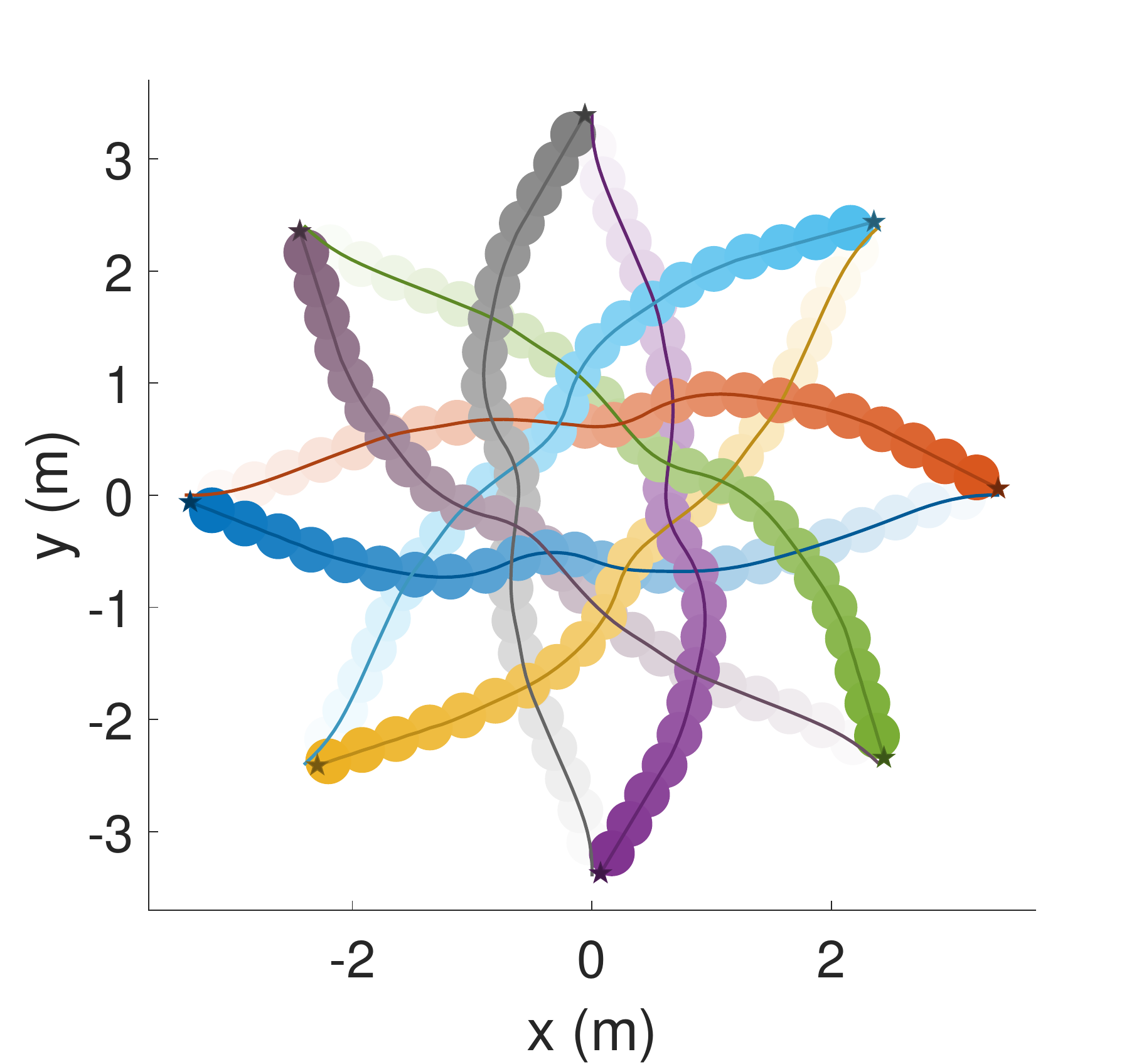}
\label{fig:rl_8_traj}
\end{subfigure}
\begin{subfigure}{0.32\textwidth}
\includegraphics[trim=15 5 20 15, clip, width=1.0\linewidth]{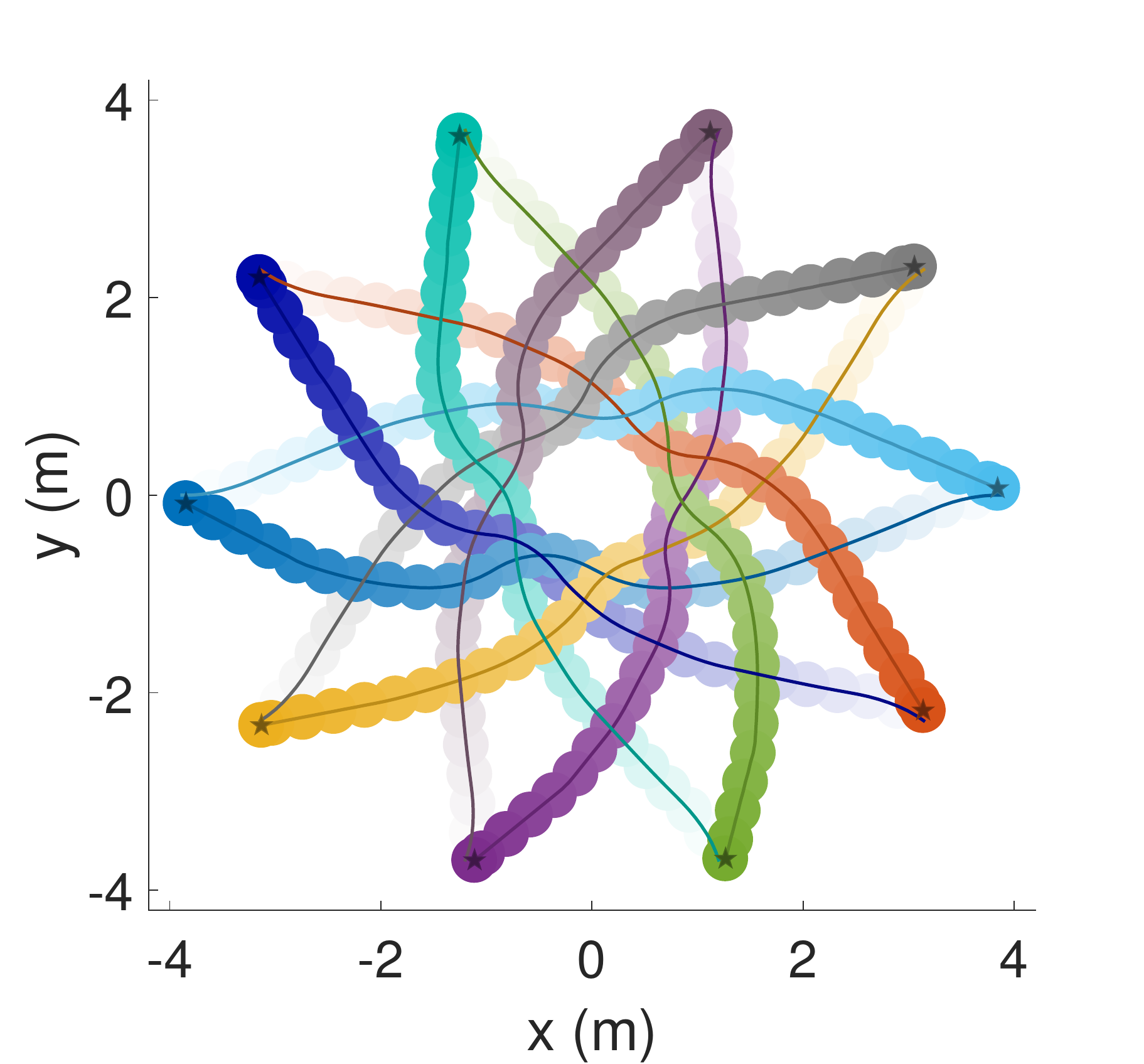}
\label{fig:rl_10_traj}
\end{subfigure}
\begin{subfigure}{0.32\textwidth}
\includegraphics[trim=15 5 20 15, clip, width=1.0\linewidth]{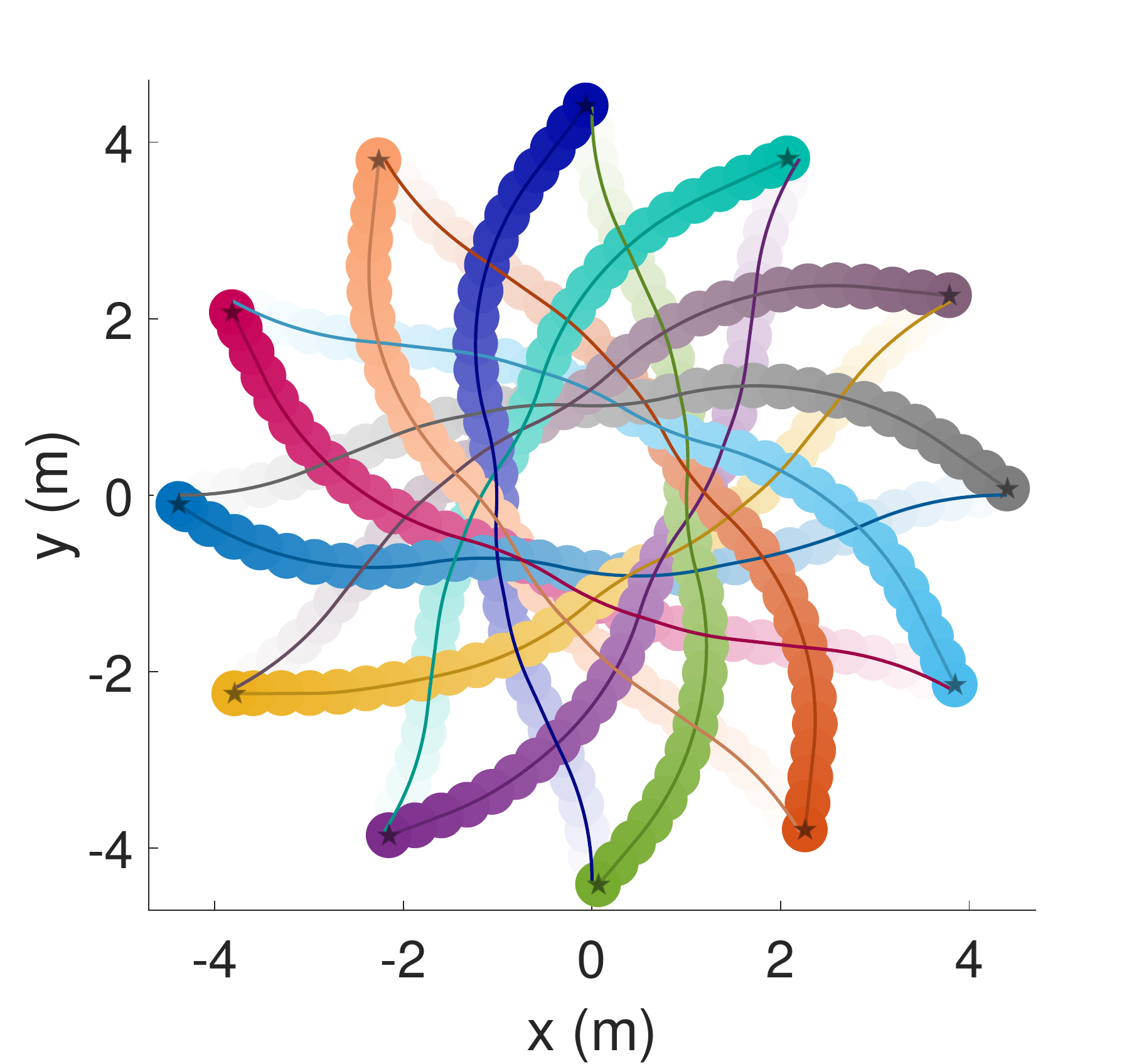}
\label{fig:rl_12_traj}
\end{subfigure}
\begin{subfigure}{0.32\textwidth}
\includegraphics[trim=15 5 20 15, clip, width=1.0\linewidth]{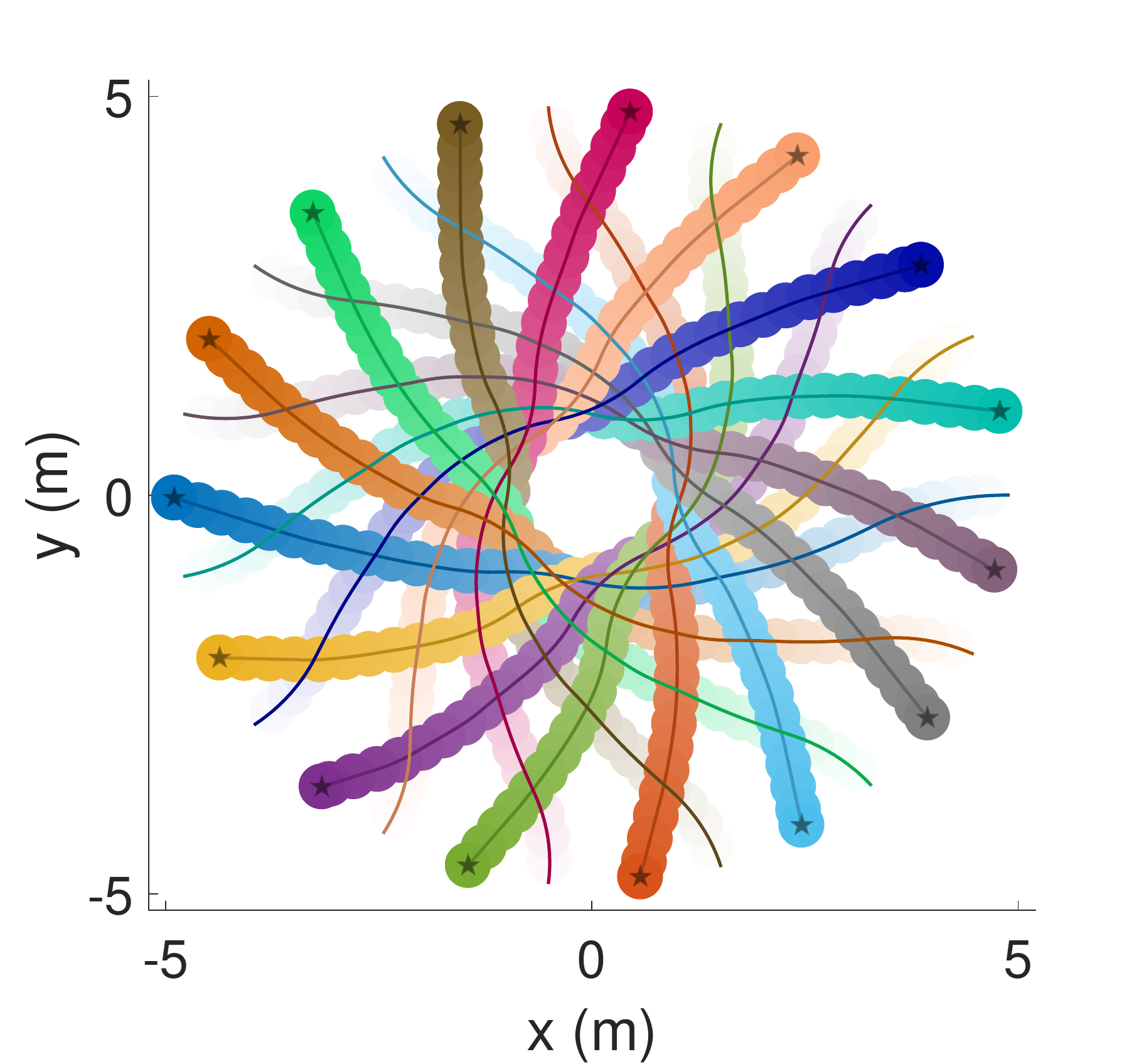}
\label{fig:rl_15_traj}
\end{subfigure}
\vspace*{-0.15in}
\caption{Trajectories of agents executing our Hybrid-RL policy in circle scenarios with $4$, $6$, $8$, $10$, $12$ and $15$ agents respectively. We use different colors to represent trajectories for different robots and use the color transparency to indicate the temporal state along each robot's trajectory sequence.}
\label{fig:rl_circle}
\end{figure*}

Furthermore, as indicated by the \textit{extra distance}, the reinforcement learned policies provide a comparable or even shorter traveled path than the NH-ORCA policy, as shown in the third row of \prettyref{tab:1circle}. In scenarios with $6$ and $10$ robots, our method produces slightly longer paths, because the robots need more space to decelerate from the high speed before stopping at goals or to rotate at a larger radius of curvature due to the higher angular velocity. 
Compared to the RL policy, our Hybrid-RL policy rarely needs to increase the extra distance for obtaining better performance in efficiency and safety. 
The quantitative difference between the learned policies and the NH-ORCA policy in \prettyref{tab:1circle} is also verified qualitatively by the difference of their trajectories illustrated in \prettyref{fig:orca_circle} and \prettyref{fig:rl_circle}. In particular, the NH-ORCA policy sacrifices the speed of each robot to avoid collisions and to generate shorter trajectories. By contrast, our reinforcement learned policies produce more consistent behavior for robots to resolve the congestion situation efficiently, which avoids unnecessary velocity reduction and thus obtains a higher average speed.

\begin{table*}
  \resizebox{\textwidth}{!}{  
  \begin{tabular}{l|l|l|l|l|l|l|l|l}
    \hline
    \multicolumn{1}{c|}{\multirow{2}{*}{Metrics}} & \multicolumn{1}{|c|}{\multirow{2}{*}{Method}} & \multicolumn{7}{c}{\#agents (radius of the scene)}\\ \cline{3-9}
    & & 4 (\SI{2.5}{m}) & 6 (\SI{3.0}{m}) & 8 (\SI{3.5}{m}) & 10 (\SI{4.0}{m}) & 12 (\SI{4.5}{m}) & 15 (\SI{5.0}{m}) & 20 (\SI{6.0}{m}) \\
    \hline
   	\multirow{4}{*}{Success Rate} & SL & 0.6  & 0.7167 & 0.6125 & 0.71 & 0.6333 & - & -  \\
      & NH-ORCA & \textbf{1.0}  & 0.9667 & 0.9250 & 0.8900 & 0.9000 & 0.8067 & 0.7800 \\
      & RL & \textbf{1.0}  & \textbf{1.0} & \textbf{1.0} & \textbf{1.0} & \textbf{1.0} & \textbf{1.0} & 0.9827 \\
      & Hybrid-RL & \textbf{1.0} & \textbf{1.0} & \textbf{1.0} & \textbf{1.0} & \textbf{1.0} & \textbf{1.0} & \textbf{1.0} \\ 
    \hline
    \multirow{4}{*}{Extra Time} & SL & 9.254 / 2.592  & 9.566 / 3.559 & 12.085 / 2.007 & 13.588 / 1.206 & 19.157 / 2.657 & - & -  \\
      & NH-ORCA & 0.622 / 0.080  & 0.773 / 0.207 & 1.067 / 0.215 & 0.877 /0.434 & 0.771 / 0.606 & 1.750 / 0.654 & 1.800 / 0.647 \\
      & RL & 0.323 / 0.025  & \textbf{0.408 / 0.009} & 0.510 / 0.005 & 0.631 / 0.011 & 0.619 / 0.020 & 0.490 / 0.046 & 0.778 / 0.032 \\
      & Hybrid-RL & \textbf{0.251 / 0.007} & \textbf{0.408 / 0.008} & \textbf{0.494/ 0.006} & \textbf{0.629 / 0.009} & \textbf{0.518 / 0.005} & \textbf{0.332 / 0.007} & \textbf{0.702 / 0.013} \\
      \hline
      \multirow{4}{*}{Extra Distance} & SL & 0.358 / 0.205  & 0.181 / 0.146 & 0.138 / 0.079 & 0.127 / 0.047 & 0.141 / 0.027 & - & -  \\
      & NH-ORCA & 0.017 / 0.004  & \textbf{0.025 / 0.005} & 0.041 / 0.034 & \textbf{0.034 / 0.009} & 0.062 / 0.024 & 0.049 / 0.019 & \textbf{0.056 / 0.018} \\
      & RL & 0.028 / 0.006  & 0.028 / 0.001 & 0.033 / 0.001 & 0.036 / 0.001 & \textbf{0.038 / 0.002} & 0.049 / 0.005 & 0.065 / 0.002 \\
      & Hybrid-RL & \textbf{0.013 / 0.001} & 0.028 / 0.001 & \textbf{0.031 / 0.001} & 0.036 / 0.001 & 0.039 / 0.001 & \textbf{0.033 / 0.001} & 0.058 / 0.001 \\
      \hline
      \multirow{4}{*}{Average Speed} & SL & 0.326 / 0.072 & 0.381 / 0.087 & 0.354 / 0.042 & 0.355 / 0.022 & 0.308 / 0.028 & - & -  \\
      & NH-ORCA & 0.859 / 0.012  & 0.867 / 0.026 & 0.839 / 0.032 & 0.876 / 0.045 & 0.875 / 0.054 & 0.820 / 0.052 & 0.831 / 0.042 \\
      & RL & 0.939 / 0.004  & \textbf{0.936 / 0.001} & 0.932 / 0.001 & 0.927 / 0.001 & 0.936 / 0.002 & 0.953 / 0.004 & 0.939 / 0.002 \\
      & Hybrid-RL & \textbf{0.952 / 0.001} & \textbf{0.936 / 0.001} & \textbf{0.934 / 0.001} & \textbf{0.927 / 0.001} & \textbf{0.946 / 0.001} & \textbf{0.968 / 0.001} & \textbf{0.945 / 0.001} \\
      \hline
 \end{tabular}
 }
\caption{Performance metrics (as mean/std) evaluated for different methods on the circle scenarios with varied scene sizes and different number of robots.}
\label{tab:1circle}
\end{table*}

\subsubsection{\textbf{Random scenarios}}
\begin{figure*}[htb] 
\captionsetup[subfigure]{justification=centering}
\centering
\begin{subfigure}{0.48\textwidth}
\includegraphics[trim=5 15 20 15, clip, height=6cm]{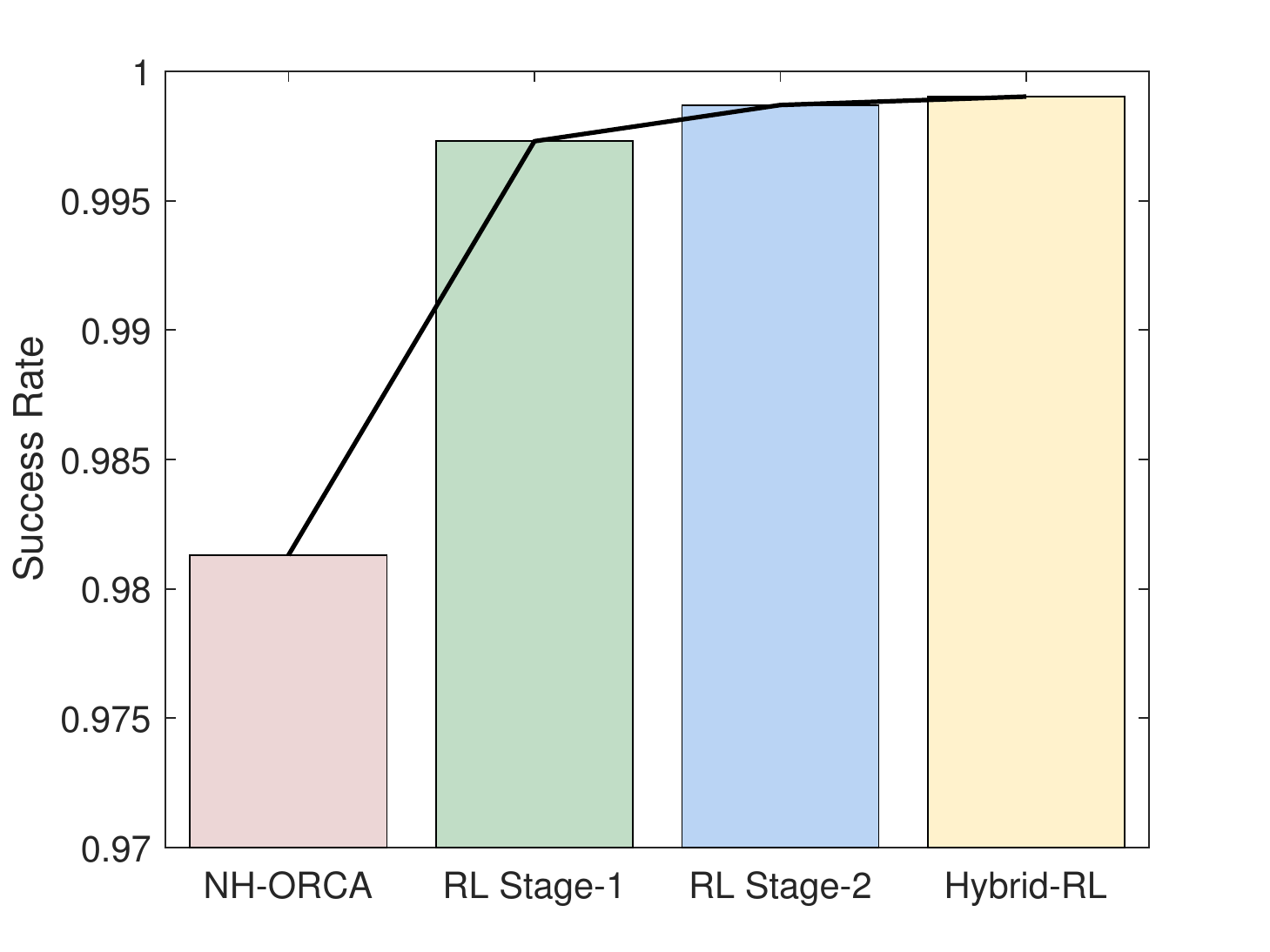}
\caption{Success rate}
\label{fig:2succ}
\end{subfigure}
\begin{subfigure}{0.48\textwidth}
\includegraphics[trim=15 15 20 15, clip, height=6cm]{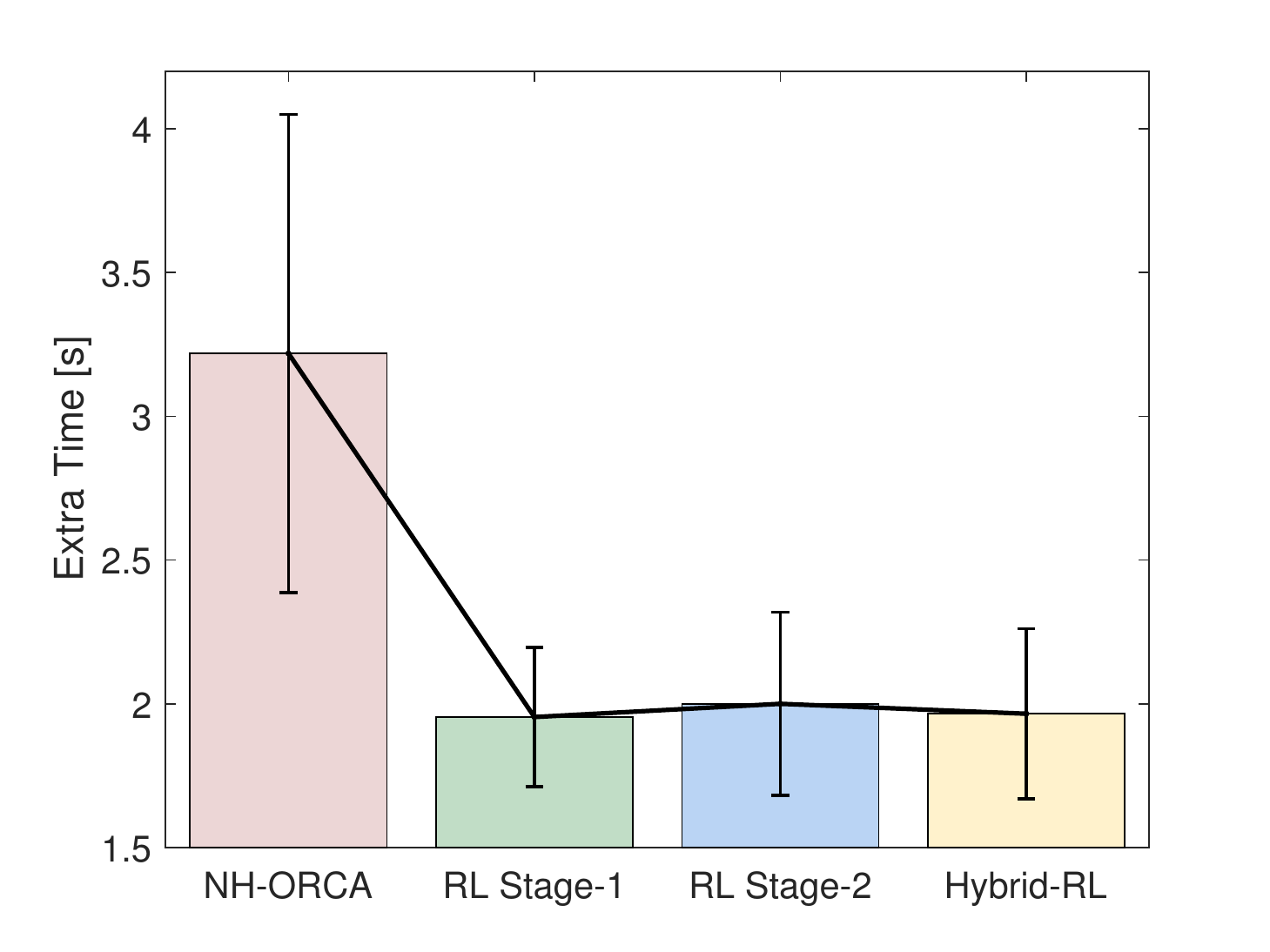}
\caption{Extra time}
\label{fig:2time}
\end{subfigure} 
\centering
\begin{subfigure}{0.48\textwidth}
\includegraphics[trim=15 15 20 15, clip, height=6cm]{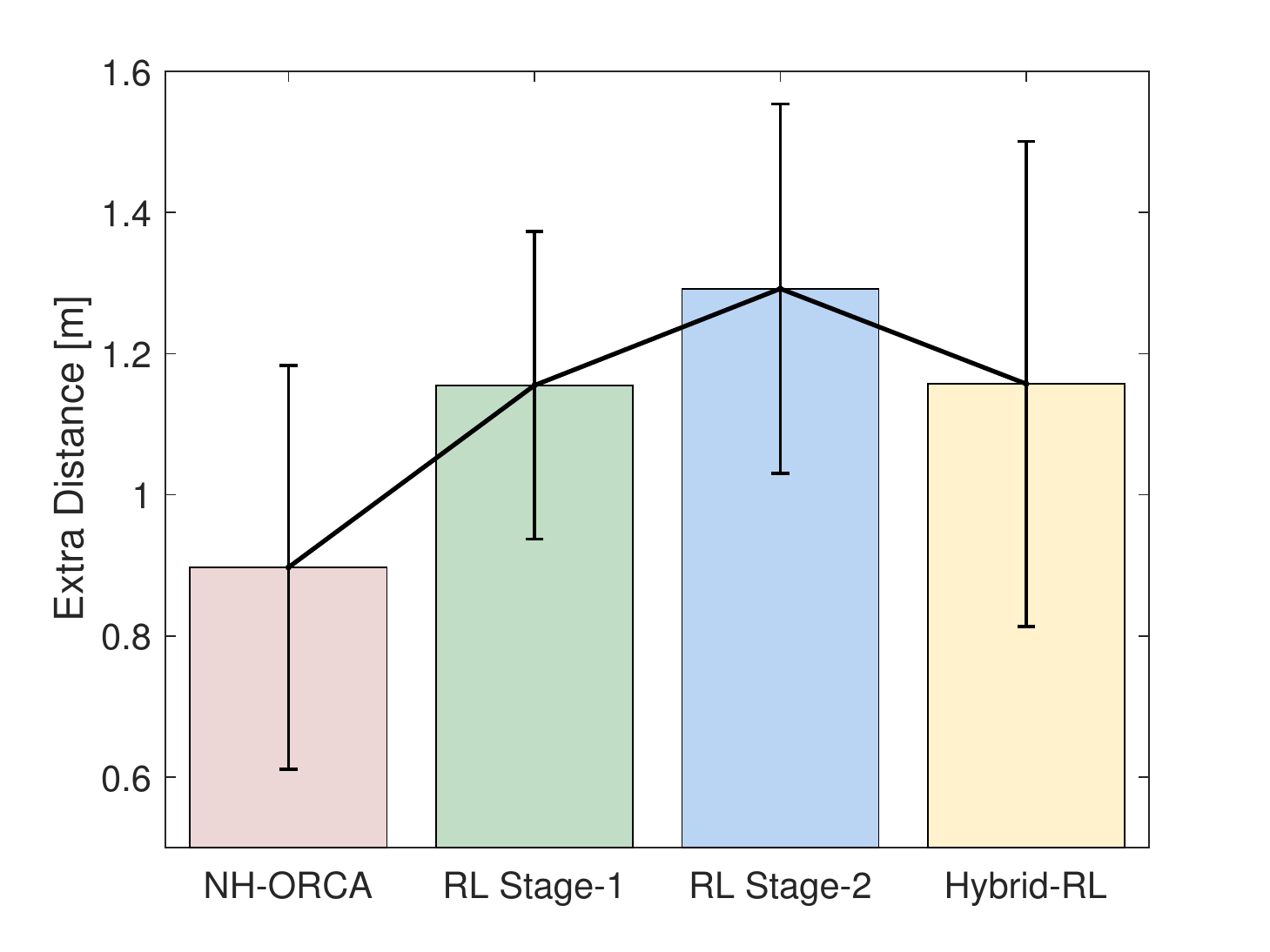}
\caption{Extra distance}
\label{fig:2dist}
\end{subfigure}
\begin{subfigure}{0.48\textwidth}
\includegraphics[trim=5 15 20 15, clip, height=6cm]{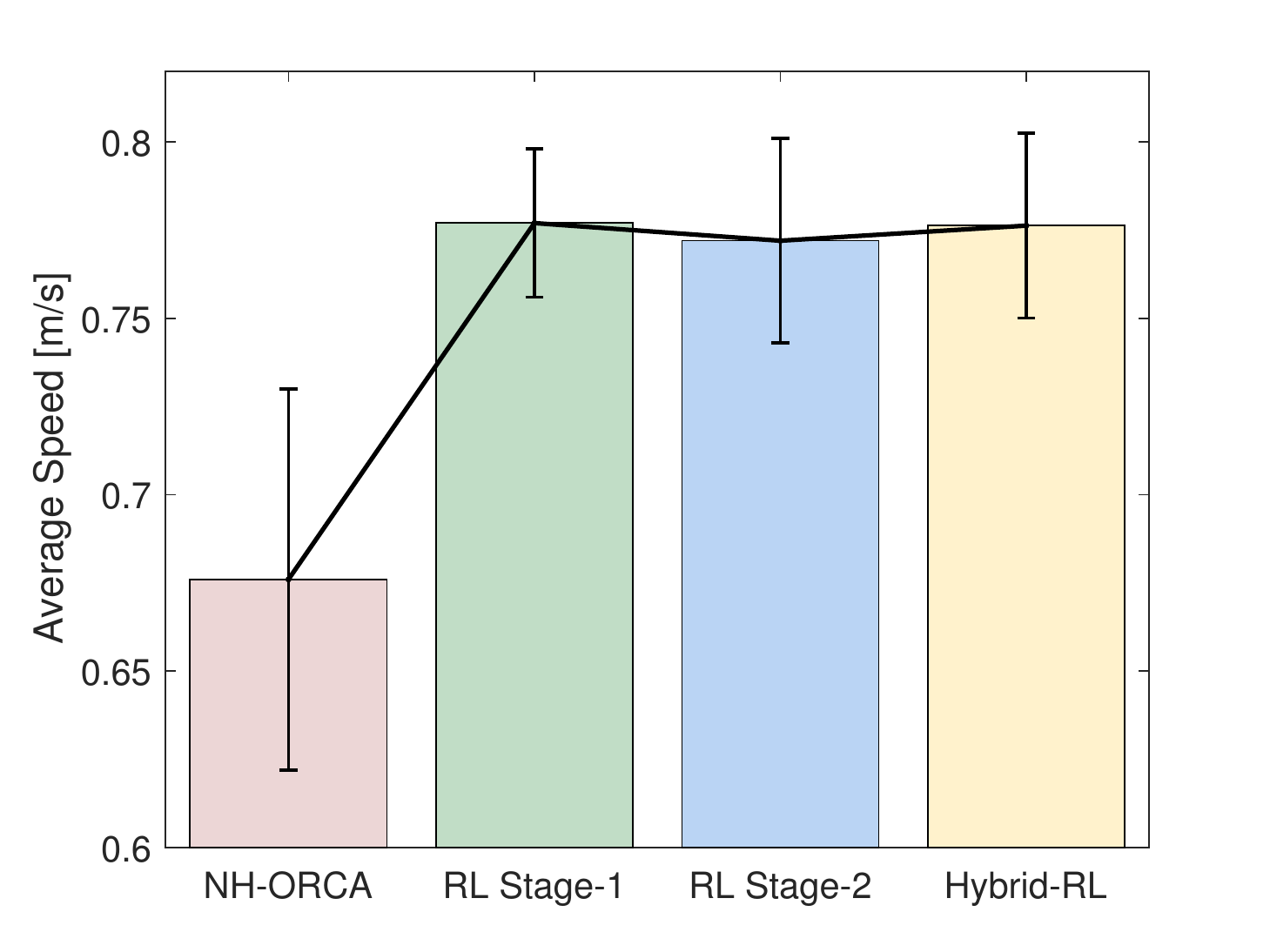}
\caption{Average speed}
\label{fig:2speed}
\end{subfigure}
\caption{Performance metrics evaluated for reinforcement learning based policies (RL Stage-1 policy, RL Stage-2 policy, and Hybrid-RL policy) and the NH-ORCA policy on random scenarios. }
\label{fig:2random}
\end{figure*}

Random scenarios are another type of scenarios frequently used to evaluate the performance of multi-robot collision avoidance. One example of the random scenarios is shown in the Scenario $7$ in \prettyref{fig:scene}, where each robot is assigned a random start point and a random goal position. To measure the performance of our method on random scenarios, we create $5$ different random scenarios with $15$ robots in each. For each random scenario, we repeat our evaluations $50$ times and the evaluation results are summarized in \prettyref{fig:2random}. We compare the Hybrid-RL policy with the RL Stage-1 policy, the RL Stage-2 policy, and the NH-ORCA policy. Note that the difficulty of the random scenarios is lower compared to the circle scenarios, since the random distribution of agents makes the traffic congestion unlikely to happen. 

As shown in \prettyref{fig:2succ}, we observe that all policies trained using deep reinforcement learning achieve success rates close to 100\%, which means they are safer than the NH-ORCA policy whose success rate is about 96\%. 
In addition, as shown in \prettyref{fig:2time}, robots using the learned policies reach their goals much faster than those using the NH-ORCA policy. 
Although the learned policies generate longer trajectories than the NH-ORCA policy (as shown in \prettyref{fig:2dist}), the higher average speed (\prettyref{fig:2speed}) and success rate indicate that our policies enable a robot to better cooperate with its nearby agents for higher navigation throughput. 
Similar to the circle scenarios above, the slightly longer path is due to robots' inevitable deceleration before stopping at goals or the larger radius of curvature taken to deal with the higher angular velocity.

As depicted in \prettyref{fig:2random}, the three reinforcement learning based policies have similar overall performance. The RL Stage-1 policy's performance is bit higher than the RL Stage-2 policy (e.g., according to the extra distance metric). This is probably because the RL Stage-1 policy is trained in similar scenarios as the test scenarios and thus it may be overfitted, while the RL Stage-2 policy is trained in miscellaneous scenarios for better generalization. 
Our Hybrid-RL policy further improves the RL Stage-2 policy and achieves similar overall performance as the RL Stage-1 policy, because the traditional control sub-policies improve the trajectory's optimality, and thus help to make a better balance between the optimality and generalization. 

\subsubsection{\textbf{Group scenarios}}
In order to evaluate the cooperation between robots, we test our learned policy on more challenging group scenarios, including group swap, group crossing, and group moving in the corridors. In the group swap scenario, two groups of robots, each with $5$ robots, are moving in opposite directions to swap their positions. As for the group crossing scenario, robots are separated in two groups with $4$ robots each, and their paths intersect in the center of the scenario.  

\begin{figure*}[!htb]
\centering
\includegraphics[trim=35 8 35 22, clip, width=0.9\linewidth]{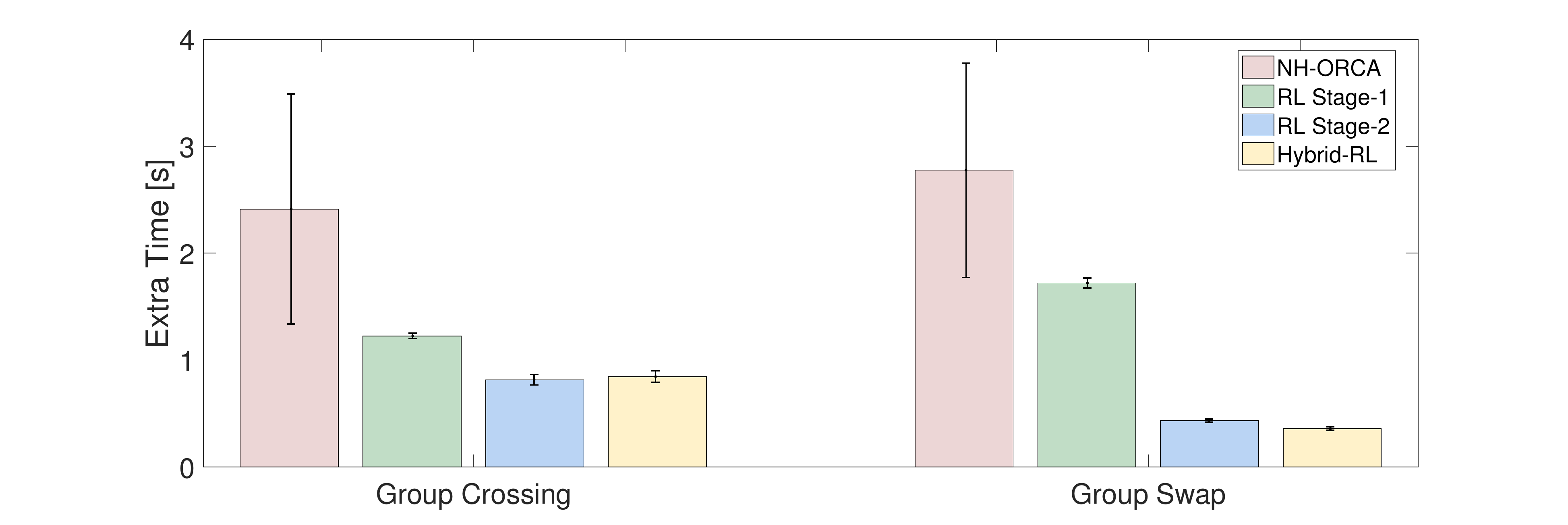}
\caption{Extra time $\bar{t}_e$ of our policies (RL Stage-1 policy and RL Stage-2 policy) and the NH-ORCA policy on two group scenarios.}
\label{fig:3group}
\end{figure*}

\begin{figure*}[!htb]
\captionsetup[subfigure]{justification=centering}
\centering
\begin{subfigure}{0.24\textwidth}
\includegraphics[height=4.5cm]{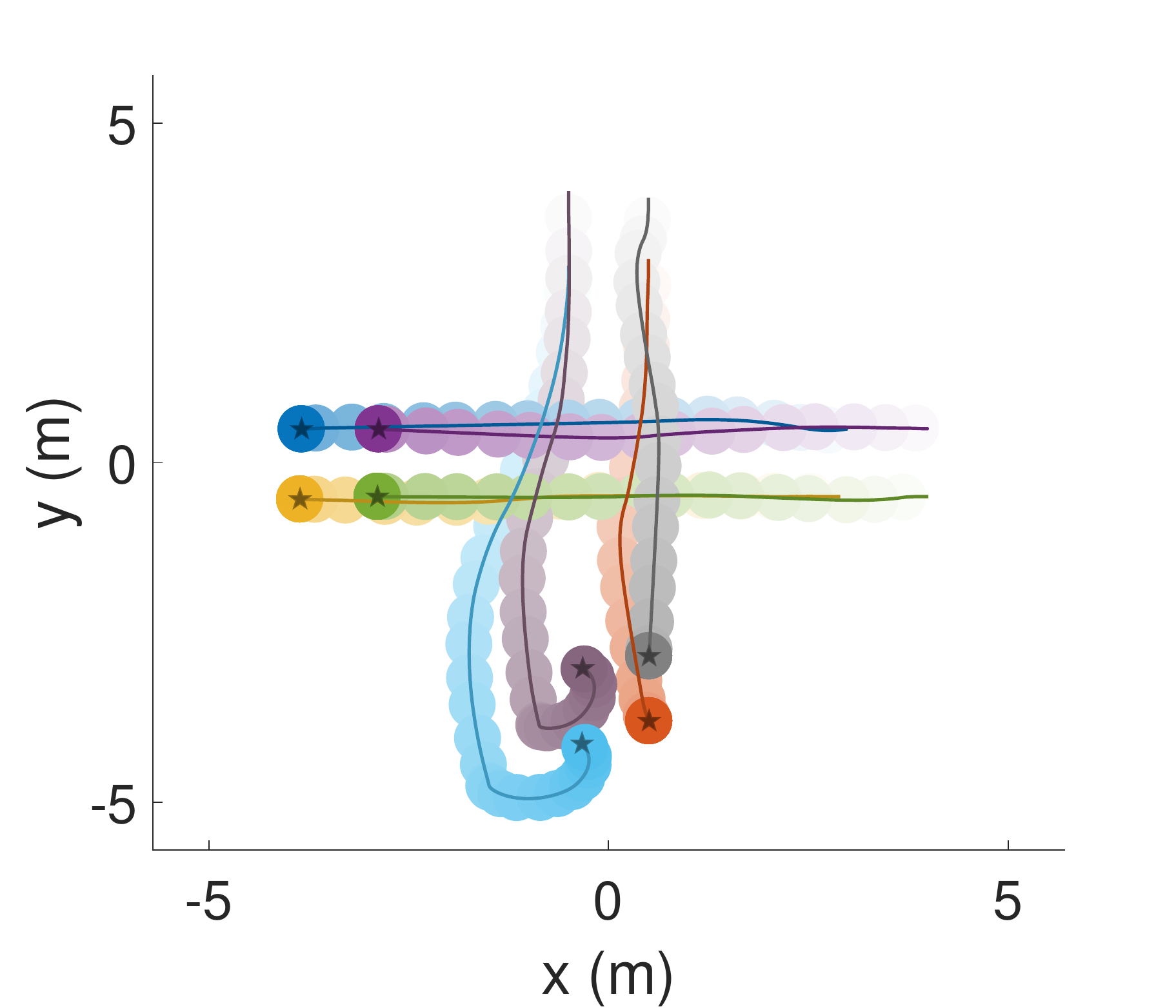}
\caption{NH-ORCA policy}
\label{fig:orca_cross_traj}
\end{subfigure}
\begin{subfigure}{0.24\textwidth}
\includegraphics[height=4.5cm]{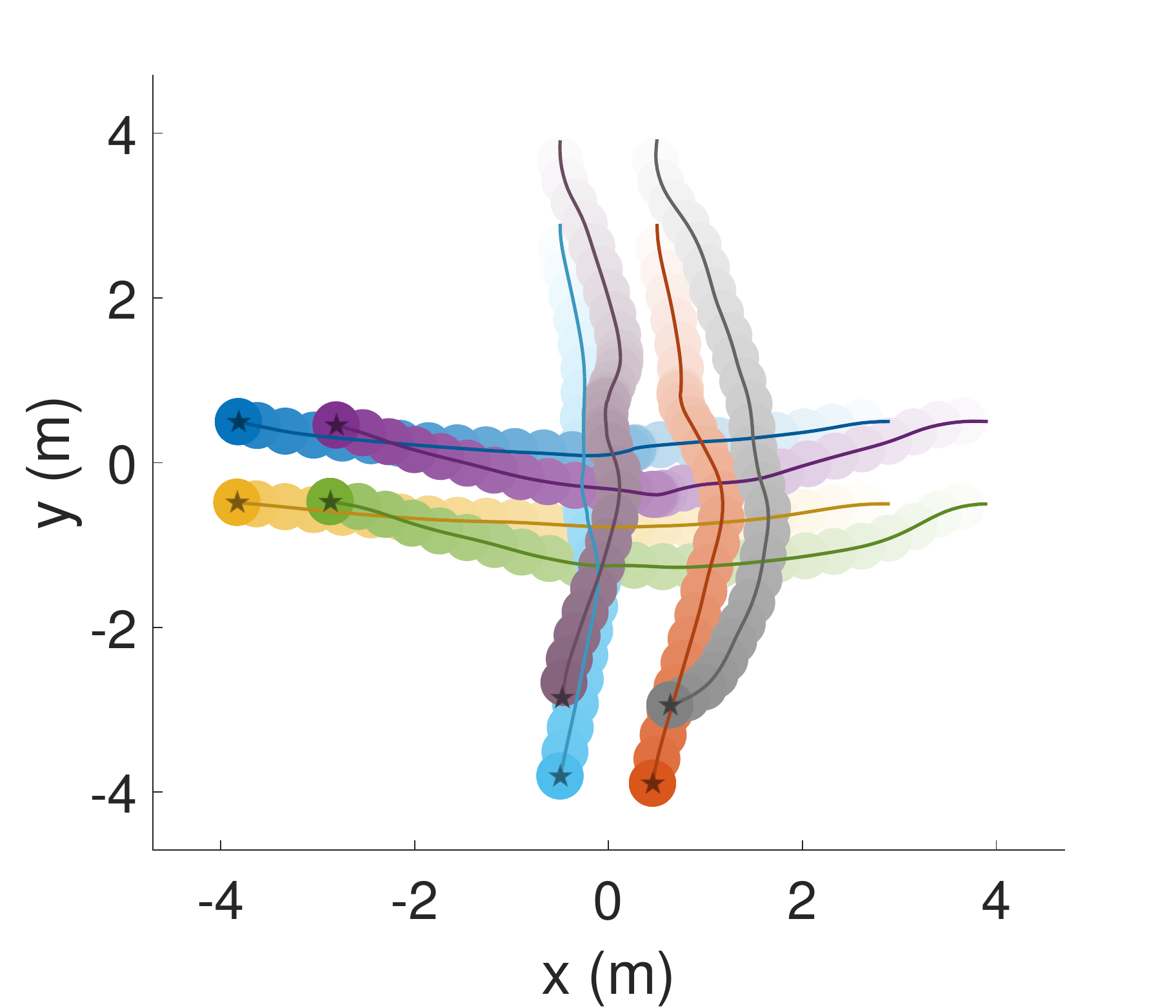}
\caption{RL Stage-1 policy}
\label{fig:stage1_cross_traj}
\end{subfigure} 
\begin{subfigure}{0.24\textwidth}
\includegraphics[height=4.5cm]{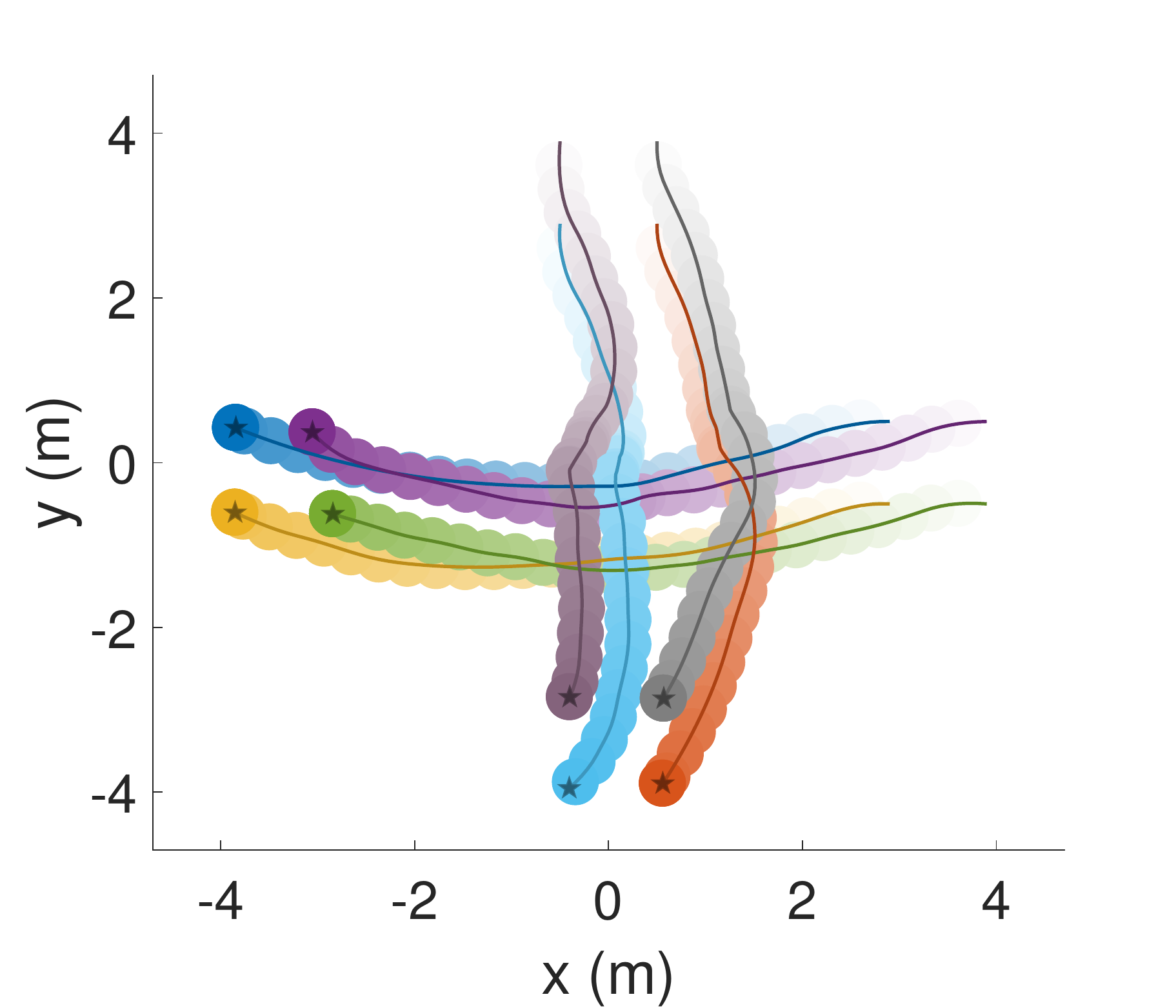}
\caption{RL Stage-2 policy}
\label{fig:stage2_cross_traj}
\end{subfigure}
\begin{subfigure}{0.24\textwidth}
\includegraphics[height=4.5cm]{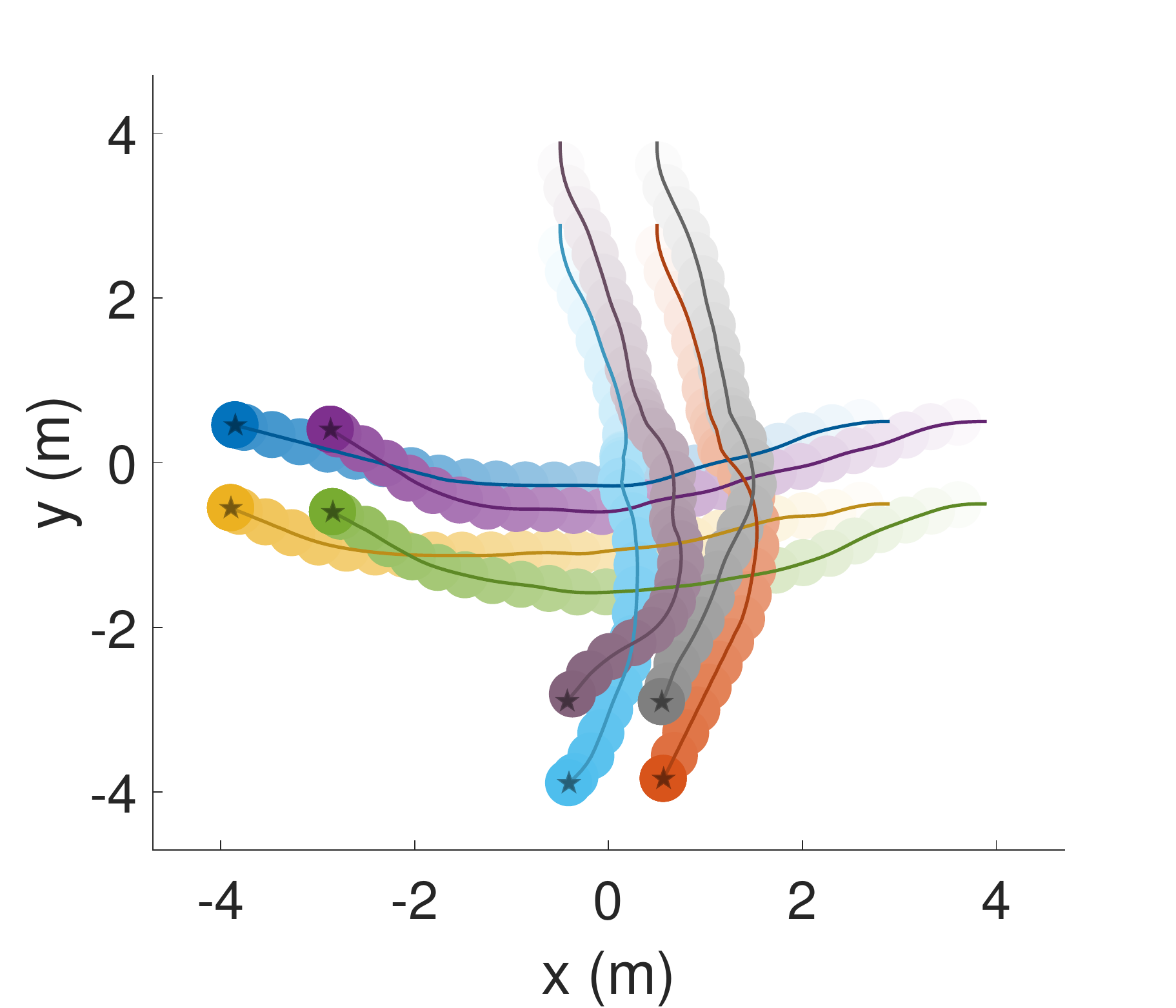}
\caption{Hybrid-RL policy}
\label{fig:hybrid_cross_traj}
\end{subfigure}
\caption{Comparison of trajectories generated by different policies in group crossing scenarios. We use different colors to distinguish trajectories of different agents and use the color transparency to indicate the timing of a trajectory sequence. }
\label{fig:cross_traj}
\end{figure*}

\begin{figure*}[!htb]
\captionsetup[subfigure]{justification=centering}
\centering
\begin{subfigure}{0.24\textwidth}
\includegraphics[height=4.5cm]{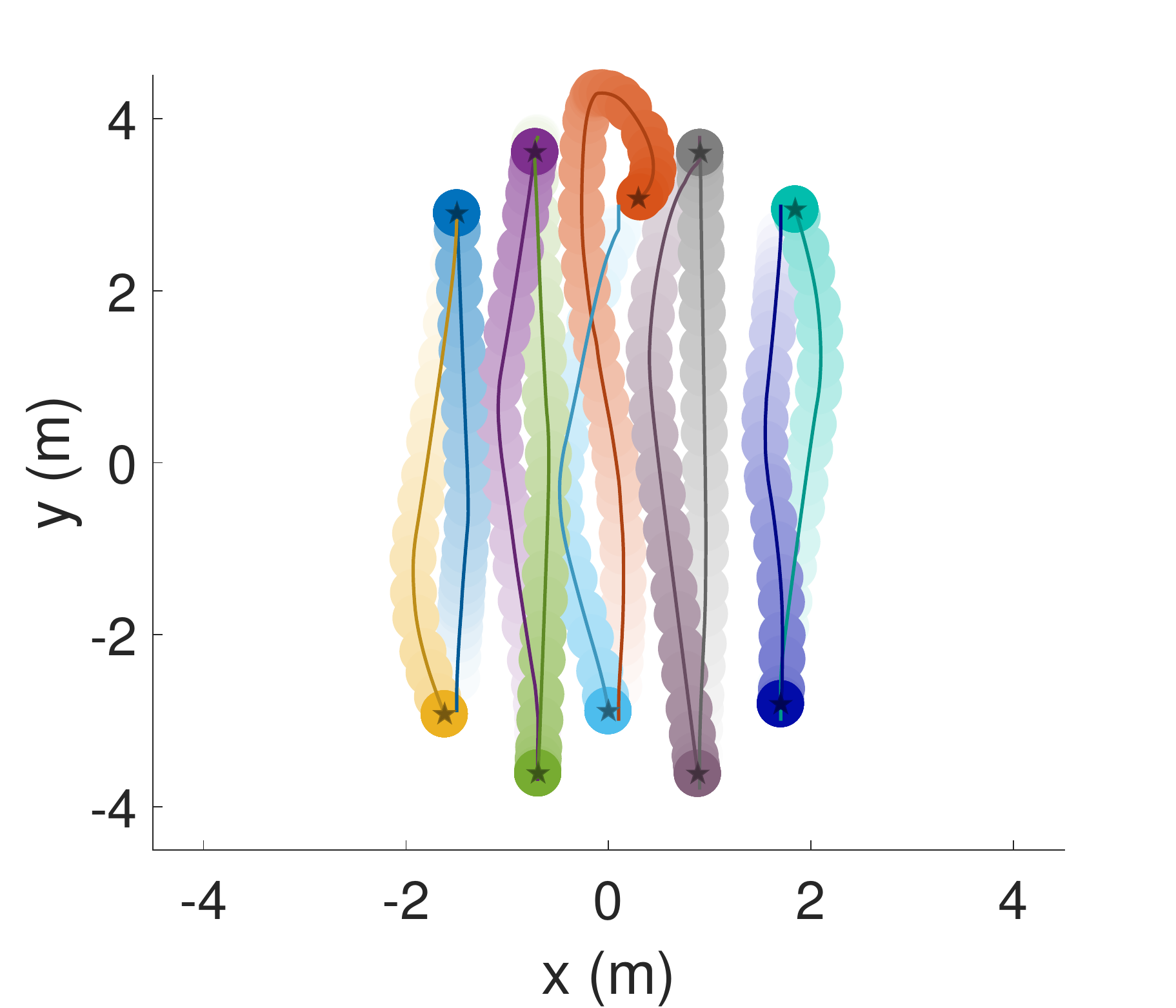}
\caption{NH-ORCA policy}
\label{fig:orca_swap_traj}
\end{subfigure}
\begin{subfigure}{0.24\textwidth}
\includegraphics[height=4.5cm]{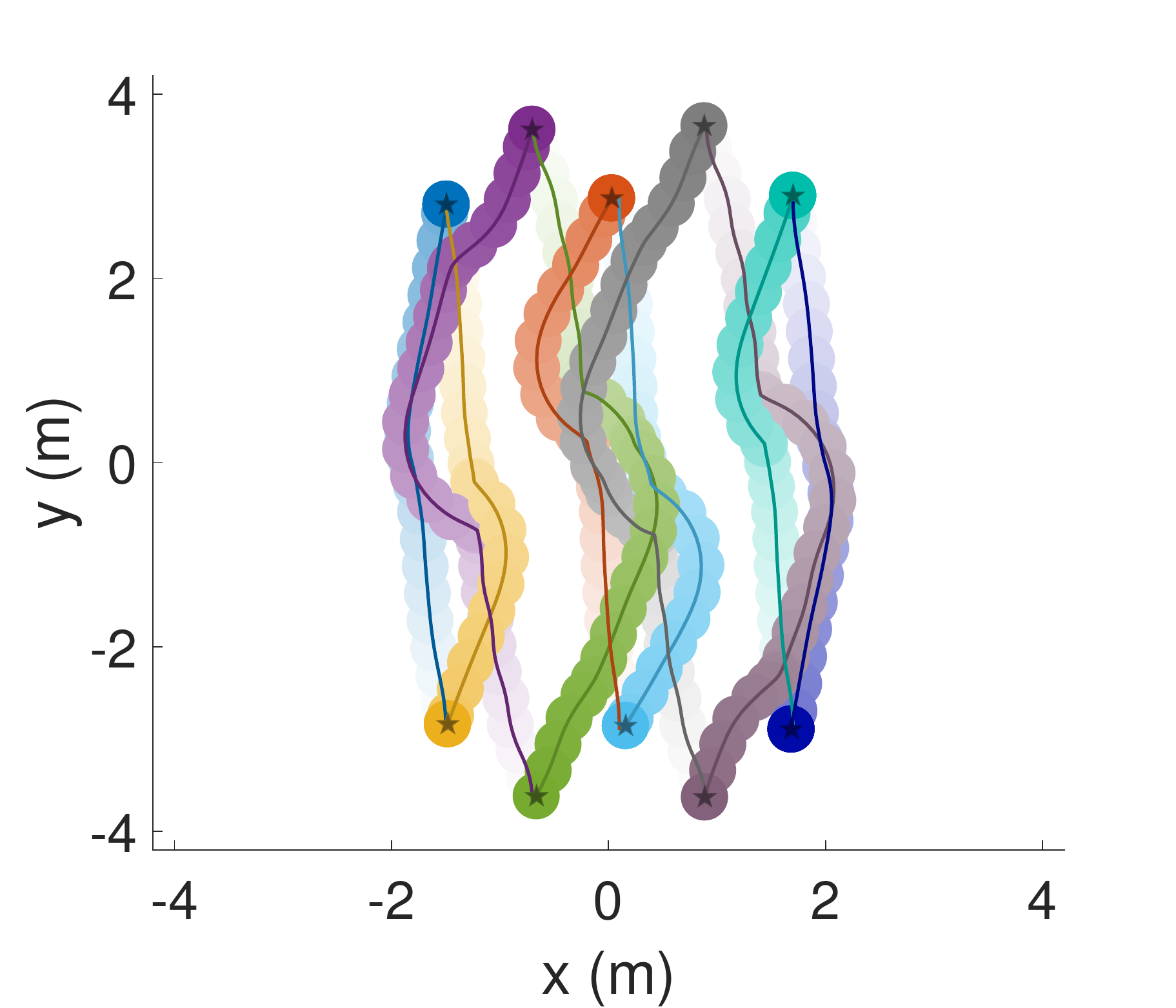}
\caption{RL Stage-1 policy}
\label{fig:stage1_swap_traj}
\end{subfigure} 
\begin{subfigure}{0.24\textwidth}
\includegraphics[height=4.5cm]{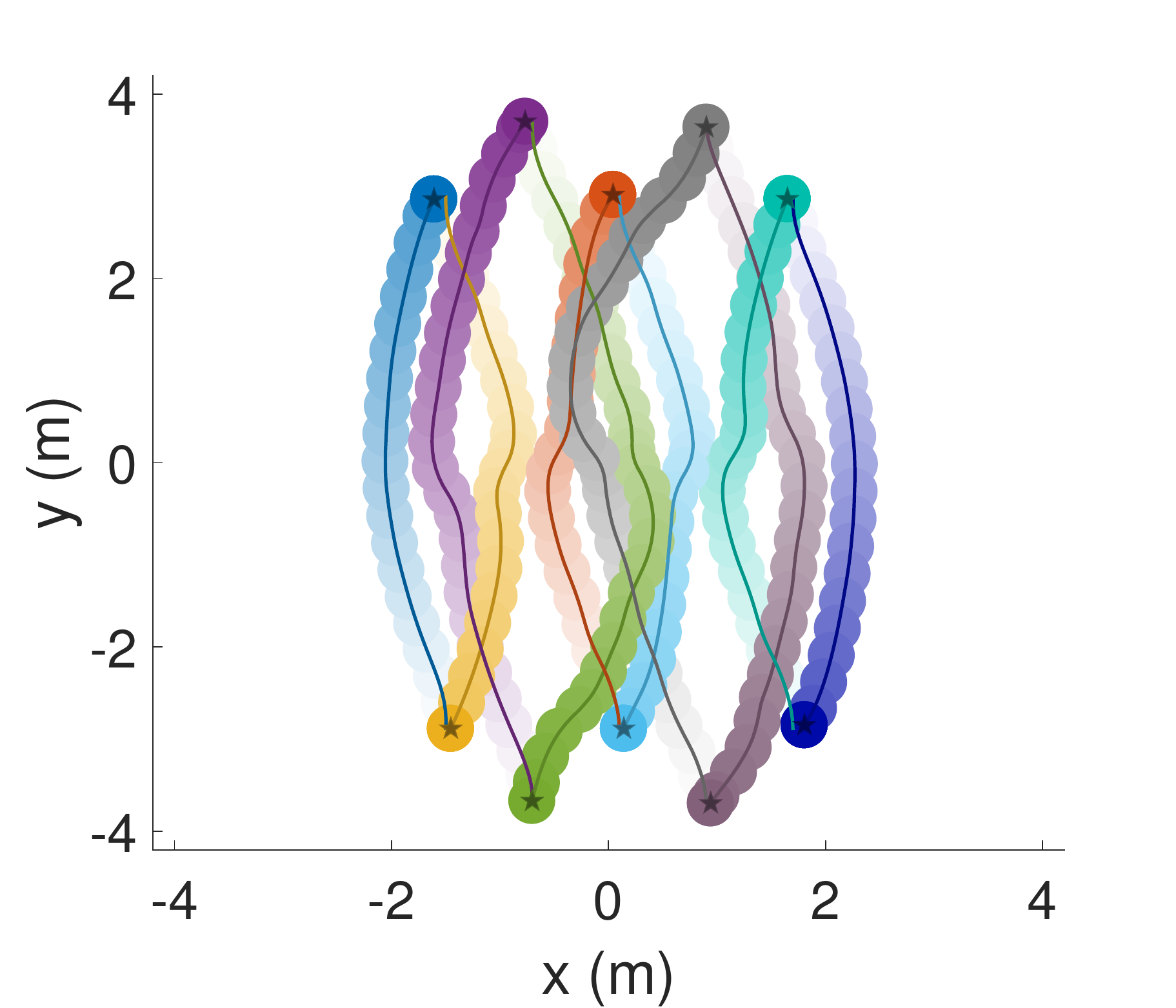}
\caption{RL Stage-2 policy}
\label{fig:stage2_swap_traj}
\end{subfigure}
\begin{subfigure}{0.24\textwidth}
\includegraphics[height=4.5cm]{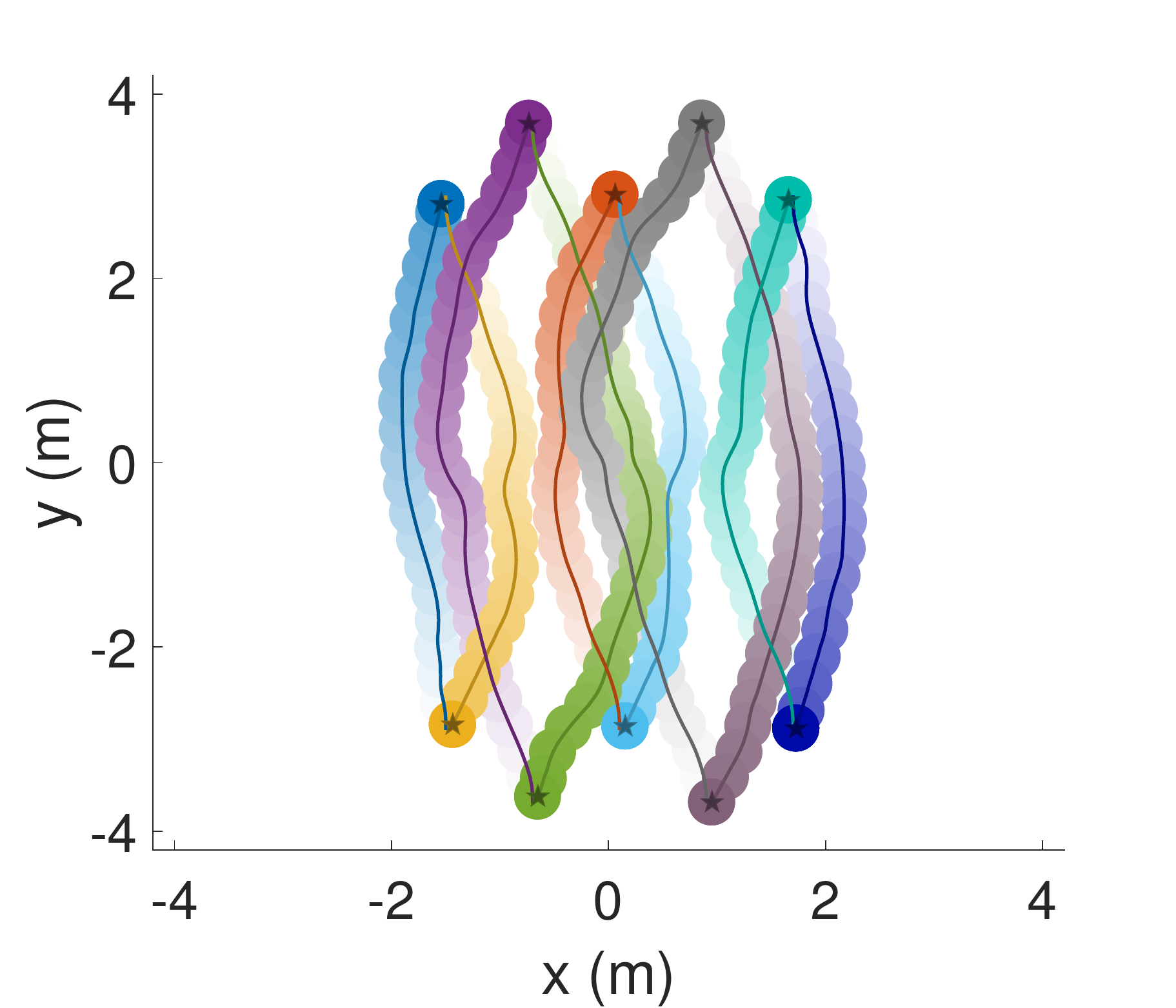}
\caption{Hybrid-RL policy}
\label{fig:hybrid_swap_traj}
\end{subfigure}
\caption{Comparison of trajectories generated by different policies in group swap scenarios. We use different colors to distinguish trajectories of different agents and use the color transparency to indicate the timing of a trajectory sequence. }
\label{fig:swap_traj}
\end{figure*}

We compare our method with the NH-ORCA policy on these two cases by measuring the average extra time $\bar{t}_e$ with $50$ trials. As summarized in \prettyref{fig:3group}, the reinforcement learned policies perform much better than the NH-ORCA policy on both benchmarks. The robots take shorter time to reach goals when using the learned policies, which indicates that the learned policies produce more cooperative behaviors than the NH-ORCA policy. Among the three learned policies, the Hybrid-RL policy provides the best performance. Besides, we also illustrate the trajectories of different methods in both scenarios in \prettyref{fig:cross_traj} and \prettyref{fig:swap_traj}. From these trajectories, we observe that the NH-ORCA policy generates unnecessary circuitous trajectory. The RL Stage-1 policy tends to generate stopping behaviors when one agent is blocked by another, while the RL Stage-2 policy and Hybrid-RL policy provide smooth and collision-free trajectories. 

\subsubsection{\textbf{Corridor scenario}}
\begin{figure*}[!htb] 
\captionsetup[subfigure]{justification=centering}
\centering
\begin{subfigure}{0.48\textwidth}
\includegraphics[width=1.0\linewidth]{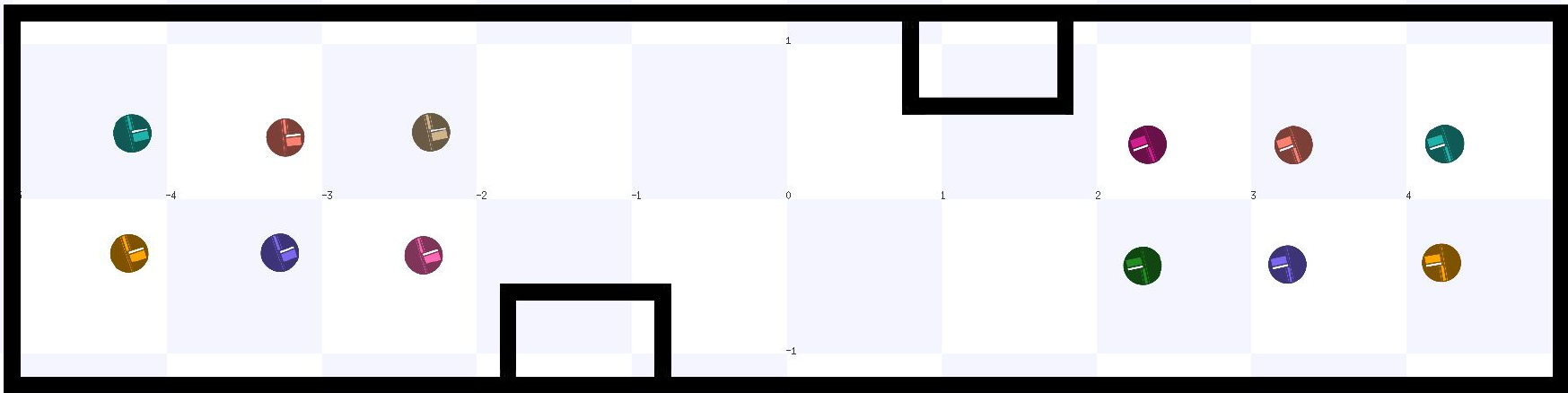}
\caption{Corridor scenario}
\label{fig:3corridor_scene}
\end{subfigure}
\begin{subfigure}{0.48\textwidth}
\includegraphics[width=1.0\linewidth]{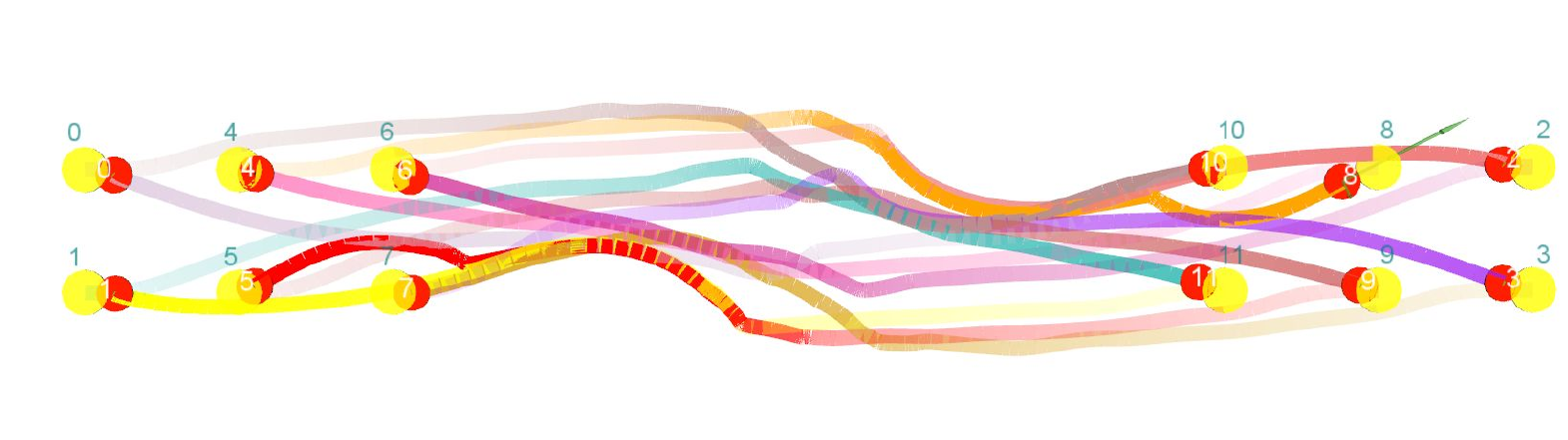}
\caption{Robot trajectories}
\label{fig:3corridor_path}
\end{subfigure}
\caption{Two groups of robots moving in a corridor with obstacles. (a) shows the corridor scenario and robots' initial positions. (b) shows trajectories generated by the Hybrid-RL policy, where the initial points are in red and the goal points are in yellow. We use different colors to distinguish trajectories of different agents and use the color transparency to indicate the timing of a trajectory sequence. Please also refer to the video for the comparison between the RL policy and the NH-ORCA policy in this scenario.}
\label{fig:3corridor}
\end{figure*}

We further evaluate the four policies in a corridor scene, where two groups exchange their positions via a narrow corridor connecting two open regions, as shown in \prettyref{fig:3corridor_scene}. 

Note that this is a benchmark with static obstacles, and we have to use different pipelines when using the NH-ORCA policy and the learned policies. In particular, the NH-ORCA requires prior knowledge about the global environment (usually a map from SLAM) to identify the static obstacles and then depends on global planners to guide robots navigating in the complex environment. The global planners do not consider the other moving agents, and thus such guidance is sub-optimal and significantly reduces the efficiency and the safety of the navigation. For the learned policies, we are using them in a map-less manner, i.e., each robot only knows its goal but without any knowledge about the global map. The goal information can be provided by global localization system such as the GPS or UWB (Ultra Wide-Band) in practice. A robot will use the vector pointing from its current position to the goal position as a guidance toward the target. Such guidance may fail in buildings with complex topologies but works well for common scenarios with moving obstacles. 

Navigation in the corridor scenario is a challenging task, and only the Hybrid-RL policy and the RL Stage-2 policy can complete it. The trajectories generated by the Hybrid-RL policy are illustrated in \prettyref{fig:3corridor_path}. The failure of the RL Stage-1 policy indicates that the co-training on a variety of scenarios is beneficial for the robustness across different situations. The NH-ORCA policy fails with many collisions.

\subsubsection{\textbf{Summary}}
In this section, we have validated the superiority of our learned policies in multiple scenarios by comparing with the state-of-the-art rule-based reaction controller NH-ORCA. We also show that the hybrid control framework can effectively improve the learned policy's performance in many scenarios. 

\subsection{Robustness evaluation}
\label{sec:robust}
Besides the generalization and the quality of the policy, another main concern about deep reinforcement learning is whether the learned policy is stable and robust to model uncertainty and input noises.
In this section, we design multiple experiments to verify the robustness of our learned policy.

\subsubsection{\textbf{Performance metrics}}
In order to quantify the robustness of our policy with respect to the workspace and robot setup, we design the following performance metrics. 
\renewcommand{\labelitemi}{\textbullet}
\begin{itemize}
\item \textit{Failure rate} is the ratio of robots that cannot reach their goals within a certain time limit.
\item \textit{Collision rate} is the ratio of robots that cannot reach their goals due to colliding with other robots or static obstacles during the navigation.
\item \textit{Stuck rate} is the ratio of robots that cannot reach their goals because they stuck in some position but without any collisions. 
\end{itemize}
Note that the failure rate is equal to the collision rate plus the stuck rate.
Similar to the experiments in \prettyref{sec:efficiency}, when computing each metric for a given method, we repeat the experiment $50$ times and report the mean value.

\subsubsection{\textbf{Different agent densities}}
We first evaluate the performance of different policies in scenarios with varied agent densities. In particular, we design five scenarios which is a $7$-meter by $7$-meter square region and allocate $20$, $30$, $40$, $50$, and $60$ agents, respectively. The agent density for these scenarios are about $0.4$, $0.6$, $0.8$, $1$, $1.2$ robots per square meter, respectively. Note that the agent densities of these five scenarios are higher than densities of the training and test scenarios in previous experiments. For instance, the agent density in circle scenarios in \prettyref{tab:1circle} is only $0.2$ agents \SI{}{/m^2}. As a result, this experiment is used to challenge the learned policy's robustness when handling the high agent density.

\begin{figure}[!htb] 
\captionsetup[subfigure]{justification=centering}
\centering
\begin{subfigure}{1\linewidth}
\includegraphics[trim=15 0 20 0, clip, width=1.0\linewidth]{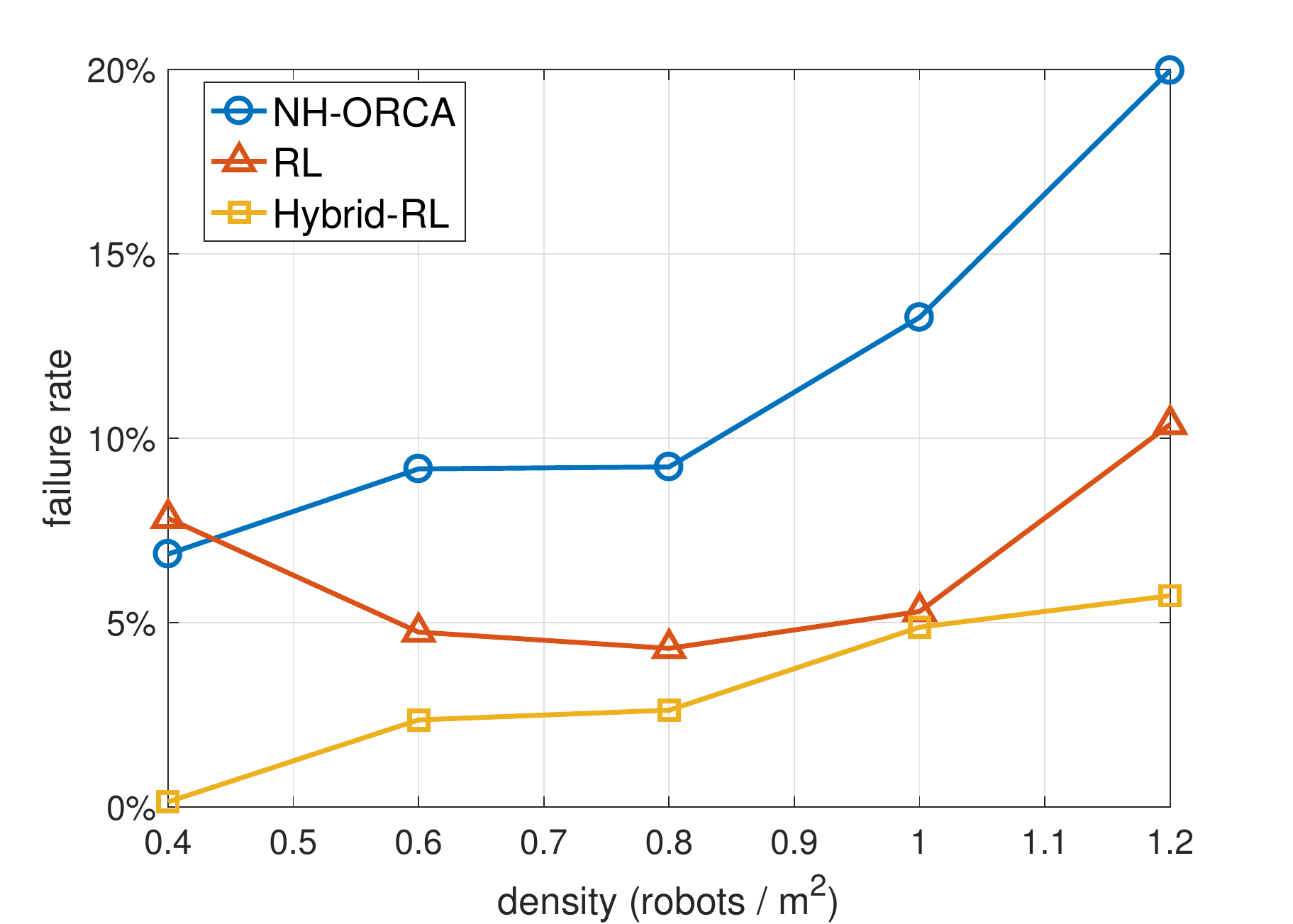}
\caption{Failure rate w.r.t. agent density}
\label{fig:density_failure}
\end{subfigure}
\begin{subfigure}{1\linewidth}
\includegraphics[trim=15 0 20 0, clip, width=1.0\linewidth]{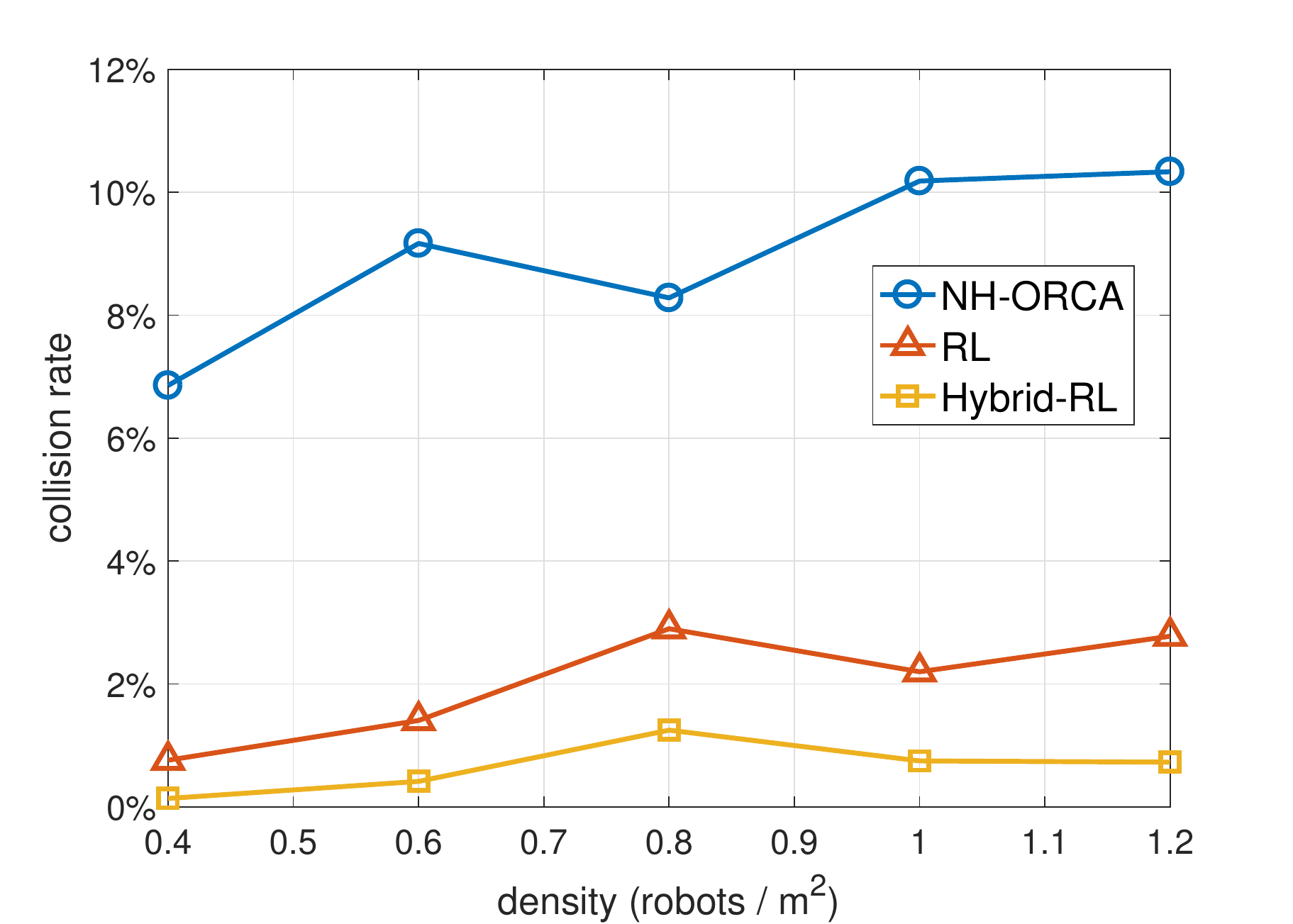}
\caption{Collision rate w.r.t. agent density}
\label{fig:density_collision}
\end{subfigure} 
\begin{subfigure}{1\linewidth}
\includegraphics[trim=15 0 20 0, clip, width=1.0\linewidth]{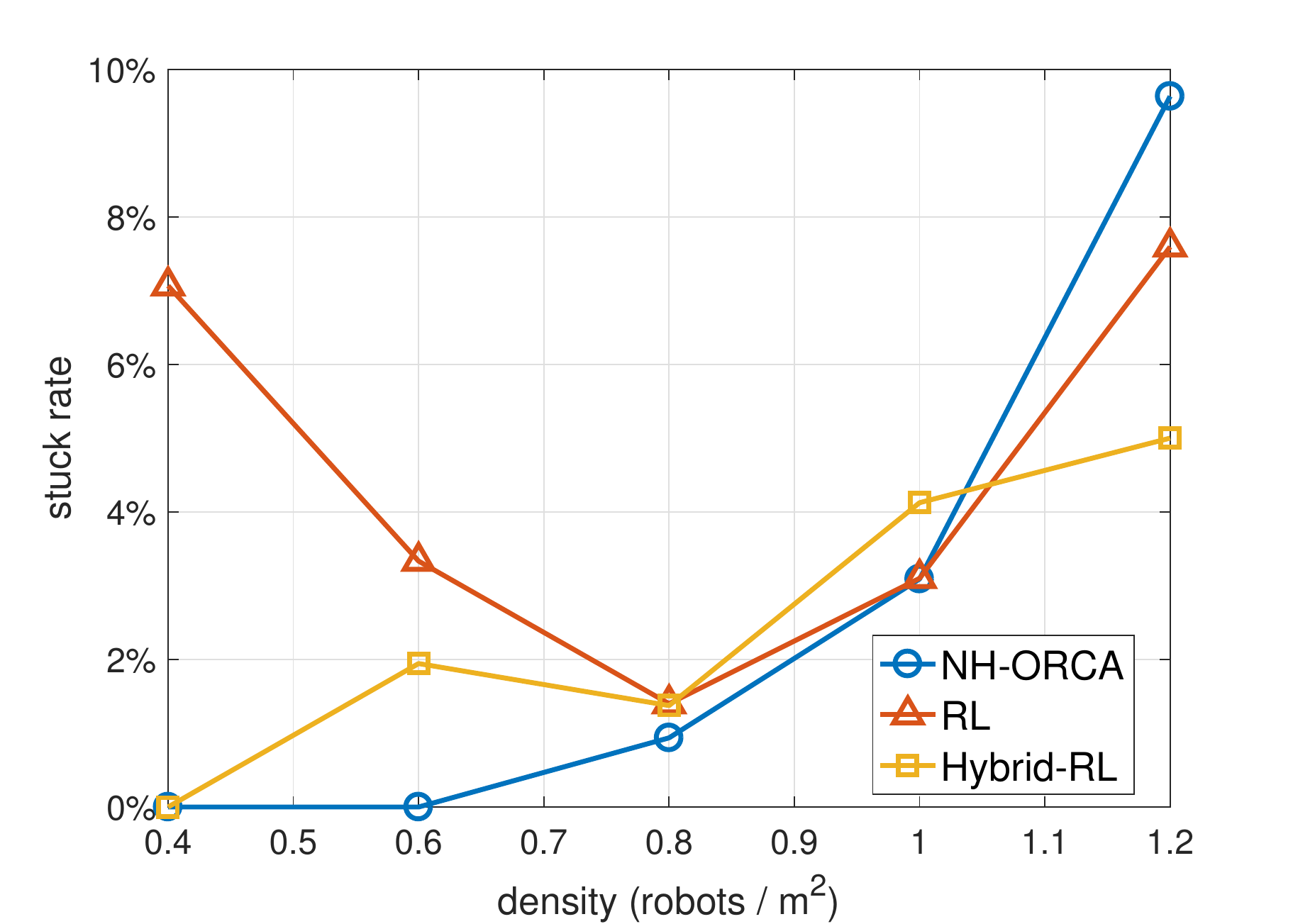}
\caption{Stuck rate w.r.t. agent density}
\label{fig:density_stuck}
\end{subfigure}
\caption{Comparison of the NH-ORCA policy, the RL policy and the Hybrid-RL policy in scenarios with different agent densities.}
\label{fig:robustness_density}
\end{figure}

From the experimental results shown in \prettyref{fig:density_failure}, we observe that the failure rates of all methods increase when the agent density increases. This is because in a crowded environment, each robot has limited space and time window to accomplish the collision-free movement and it has to deal with many neighboring agents. Thus, the robots are more likely running into collisions and getting stuck in congestion. Among three methods, the Hybrid-RL policy achieves the lowest failure rate and collision rate, as shown in \prettyref{fig:density_failure} and \prettyref{fig:density_collision}. 
Both the RL and Hybrid-RL policy's collision rates and overall failure rates are significantly lower than that of the NH-ORCA policy. However, as we observe in \prettyref{fig:density_stuck}, the RL policy's stuck rate is slightly higher than that of the NH-ORCA policy in low density situations. This is due to the limitation of the RL policy that we discussed in \prettyref{sec:hybrid} when introducing the hybrid control framework: the robot controlled by the RL policy may wander around its goal when it is in the close proximity of the goal. Such navigation artifact of the RL policy takes place in the high density scenarios, which leads to higher stuck rates. 
Resolving this difficulty by adopting proper sub-policies, our Hybrid-RL policy enable robots to reach goals efficiently and thus it significantly outperforms the RL policy and the NH-ORCA policy in performance, as shown in \prettyref{fig:density_stuck}.

We also observe two interesting phenomena in \prettyref{fig:robustness_density}. First, the RL policy achieves the lowest stuck rate at the density $0.8$ robots \SI{}{1/m^2}. We believe that this is because the average density in the training set is close to $0.8$ robots \SI{}{/m^2} and thus the RL policy slightly overfits in this case. Second, we observe that the stuck rate of the Hybrid-RL policy is a bit higher than RL policy at the density $1$ robots \SI{}{/m^2}. This is because in emergent cases, the RL policy has a higher probability to directly run into the obstacles and get collision while the Hybrid-RL policy will use $\pi\textsubscript{safe}$ to keep safe but may get stuck. This results in the slight lower stuck rate but much higher collision rate of the RL policy compared to the Hybrid-RL policy at the density $1$ robots \SI{}{/m^2}.

\subsubsection{\textbf{Different agent sizes}}
In our training scenarios, all robots are of the same disc shape with a radius of \SI{0.12}{m}. When transferring the learned policy to unseen simulated or real world scenarios, the robots may have different shapes and different sizes. 

In \prettyref{sec:gen}, we directly deploy our learned policy to agents with similar sizes but shapes different with the prototype agents used in training, and we demonstrate that our method still provides satisfactory performance. In this part, we focus on evaluating the robustness of our policy when the robot's size is significantly different from the setup in training. 
In particular, we generate five testing scenarios, each of which is a $7$-meter by $7$-meter square region with $10$ round shape robots with the same radius, where the radius is $0.12$, $0.18$, $0.24$, $0.30$, and $0.36$ meters, respectively.
In other words, the smallest agent has the same size as that of the prototype agent, and the largest agent has the radius $3$ times larger than that of the prototype agent.

\begin{figure*}[h] 
\centering
\includegraphics[width=1.0\linewidth]{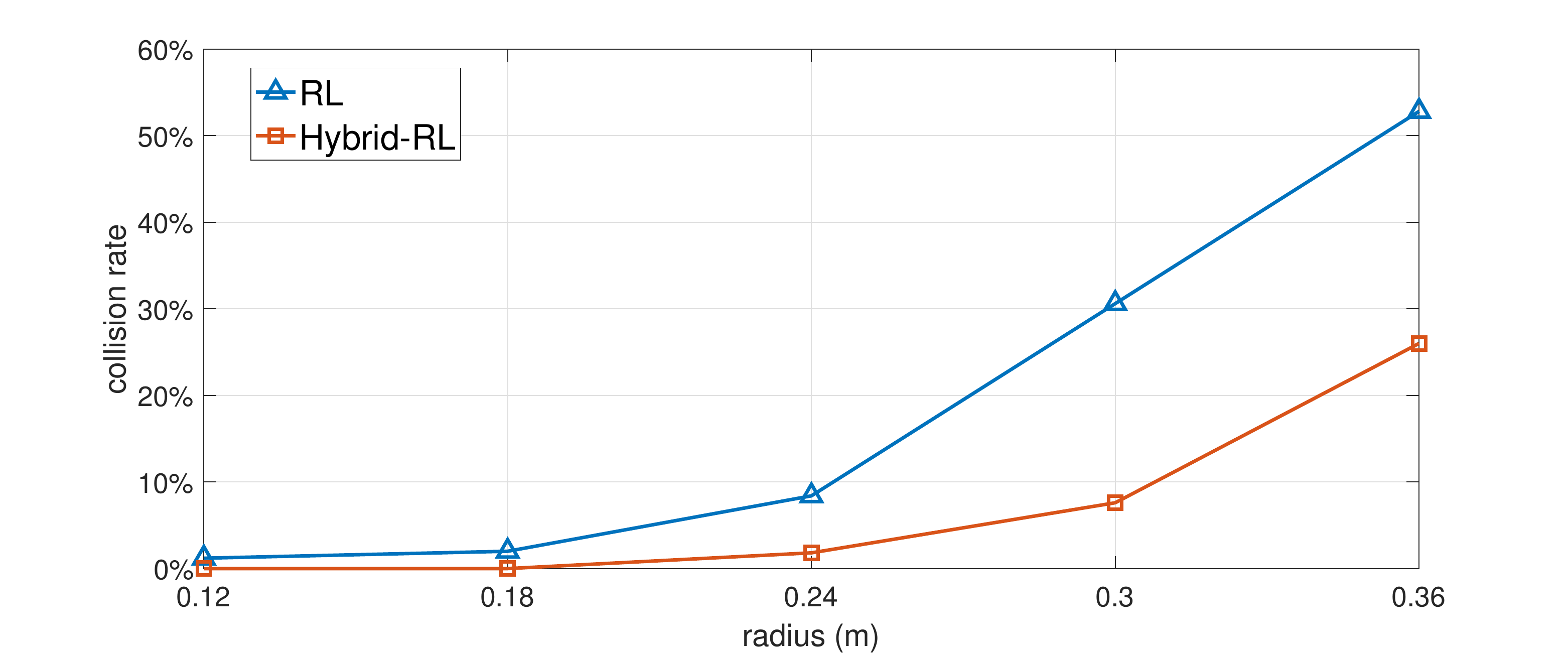}
\caption{Comparison of the collision rates of the RL policy and the NH-ORCA policy as the robot radius increases.}
\label{fig:size_collision}
\end{figure*}

From the results shown in \prettyref{fig:size_collision}, we observe that our Hybrid-RL policy can be directly deployed to robots that are $2$ times larger in radius (and $4$ times greater in the area) than the prototype agent used for training, and the robust collision avoidance behavior can still be achieved, with a failure rate of about $3$\%. When the size disparity between the training and test becomes larger, the collision rate of the Hybrid-RL policy increases but is still significantly lower than that of the RL policy. This implies that the Hybrid-RL policy has learned to maintain an appropriate safety margin with other obstacles when navigating through them.
In addition, the hyper-parameter $r\textsubscript{safe}$ about the size of safety radius in the hybrid control framework also contributes to an excellent balance between the navigation efficiency and safety.

\subsubsection{\textbf{Different maximum velocities}}
In our proposed policy, the output linear and angular velocities of the policy network are truncated to a certain range ($[0, 1]$ \SI{}{m/s} for the linear velocity and $[0, 1]$ \SI{}{rad/s} for the angular velocity in our training setup) to reduce the exploration space of reinforcement learning and thus to speed up the training process.
However, since the robots in a variety of applications may have different dynamics and different ranges for actions, we here focus on studying the robustness of the learned policy for the testing robots with different maximum linear and angular velocities.

In particular, we allocate $10$ robots with the same range for the linear and angular velocities in a $7$-meter by $7$-meter square region. We create five testing scenarios, whose maximum setups for linear and angular velocities are (\SI{1}{m/s}, \SI{1}{rad/s}), (\SI{1.5}{m/s}, \SI{1.5}{rad/s}), (\SI{2}{m/s}, \SI{2}{rad/s}), (\SI{2.5}{m/s}, \SI{2.5}{rad/s}), and (\SI{3}{m/s}, \SI{3}{rad/s}), respectively. Note that since the network output is constrained in training, there is no need to evaluate the velocities lower than (\SI{1}{m/s}, \SI{1}{rad/s}).

\begin{figure}[!htb] 
\captionsetup[subfigure]{justification=centering}
\centering
\begin{subfigure}{1\linewidth}
\includegraphics[trim=10 0 20 0, clip, width=1.0\linewidth]{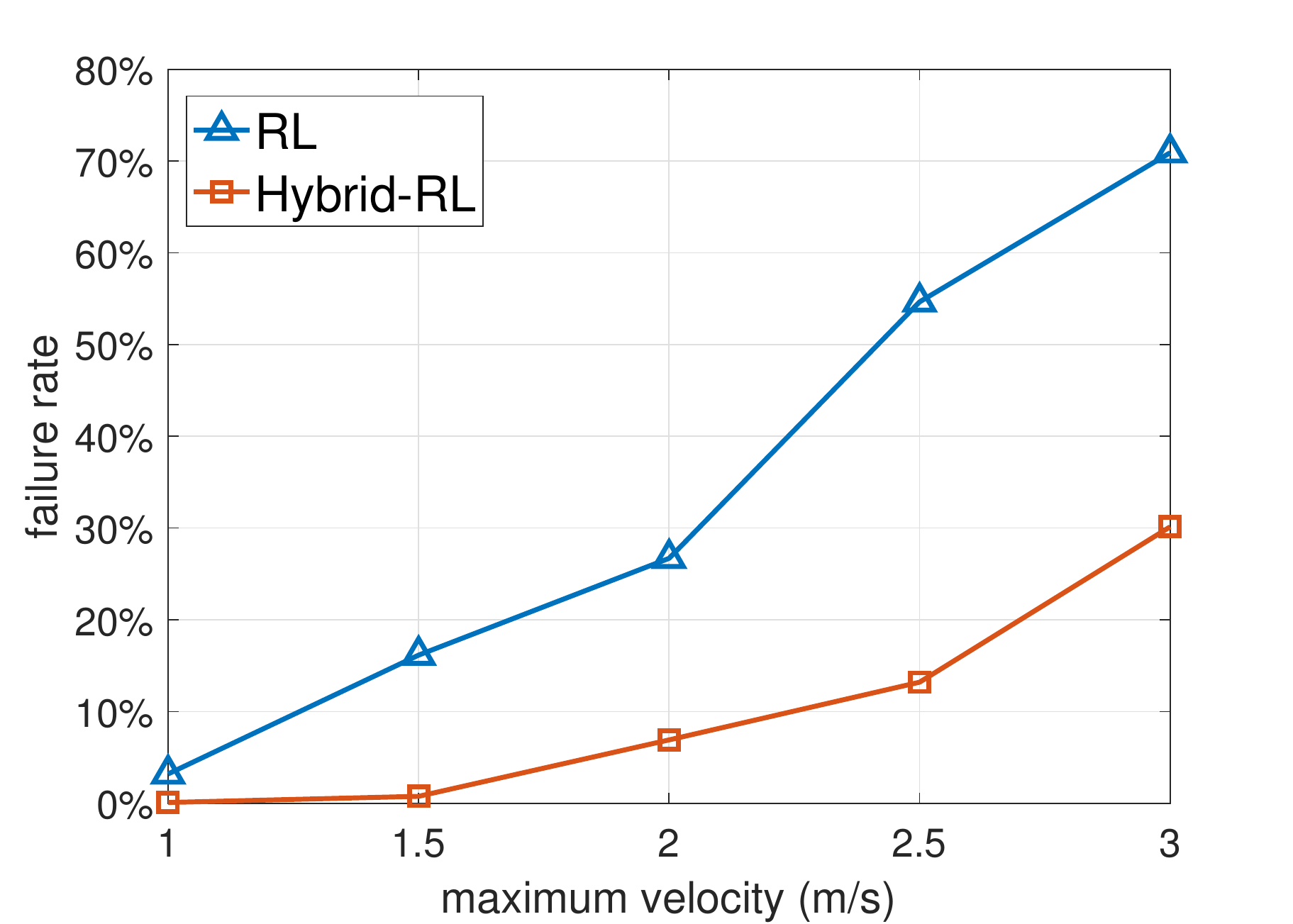}
\caption{Failure rate w.r.t. maximum velocity}
\label{fig:velocity_failure}
\end{subfigure}
\begin{subfigure}{1\linewidth}
\includegraphics[trim=10 0 20 0, clip, width=1.0\linewidth]{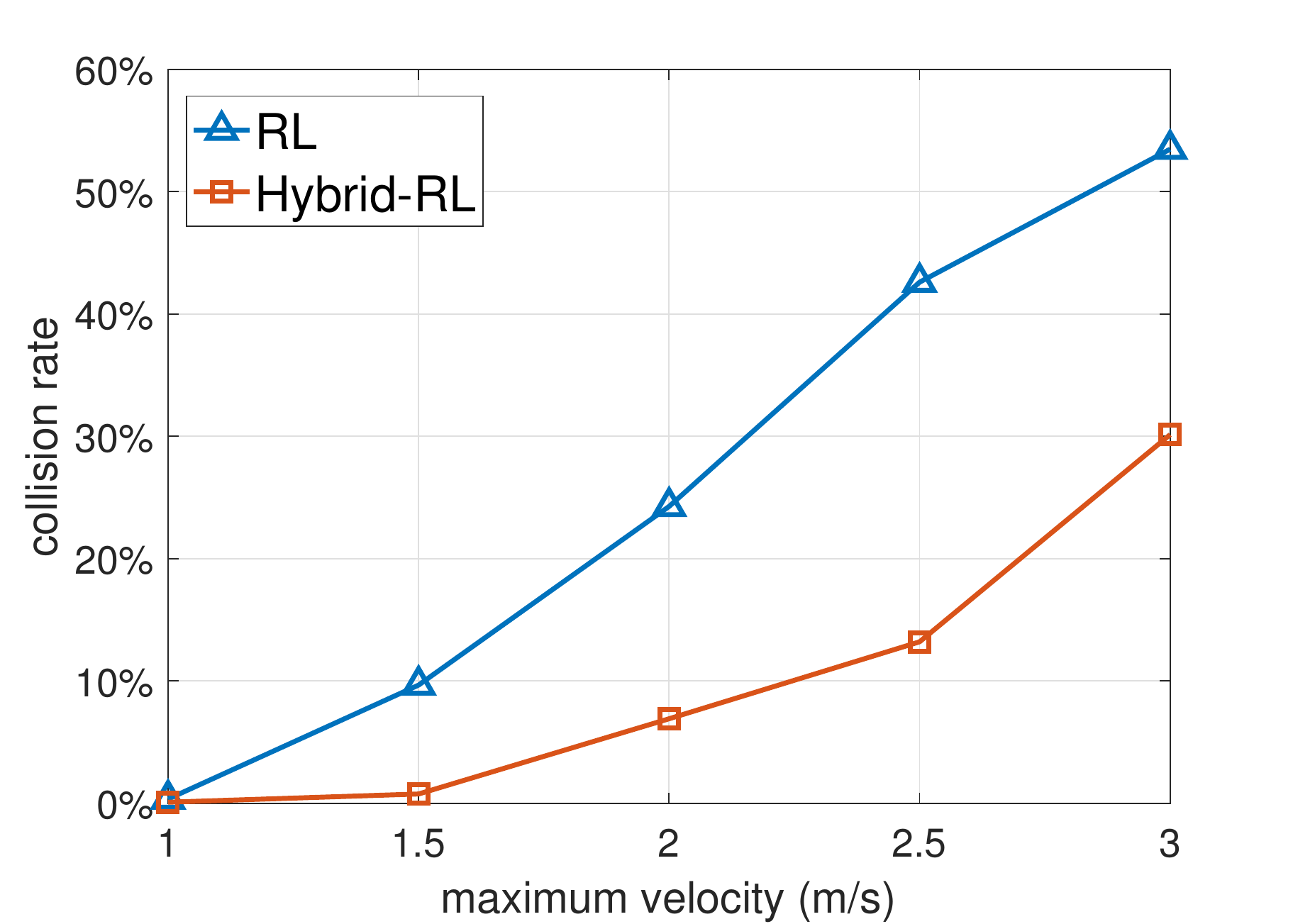}
\caption{Collision rate w.r.t. maximum velocity}
\label{fig:velocity_collision}
\end{subfigure} 
\begin{subfigure}{1\linewidth}
\includegraphics[trim=10 0 20 0, clip, width=1.0\linewidth]{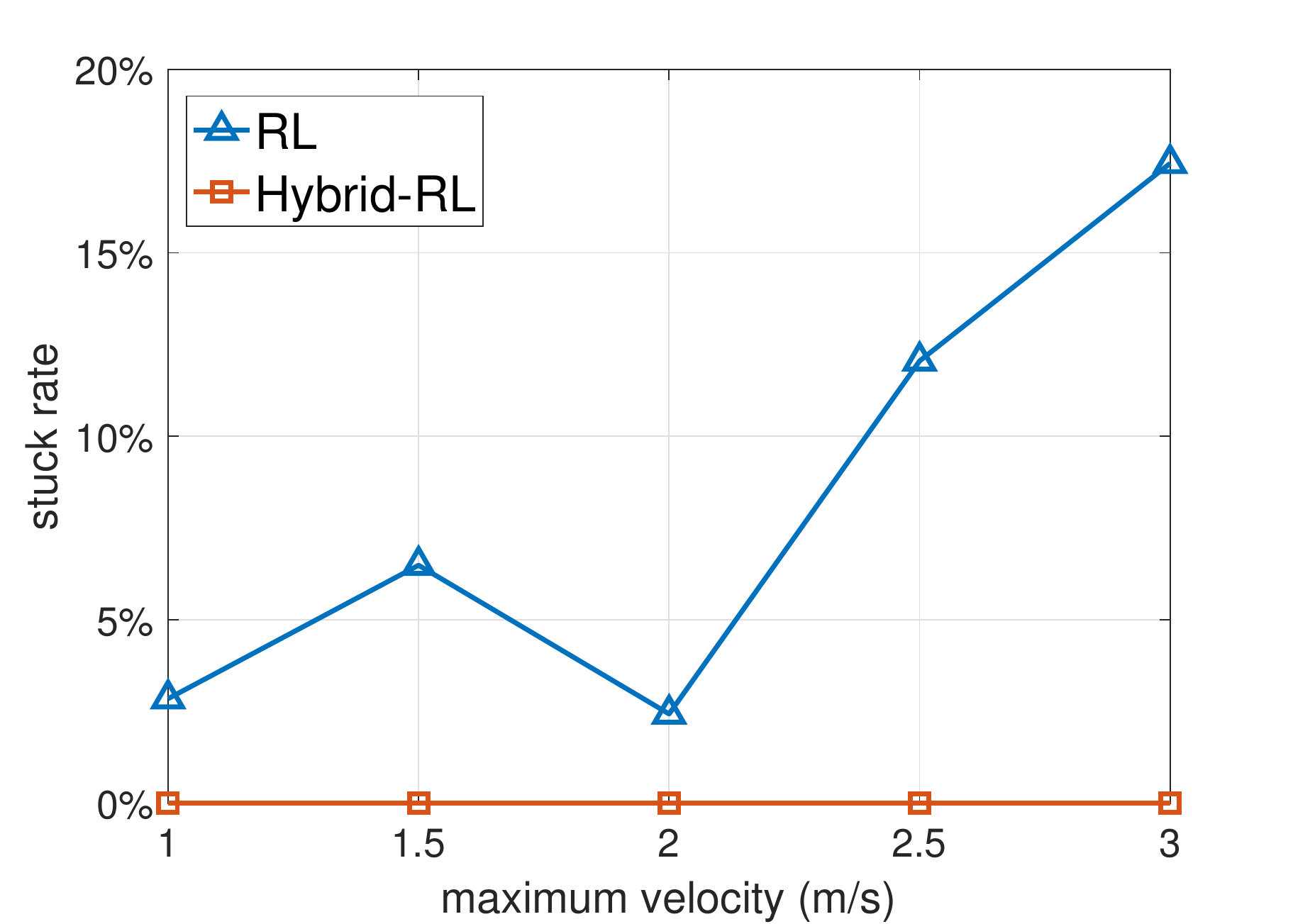}
\caption{Stuck rate w.r.t maximum velocity}
\label{fig:velocity_stuck}
\end{subfigure}
\caption{Comparison of the RL policy and the Hybrid-RL policy for scenarios with different maximum velocity setup for robots.}
\label{fig:robustness_velocity}
\end{figure}

The experimental result in \prettyref{fig:robustness_velocity} shows that the RL policy is very sensitive to the changes of the network output range, i.e., as the maximum speed increases, its failure rate increases quickly to an unacceptable level, while our Hybrid-RL policy's collision rate is still below $10$\% and thus is still relatively safe even when the maximum speed is two times the value in training. In addition, the Hybrid-RL policy's stuck rate remains close to zero when the maximum speed is three times of the training setup, while the RL policy's stuck rate increases to be about $20$\%, as shown in \prettyref{fig:velocity_stuck}. Thanks to the special sub-policy for the emergent situation (as discussed in \prettyref{sec:hybrid}), our Hybrid-RL policy outperforms the RL policy significantly.

\subsubsection{\textbf{Different control frequencies}}
In our training setup, we set the control frequency of executing our collision avoidance policy to \SI{10}{\Hz}. 
Nevertheless, the actual control frequency depends on many factors, including the available computation resources of the on-board computer, the sensor's measurement frequency, and the localization system's frequency. As a result, it is difficult to maintain the control frequency at \SI{10}{\Hz} for long-term execution. In this part, we evaluate whether our collision avoidance policy are able to avoid obstacles when the robot's control frequency varies. 

\begin{figure*}[h] 
\centering
\includegraphics[width=1.0\linewidth]{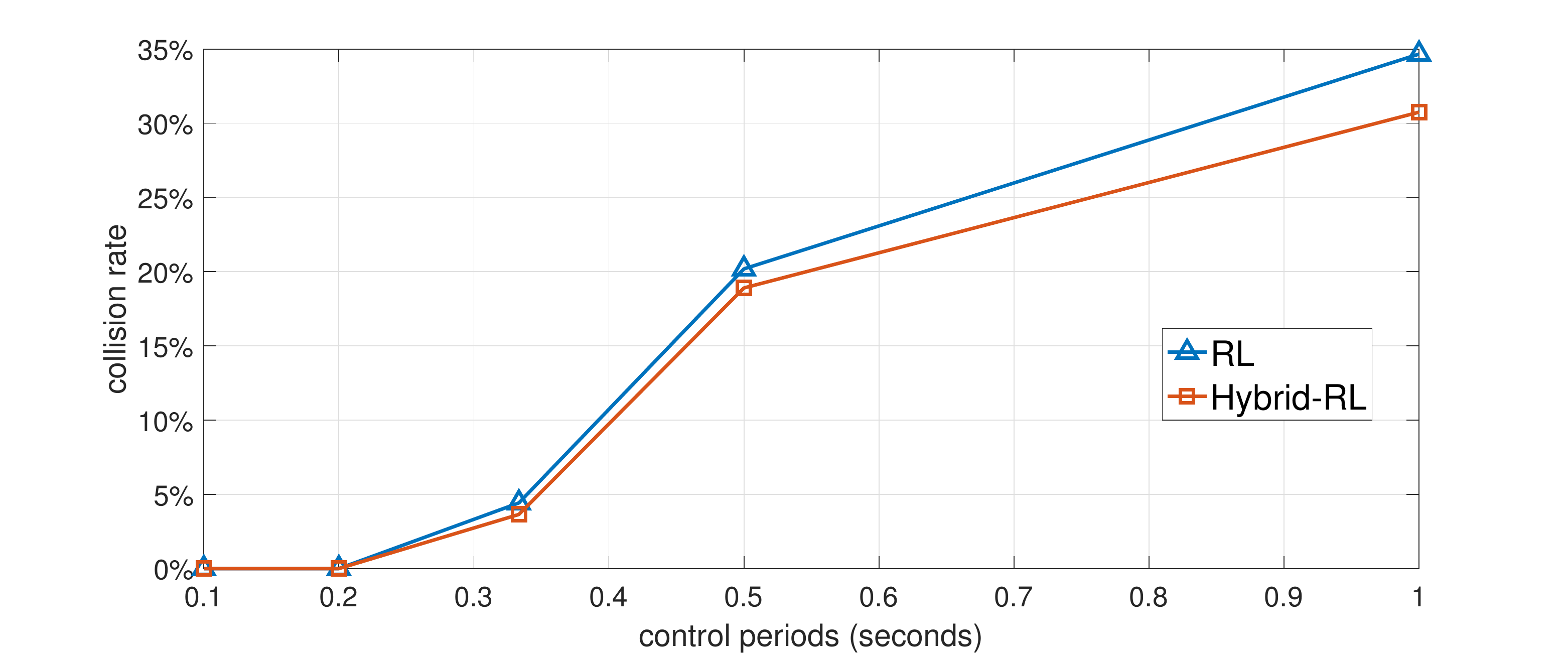}
\caption{Comparison of collision rates between the RL policy and the Hybrid-RL policy for different control periods.}
\label{fig:hz_collision}
\end{figure*}

We deploy $10$ robots in a $7$-meter by $7$-meter square region, and run five experiments with the control frequencies to be \SI{10}{\Hz}, \SI{5}{\Hz}, \SI{3}{\Hz}, \SI{2}{\Hz} and \SI{1}{\Hz} respectively. In other words, the robots' control period is \SI{0.1}{s}, \SI{0.2}{s}, \SI{0.3}{s}, \SI{0.5}{s} and \SI{1}{s}, respectively. From the experimental results in \prettyref{fig:hz_collision}, we observe that the collision rate of our policy increases dramatically when the control frequency is below \SI{3}{\Hz}. This is consistent with human experience: the navigation is difficult when blinking eyes frequently. The Hybrid-RL policy slightly outperforms the RL policy when the control frequency varies.
 
\subsubsection{\textbf{Summary}}
After the experimental evaluations above, we demonstrate that our Hybrid-RL policy can be well generalized to unseen scenarios and is robust to the variations of the scenario configuration including densities, sizes, maximum velocities and control frequencies of the robots. In this way, we validate that the Hybrid-RL policy is able to bridge the sim-to-real gap and has the potential to be directly deployed on physical robots with different setups. 
\section{Real World Experiments}
\label{sec:real_exp}

In this section, we demonstrate that our reinforcement learning based policy can be deployed to physical robots. Besides the generalization capability and uncertainty modeling that we discussed in \prettyref{sec:sim_exp}, there still exist other challenges of transferring our learned policy from simulation to the real world. First, in real situations, noises in observation are ubiquitous due to the imperfect sensing. Second, the clock of each individual robot is not synchronized with each other and such an asynchronized distributed system is challenging for control. Finally, it is not guaranteed that all robots can provide consistent behavior given the same control command, because many real-world factors such as mechanics details, motor properties, and frictions cannot be accurately modeled in a simulation environment. 

In the following sub-sections, we first briefly introduce the hardware setup of our robots. Then, we present the multi-robot scenarios for evaluation. Lastly, we demonstrate the collision avoidance behavior of the robot controlled by our learned policy when interacting the real human crowd.

\subsection{\textbf{Hardware setup}}

\begin{figure}[htb]
\includegraphics[width=1.0\linewidth]{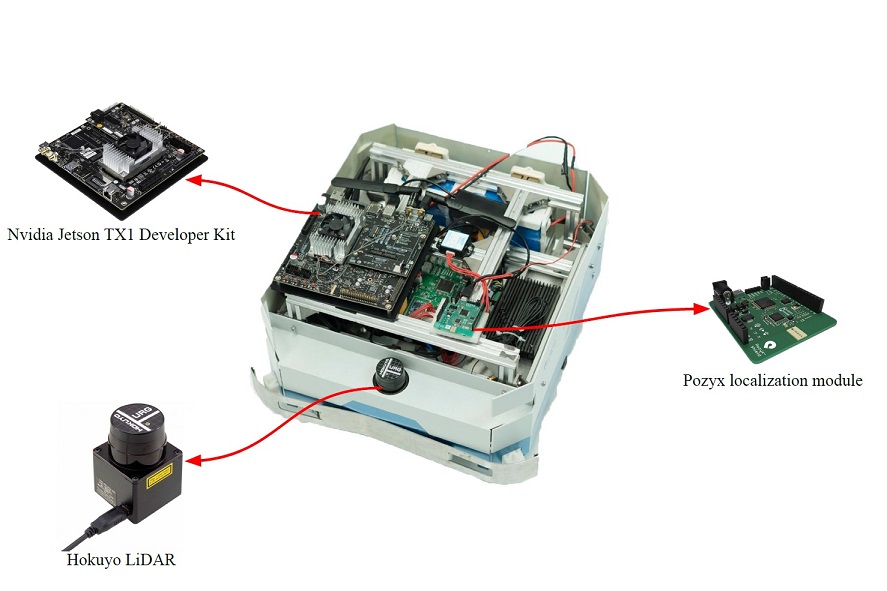}
\caption{The self-developed mobile platform used to verify the transferability of our reinforcement learning based policy deployed on physical robots.}
\label{fig:hardware_ourselves}
\end{figure}

To validate the transferability of our Hybrid-RL policy, we use a self-developed mobile robot platform for the real-world experiments. The mobile platform has a squared shape with a side length of \SI{46}{cm}, as shown in \prettyref{fig:hardware_ourselves}. Its shape, size, and dynamic characteristics are completely different from the prototype agents used in the training scenarios. Hence, the experiments based on such a physical platform provide a convincing evaluation about whether our method can bridge the sim-to-real gap. 

In our distributed multi-robot systems, each robot is mounted a sensor for measuring the distance to the surrounding obstacles, a localization system for measuring the distance from the robot to its goal, and an on-board computer for computing the navigation velocity in real-time. We use the Hokuyo URG-04LX-UG01 2D LiDAR as our laser scanner, the Pozyx localization system based on the UWB (Ultra-Wide Band) technology to localize our ground vehicles' 2D positions, and the Nvidia Jetson TX1 as our on-board computing device. 
The detail about each component of our sensor kit is as follows:
\begin{itemize}
\item \textit{LiDAR}: Its measurement distance is from \SI{20}{mm} to \SI{4000}{mm}. Its precision is $\pm$ \SI{30}{mm} when the measuring distance is between \SI{60}{mm} and \SI{1000}{mm}, and the precision is $3 \%$ when the measuring distance is between \SI{1000}{mm} to \SI{4000}{mm}. Its angle range is 
[\minus \ang{ 120;;}, \ang{120;;}] (but we only use the measurements from \minus \ang{90;;} to \ang{90;;}), and the angle resolution is \ang{0.36;;}. The sensor's update frequency is \SI{10}{\Hz}.
\item \textit{UWB localization system}: 
Its measurement accuracy is around $\pm$\SI{150}{mm} in our experiment. Note that unlike other centralized collision avoidance systems, our solution does not need the localization for collision avoidance and thus the $\pm$\SI{150}{mm} precision is sufficient.
\item \textit{On-board computer}: Nvidia Jetson TX1 is an embedded computing platform consisting of a Quad-core ARM A57 CPU and a 256-core Maxwell graphics processor.
\end{itemize}

\subsection{\textbf{Multi-robot scenarios}}
In this section, we design a series of real-world multi-robot navigation experiments to test the performance of our Hybrid-RL policy on the physical robots. First, we test the performance of the robots moving in a basic scenario without any obstacles. Next, we validate the policy on robotic platforms with more complex scenarios by adding static and moving obstacles (e.g. pedestrians). In addition, we increase the number of robots in the same workspace (i.e., increasing the density of robots) to further challenge the navigation algorithm's capability. Finally, we design a scenario which emulates the autonomous warehouse application to provide a comprehensive evaluation about our distributed collision avoidance policy. 

\begin{figure*}
\centering
\begin{subfigure}{1\linewidth}
\includegraphics[width=1.0\linewidth]{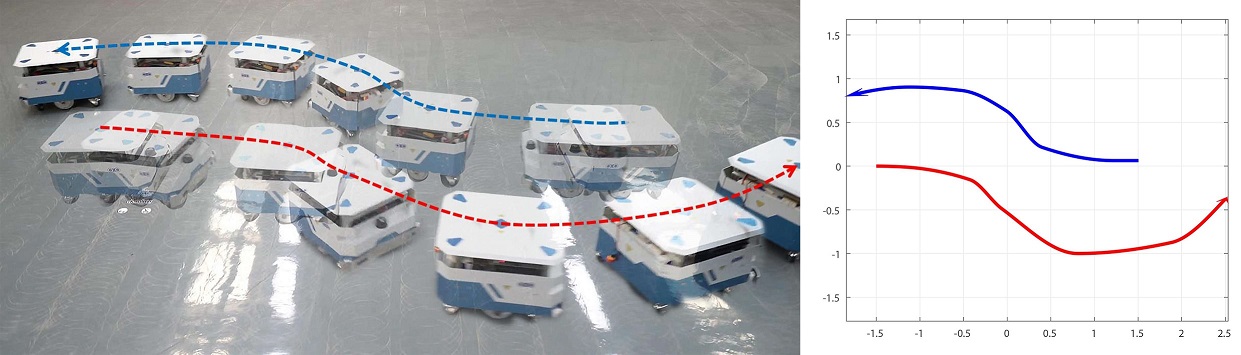}
\caption{Trajectories of two robots in the swap scenario}
\label{fig:2_encounter} 
\end{subfigure} 
\begin{subfigure}{1\linewidth}
\includegraphics[width=1.0\linewidth]{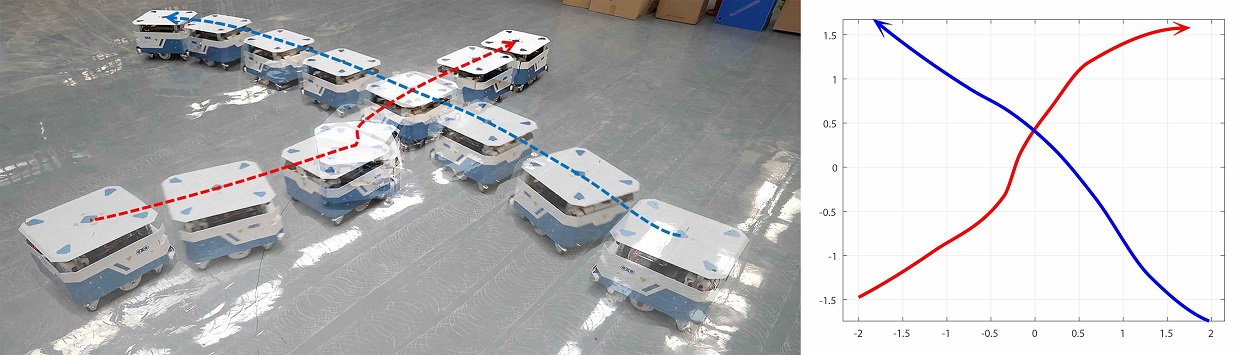}
\caption{Trajectories of two robots in crossing scenario}
\label{fig:2_cross} 
\end{subfigure} 
\begin{subfigure}{1\linewidth}
\includegraphics[width=1.0\linewidth]{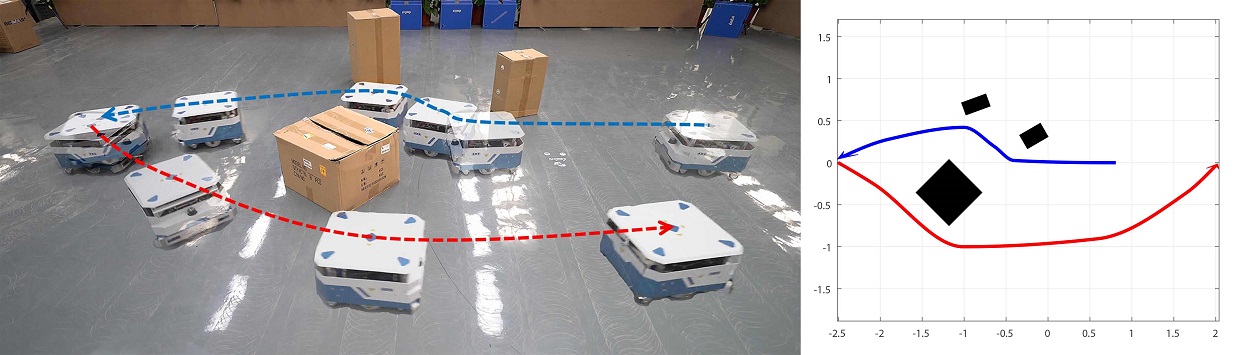}
\caption{Trajectories of two robots in the swap scenario with static obstacles}
\label{fig:2_encounter_obstacles} 
\end{subfigure} 
\begin{subfigure}{1\linewidth}
\includegraphics[width=1.0\linewidth]{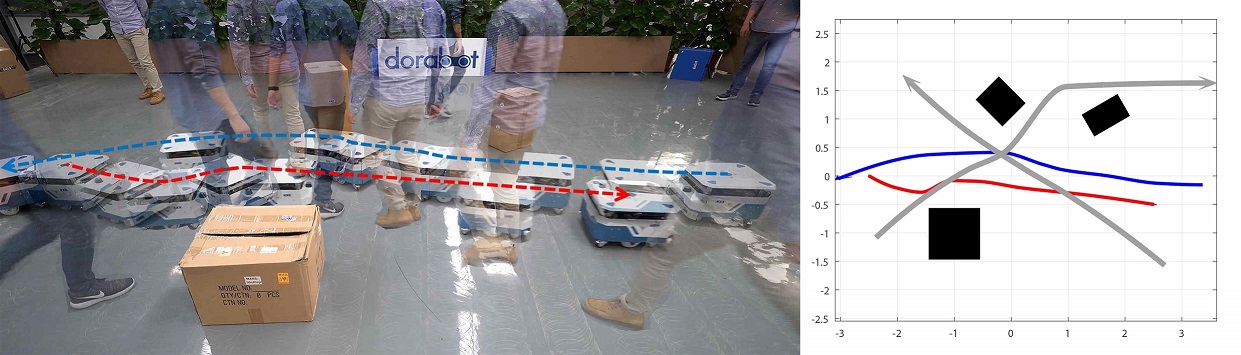}
\caption{Trajectories of two robots in the swap scenario with static obstacles and moving obstacles}
\label{fig:2_encounter_people_obstacles} 
\end{subfigure} 
\caption{(a) Swap scenario: two robots moving in the opposite directions swap their positions. (b) Crossing scenario: the trajectories of two robots intersect. (c) Static obstacles are placed in a swap scenario. (d) Both static obstacles and moving obstacles (i.e. pedestrians) are placed in the swap scenario. The figures on the left show the trajectories of the physical robots in the real scenario, and the figures on the right show the trajectories of the physical robots in the 2D coordinate system, which correspond to the figures on the left. In the figures on the right, the black boxes refer to the static obstacles; the gray trajectories refer to the pedestrians' paths; other colored trajectories are the robots' paths. Please refer to the video for more details about the physical robots' behavior in these scenarios.}
\label{fig:2_benchmark}
\end{figure*}

As the first step, we use a swap scenario (\prettyref{fig:2_encounter}) and a crossing scenario (\prettyref{fig:2_cross}) to demonstrate that two robots can reliably avoid each other based on our Hybrid-RL policy. Although the size, shape and dynamic characteristics of these physical robots are different from the agents trained in the simulation, our deep reinforcement learning based collision avoidance policy still performs well in these two scenarios. 

Next, we add static obstacles in the swap benchmark and test whether the two robots can still successfully navigate in this complex scenario in a map-less manner. From the trajectories in \prettyref{fig:2_encounter_obstacles}, we observe that both robots smoothly avoid static obstacles that are randomly placed in the environment. Then we allow two pedestrians to interfere the robots in the same environment. Taking the trajectories shown in \prettyref{fig:2_encounter_people_obstacles} for example, the two robots adaptively slow down and wait for the pedestrians and then start again towards the goal after the pedestrian passes.
It is important to note that there is no pedestrian data introduced in the training process, and the pedestrians' shape and dynamic characteristics are quite different from other robots and static obstacles. Moreover, we do not leverage any pedestrian tracking or prediction methods in the robot's navigation pipeline. As illustrated in the qualitative results, our learned policy demonstrates an excellent generalization capability in collision avoidance  no only for the static obstacles but also for the dynamic obstacles.
Hence, based on our learned policy without fine-tuning, the robots can easily adapt to new and more challenging tasks. 

\begin{figure*}
\centering
\begin{subfigure}{1\linewidth}
\includegraphics[width=1.0\linewidth]{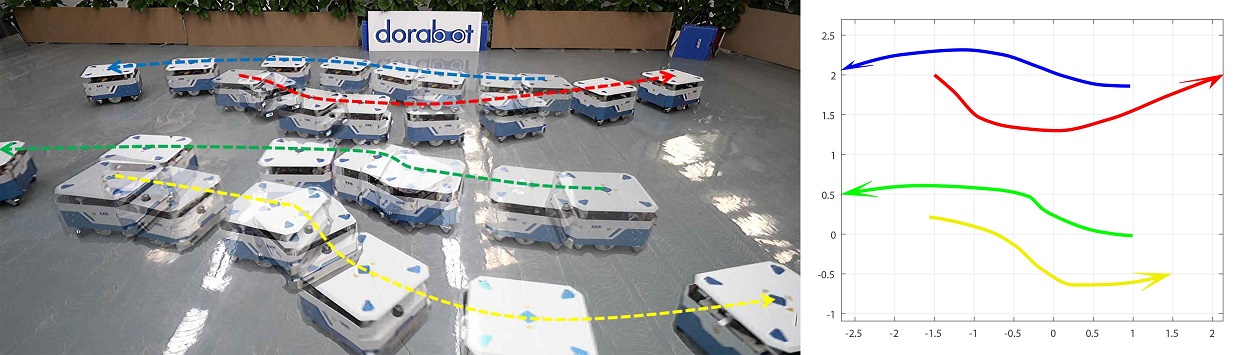}
\caption{Trajectories of two groups of robots in the swap scenario}
\label{fig:4_swap} 
\end{subfigure} 
\\
\begin{subfigure}{1\linewidth}
\includegraphics[width=1.0\linewidth]{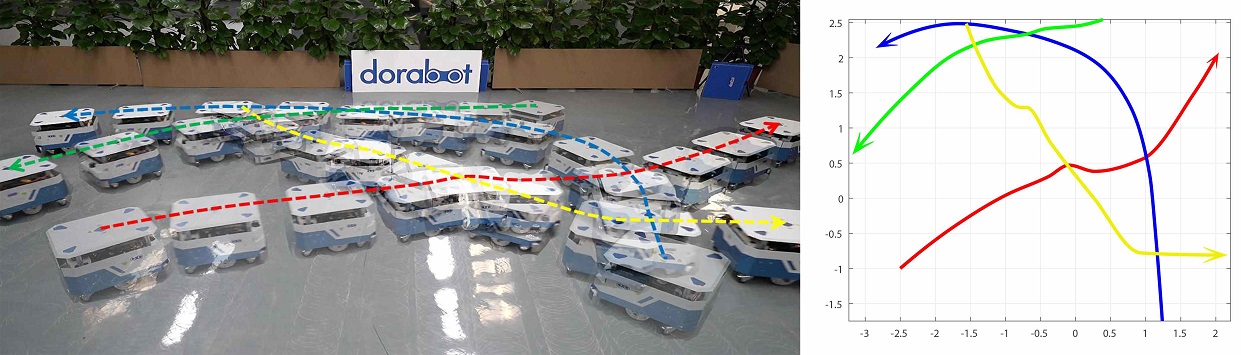}
\caption{Trajectories of four robots in the crossing scenario}
\label{fig:4_cross} 
\end{subfigure} 
\\
\begin{subfigure}{1\linewidth}
\includegraphics[width=1.0\linewidth]{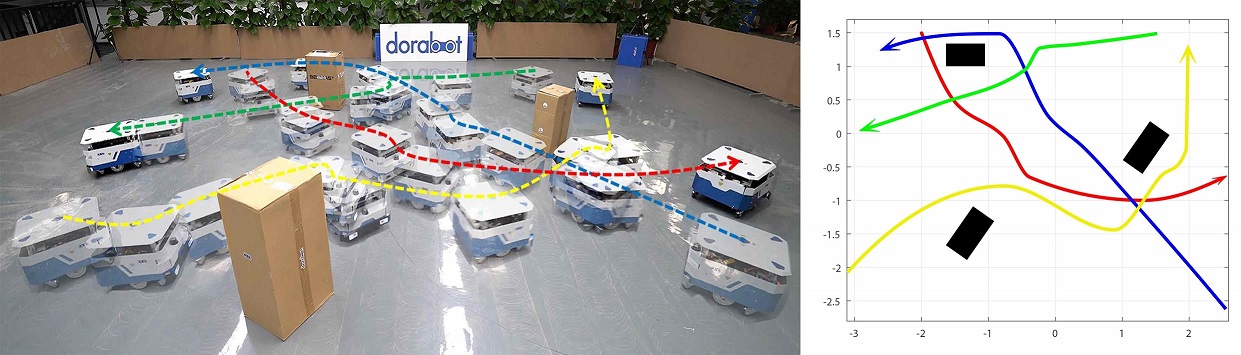}
\caption{Trajectories of four robots in the crossing scenario with static obstacles}
\label{fig:4_cross_obstacles} 
\end{subfigure} 
\\
\begin{subfigure}{1\linewidth}
\includegraphics[width=1.0\linewidth]{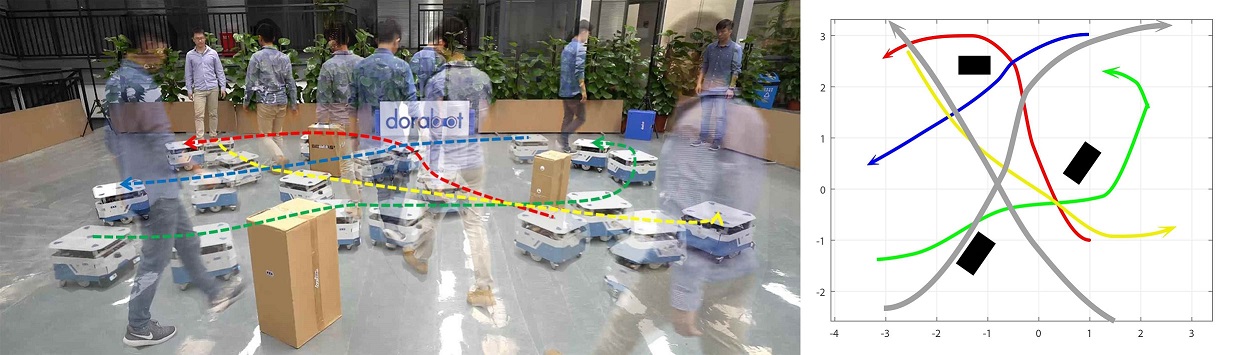}
\caption{Trajectories of four robots in the crossing scenario with static and moving obstacles}
\label{fig:4_cross_people_obstacles} 
\end{subfigure} 
\caption{(a) Swap scenario: two groups of robots moving in the opposite directions swap their positions. (b) Crossing scenario: the path of four robots will intersect. (c) Static obstacles are placed in a crossing scenario. (d) Both static obstacles and moving obstacles (i.e., pedestrians) are placed in the the crossing scenario. The left column shows the trajectories of the physical robots in the real scenario, and the right column shows the trajectories of the physical robots in a 2D coordinate system. In the right column, the black boxes are the static obstacles, the gray trajectories are the pedestrians' paths, and other colored trajectories are the robots' paths. Please also refer to the video for more details about the robots' behavior in this scenario.}
\label{fig:4_benchmark} 
\end{figure*}

We further add another two robots in the two-robot benchmarks and design a four-robot benchmark as shown in \prettyref{fig:4_benchmark}. From the resulting trajectories, we observe that all robots reach their goals successfully without any collisions. However, we also notice that the quality of the robots' trajectories is lower than that in the simpler benchmarks demonstrated in \prettyref{fig:2_benchmark}. This is mainly due to the positioning error of the UWB localization system and the disparity in robot hardwares, both of which are amplified as the number of robots increases.

\begin{figure}[h] 
\captionsetup[subfigure]{justification=centering}
\centering
\begin{subfigure}{0.48\textwidth}
\includegraphics[width=1.0\linewidth]{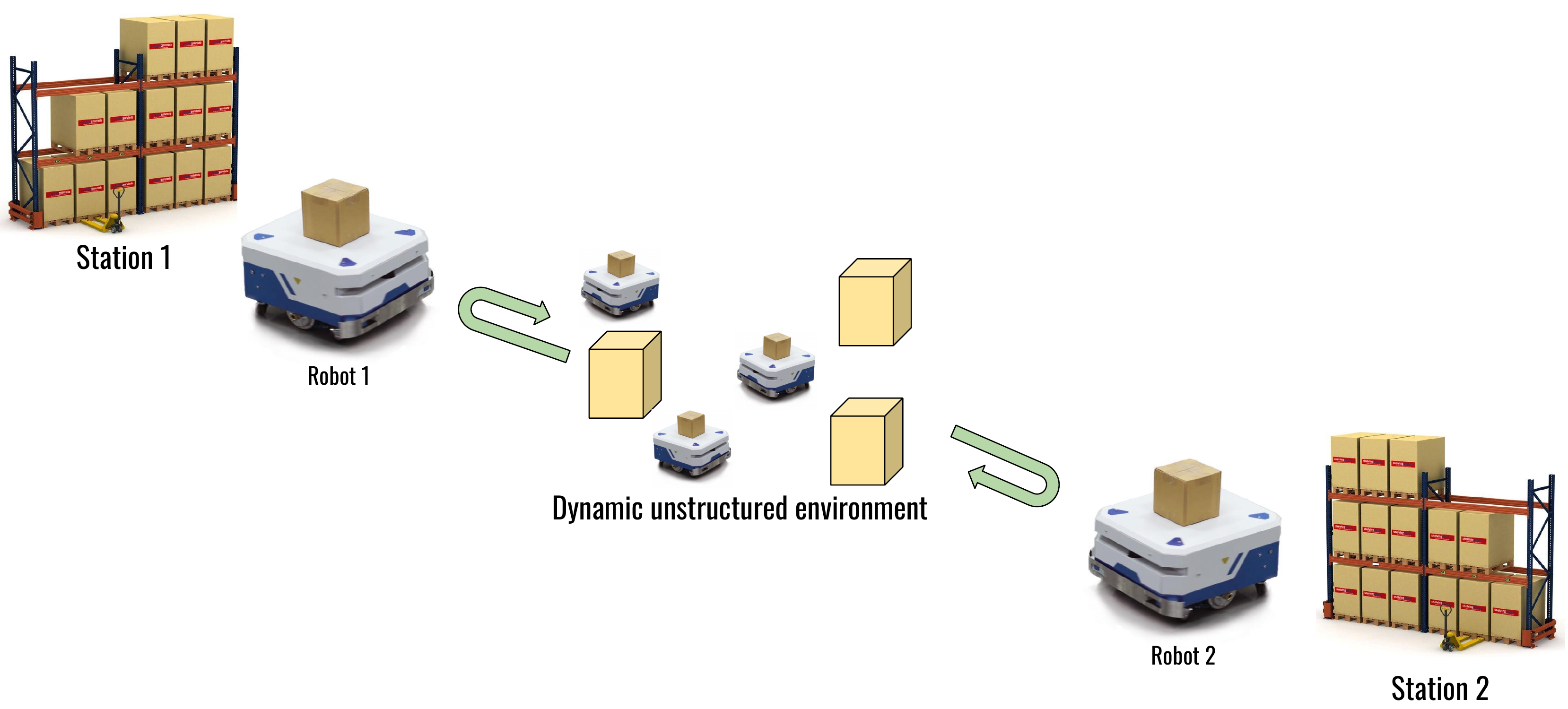}
\caption{The diagram of an autonomous warehouse benchmark without human co-workers.}
\label{fig:application}
\end{subfigure} \\
\begin{subfigure}{0.48\textwidth}
\includegraphics[width=1.0\linewidth]{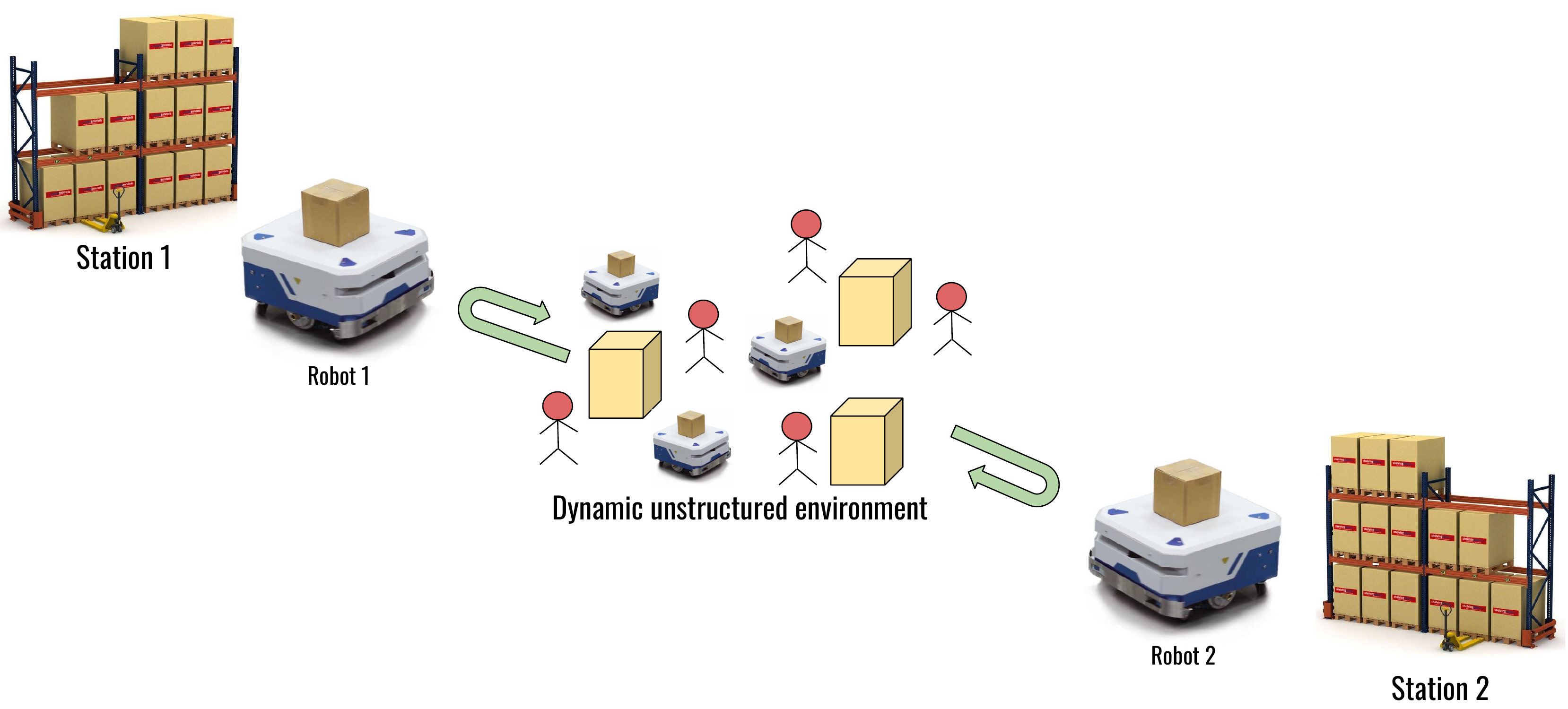}
\caption{The diagram of an autonomous warehouse benchmark with human co-workers.}
\label{fig:complex_application}
\end{subfigure}
\caption{Two types of autonomous warehouse scenarios for evaluating the performance of our multi-robot collision avoidance policy.}
\label{fig:application_scenario}
\end{figure}

\begin{figure*}[h] 
\centering
\includegraphics[width=0.7\linewidth]{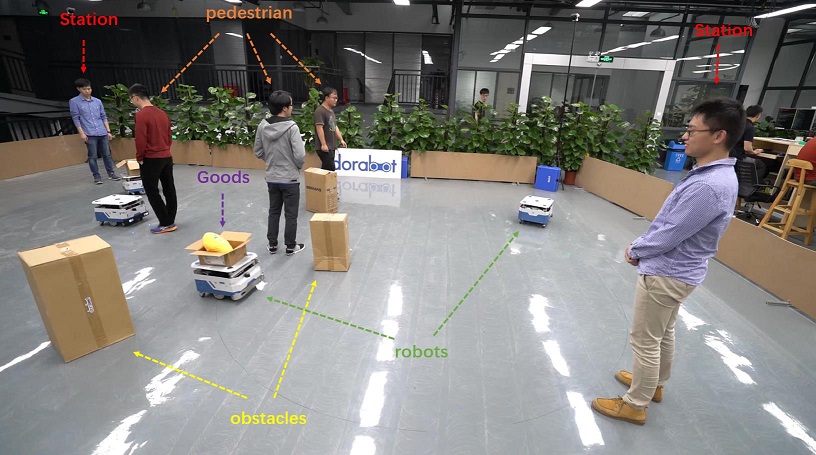}
\caption{The experimental scenario for the autonomous warehouse benchmark with human co-workers. Please refer to the video for more details about the robots' behavior in this scenario.}
\label{fig:real_complex_application}
\end{figure*}

Finally, we design two scenarios which emulate the autonomous warehouse application, where multiple robots work together to achieve efficient and flexible transportation of commodities in an unstructured warehouse environment for a long time. In the first scenario as illustrated in \prettyref{fig:application}, there are two transportation stations in the warehouse and two robots are responsible for transporting objects between these two stations. Next, we add random moving pedestrians as the co-workers in the warehouse and build the second scenario as illustrated in \prettyref{fig:complex_application} and \prettyref{fig:real_complex_application}. The second scenario is more complex, because the robots and the humans share the workspace, especially that the workers may block the robot, which challenges the safety of our multi-robot collision avoidance policy. As illustrated from the results and the attached video, our policy enables multiple robots to accomplish efficient and safe transportation in these two scenarios. In our test, the entire system can safely run for more than thirty minutes without any faults. Please refer to our attached videos for more details.

\subsection{\textbf{Robotic collision avoidance in dense human crowds}}
There are many previous work about navigating a robot toward its goals in a dense human crowd without any collisions~\cite{ess2010object}. The typical pipeline is first predicting pedestrian movements~\cite{Alahi:2016:CVPR,Yi:2016:ECCV,Gupta:2018:CVPR,Kim:2015:BRVO} and then using reactive control (similar to~\cite{Flacco:2012:ICRA}) or local planning for collision avoidance. Since it is difficult to predict the human motion accurately, such a pipeline is not reliable for practical applications. 

In this section, we use our reinforcement learning based collision avoidance policy to achieve high-quality navigation in a dense human crowd. 
Our method is applied for navigation tasks involving multiple cooperative robots in dynamic unstructured environments, but it also provides high-quality collision avoidance policy for a single robot task in a dynamic unstructured environment with non-cooperative agents (e.g., pedestrians).

Our experimental setup is as follows. We attach one UWB localization tag to a person so that the mobile robot will follow this person according to the UWB signal. Hence, the person controls the robot's desired motion direction by providing a time-varying goal. For evaluation, this person leads the robot into the dense human stream, which provides a severe challenge to the navigation policy. We test the performance of our method in a wide variety of scenarios using different types of robots. 

\begin{figure*}[htb]
\centering
\captionsetup[subfigure]{justification=centering}
\begin{subfigure}[b]{0.3\textwidth}
\includegraphics[width=1\linewidth]{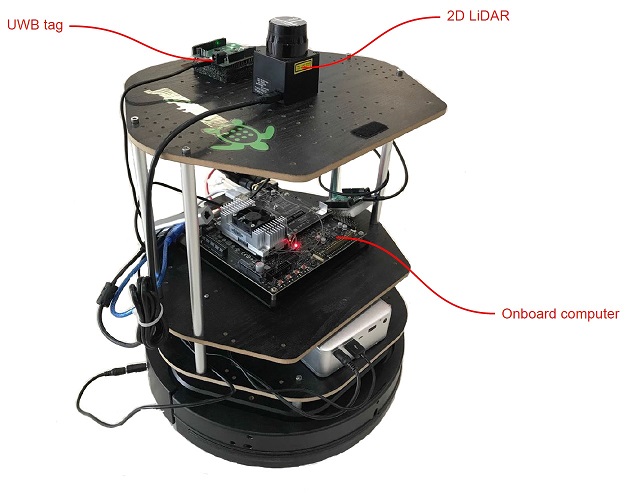}
\caption{Turtlebot}
\label{fig:hardware_turtlebot}
\end{subfigure}
\begin{subfigure}[b]{0.21\textwidth}
\includegraphics[width=0.8\linewidth]{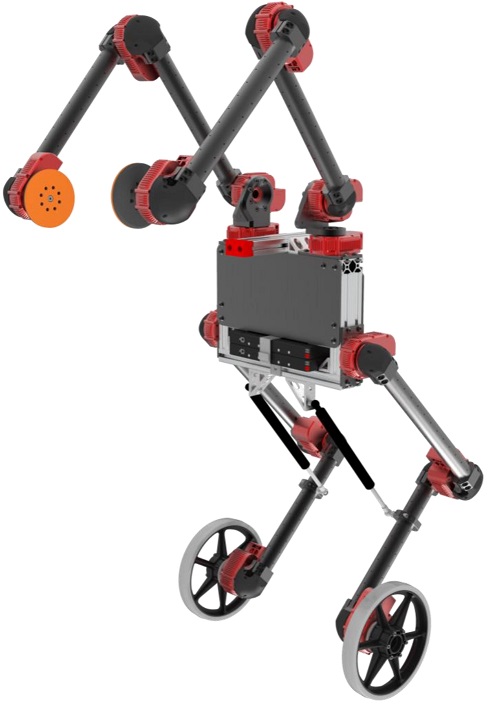}
\caption{Igor robot}
\label{fig:hardware_igor}
\end{subfigure}
\begin{subfigure}[b]{0.20\textwidth}
\centering
\includegraphics[width=0.5\linewidth]{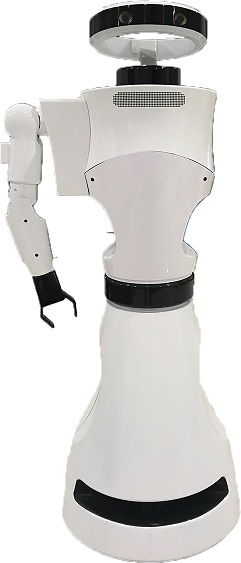}
\caption{Baidu bear robot}
\label{fig:hardware_baidubear}
\end{subfigure}
\begin{subfigure}[b]{0.20\textwidth}
\centering
\includegraphics[width=0.8\linewidth]{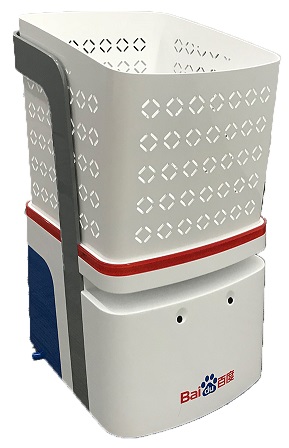}
\caption{Baidu shopping cart}
\label{fig:hardware_cart}
\end{subfigure}
\caption{Four mobile platforms are tested in our experiments for the collision avoidance in the dense human crowds, including the Turtlebot, the Igor robot from Hebi robotics, the Baidu bear robot, and the Baidu shopping cart from Baidu Inc. }
\label{fig:hardware_different}
\end{figure*}

For the testing mobile robot platform, we use four platforms, including the Turtlebot platform, the Igor robot from Hebi robotics, the Baidu Bear robot, and the Baidu shopping cart from Baidu Inc. These four different robots have their own characteristics. The Turtlebot (\prettyref{fig:hardware_turtlebot}) has a size bigger than the \SI{12}{cm}-radius prototype agent used in the training process; the Igor robot (\prettyref{fig:hardware_igor}) is a self-balancing wheeled R/C robot; the BaiduBear robot is a human-alike service robot (\prettyref{fig:hardware_baidubear}); and the Baidu shopping cart used differential wheels as the mobile base, whose weight is quite different with the Turtlebot. The maximum linear velocities for these robots are \SI{0.7}{m/s}, \SI{0.5}{m/s}, \SI{0.4}{m/s}, and \SI{0.4}{m/s}, respectively.

\begin{figure*}
\centering
\captionsetup[subfigure]{justification=centering}
\begin{subfigure}{0.45\textwidth}
\includegraphics[width=0.9\linewidth]{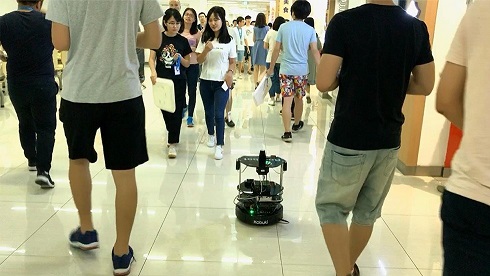}
\caption{canteen}
\label{fig:following_1}
\end{subfigure}
\begin{subfigure}{0.45\textwidth}
\includegraphics[width=0.9\linewidth]{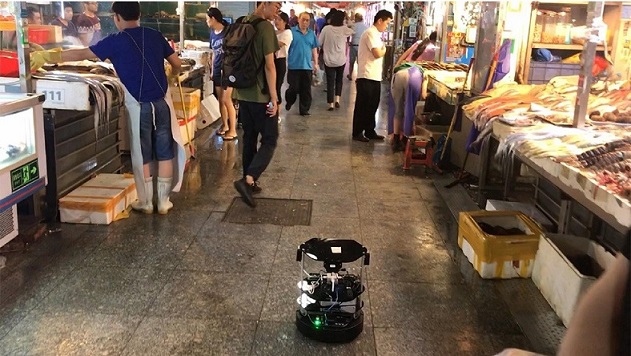}
\caption{farmer's market}
\label{fig:following_2}
\end{subfigure}
\begin{subfigure}{0.45\textwidth}
\includegraphics[width=0.9\linewidth]{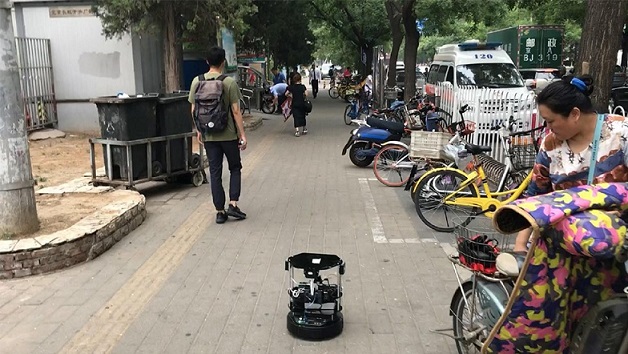}
\caption{outdoor street}
\label{fig:following_3}
\end{subfigure}
\begin{subfigure}{0.45\textwidth}
\includegraphics[width=0.9\linewidth]{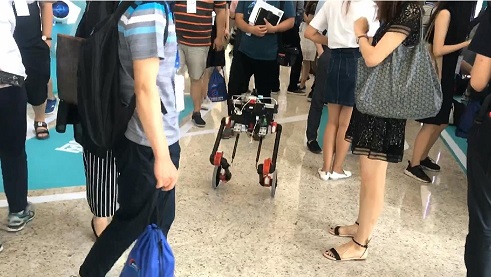}
\caption{indoor conference venue}
\label{fig:following_4}
\end{subfigure}
\caption{Map-less navigation in complex and highly dynamic environments using different mobile platforms. Please refer to the video for more details about the robots' behavior in these challenging scenarios.}
\label{fig:following_benchmarks}
\end{figure*}

Our testing scenarios cover the real-world environments that an mobile robot may encounter, including the canteen, farmer's market, outdoor street, and indoor conference room, as shown in \prettyref{fig:following_benchmarks}. The Turtlebot running our Hybrid-RL policy safely avoids pedestrians and static obstacles with different shapes, even in some extremely challenging situations, e.g., when curious people suddenly block the robot from time to time. However, the Turtlebot's behavior is not perfect. In particular, the robot tends to accelerate and decelerate abruptly within dense obstacles. Such behavior is reasonable according to the objective function of the policy optimization algorithm (in \prettyref{eq:obj}), because the robot will move to its goal as fast as possible in this way and its movement is more efficient. A refined objective function for optimizing a smooth motion in dense crowds would be left as our future work. 

\begin{figure*}
\centering
\includegraphics[width=1\linewidth]{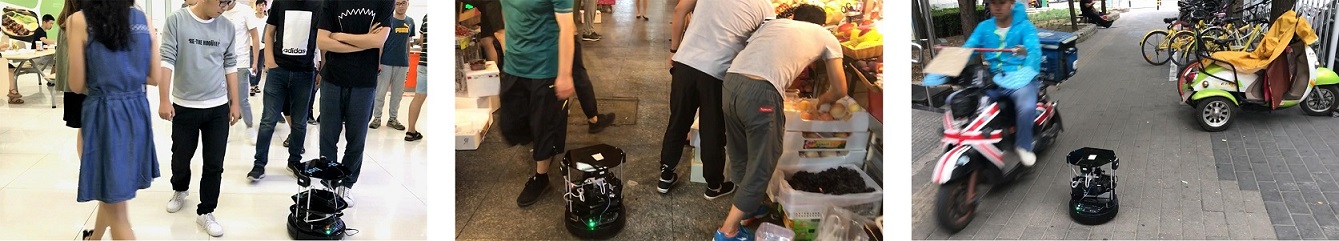}
\caption{Turtlebot works in highly dynamic unstructured scenarios. Please refer to the video for more details about the robots' behavior in these scenarios.}
\label{fig:turtlebot_block}
\end{figure*}

\begin{figure*}
\includegraphics[width=1\linewidth]{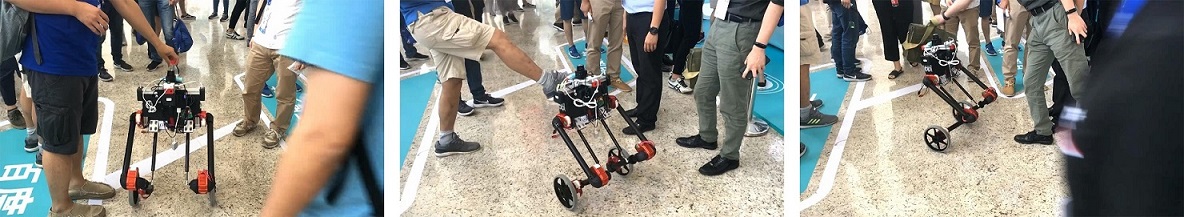}
\caption{Igor reacts quickly to the bottles, human legs, and bags that suddenly appear in its field of view. Please also refer to the video for more details about the robots' behavior in these scenarios.} 
\label{fig:igor_block}
\end{figure*}

\begin{figure*}
\centering
\includegraphics[width=1\linewidth]{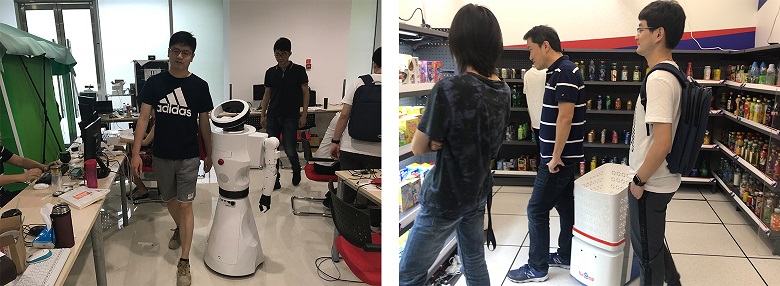}
\caption{Our collision avoidance policy is deployed on Baidu Bear robot and the Baidu shopping cart.}
\label{fig:other_block}
\end{figure*}

More experiments on different robotic platforms deploying our policy work in different highly dynamic scenarios are available in \prettyref{fig:turtlebot_block}, \prettyref{fig:igor_block}, and \prettyref{fig:other_block}. Please refer to the video for more details.

\subsection{\textbf{Summary}}
We demonstrate that our multi-robot collision avoidance policy can be well deployed to different types of real robots, though their shape, size, and dynamic characteristics are quite different from the prototype agents used in the training simulation scenarios. 
We validate the learned policy on various real-world scenarios and show that our Hybrid-RL policy can run robustly for a long time without collisions and the robots' movement easily adapt with the pedestrians in the warehouse scenario. 
We also verify the possibility of using the Hybrid-RL policy to enable a single robot to navigate through a dense human crowd efficiently and safely. 

\section{Conclusion}
\label{sec:conclusion}
In this paper, we present a multi-scenario multi-stage training framework to optimize a fully decentralized sensor-level collision avoidance policy with a robust policy gradient algorithm. We evaluate the performance of our method using a series of comprehensive experiments and demonstrate that the learned policy significantly outperforms the state-of-the-art approaches in terms of the success rate, the navigation efficiency, and the generalization capability. Our learned policy can also be used to navigate a single robot through a dense pedestrian crowd, and illustrate the excellent performance in a wide variety of scenarios and on different robot platforms.

Our work serves as the first step towards reducing the navigation performance gap between the centralized and decentralized methods, though we are fully aware of that, as a local collision avoidance method, our approach cannot completely replace a global path planner when scheduling many 
robots to navigate through complex environments with dense obstacles or with dense pedestrians. Our future work would be how to incorporate our approach with classical mapping methods (e.g. SLAM) and global path planners (e.g. RRT and A$^*$) to achieve satisfactory performance for planning a safe trajectory through a dynamic environment.

\section*{Acknowledgements}
The authors would like to thank Hao Zhang from Dorabot Inc. and Ruigang Yang from Baidu Inc. for their support to us when preparing the physical robot experiments. The authors would like to thank Dinesh Manocha from the University of Maryland for his constructive discussion about the method and the experiments.

{
\bibliographystyle{apacite}
\bibliography{references}

\begin{thebibliography}{}

\bibitem [\protect \citeauthoryear {%
Adouane%
}{%
Adouane%
}{%
{\protect \APACyear {2009}}%
}]{%
adouane2009hybrid}
\APACinsertmetastar {%
adouane2009hybrid}%
\begin{APACrefauthors}%
Adouane, L.%
\end{APACrefauthors}%
\unskip\
\newblock
\APACrefYearMonthDay{2009}{}{}.
\newblock
{\BBOQ}\APACrefatitle {Hybrid and safe control architecture for mobile robot
  navigation} {Hybrid and safe control architecture for mobile robot
  navigation}.{\BBCQ}
\newblock
\BIn{} \APACrefbtitle {9th Conference on Autonomous Robot Systems and
  Competitions.} {9th conference on autonomous robot systems and competitions.}
\PrintBackRefs{\CurrentBib}

\bibitem [\protect \citeauthoryear {%
Alahi%
\ \protect \BOthers {.}}{%
Alahi%
\ \protect \BOthers {.}}{%
{\protect \APACyear {2016}}%
}]{%
Alahi:2016:CVPR}
\APACinsertmetastar {%
Alahi:2016:CVPR}%
\begin{APACrefauthors}%
Alahi, A.%
, Goel, K.%
, Ramanathan, V.%
, Robicquet, A.%
, Fei-Fei, L.%
\BCBL {}\ \BBA {} Savarese, S.%
\end{APACrefauthors}%
\unskip\
\newblock
\APACrefYearMonthDay{2016}{}{}.
\newblock
{\BBOQ}\APACrefatitle {{Social LSTM}: Human Trajectory Prediction in Crowded
  Spaces} {{Social LSTM}: Human trajectory prediction in crowded
  spaces}.{\BBCQ}
\newblock
\BIn{} \APACrefbtitle {IEEE Conference on Computer Vision and Pattern
  Recognition.} {Ieee conference on computer vision and pattern recognition.}
\PrintBackRefs{\CurrentBib}

\bibitem [\protect \citeauthoryear {%
Alonso-Mora%
, Baker%
\BCBL {}\ \BBA {} Rus%
}{%
Alonso-Mora%
\ \protect \BOthers {.}}{%
{\protect \APACyear {2017}}%
}]{%
alonso2017multi}
\APACinsertmetastar {%
alonso2017multi}%
\begin{APACrefauthors}%
Alonso-Mora, J.%
, Baker, S.%
\BCBL {}\ \BBA {} Rus, D.%
\end{APACrefauthors}%
\unskip\
\newblock
\APACrefYearMonthDay{2017}{}{}.
\newblock
{\BBOQ}\APACrefatitle {Multi-robot formation control and object transport in
  dynamic environments via constrained optimization} {Multi-robot formation
  control and object transport in dynamic environments via constrained
  optimization}.{\BBCQ}
\newblock
\APACjournalVolNumPages{The International Journal of Robotics
  Research}{36}{9}{1000--1021}.
\PrintBackRefs{\CurrentBib}

\bibitem [\protect \citeauthoryear {%
Alonso-Mora%
, Breitenmoser%
, Rufli%
, Beardsley%
\BCBL {}\ \BBA {} Siegwart%
}{%
Alonso-Mora%
\ \protect \BOthers {.}}{%
{\protect \APACyear {2013}}%
}]{%
alonso2013optimal}
\APACinsertmetastar {%
alonso2013optimal}%
\begin{APACrefauthors}%
Alonso-Mora, J.%
, Breitenmoser, A.%
, Rufli, M.%
, Beardsley, P.%
\BCBL {}\ \BBA {} Siegwart, R.%
\end{APACrefauthors}%
\unskip\
\newblock
\APACrefYearMonthDay{2013}{}{}.
\newblock
{\BBOQ}\APACrefatitle {Optimal reciprocal collision avoidance for multiple
  non-holonomic robots} {Optimal reciprocal collision avoidance for multiple
  non-holonomic robots}.{\BBCQ}
\newblock
\BIn{} \APACrefbtitle {Distributed Autonomous Robotic Systems} {Distributed
  autonomous robotic systems}\ (\BPGS\ 203--216).
\newblock
\APACaddressPublisher{}{Springer}.
\PrintBackRefs{\CurrentBib}

\bibitem [\protect \citeauthoryear {%
Alur%
, Esposito%
, Kim%
, Kumar%
\BCBL {}\ \BBA {} Lee%
}{%
Alur%
\ \protect \BOthers {.}}{%
{\protect \APACyear {1999}}%
}]{%
alur1999formal}
\APACinsertmetastar {%
alur1999formal}%
\begin{APACrefauthors}%
Alur, R.%
, Esposito, J.%
, Kim, M.%
, Kumar, V.%
\BCBL {}\ \BBA {} Lee, I.%
\end{APACrefauthors}%
\unskip\
\newblock
\APACrefYearMonthDay{1999}{}{}.
\newblock
{\BBOQ}\APACrefatitle {Formal modeling and analysis of hybrid systems: A case
  study in multi-robot coordination} {Formal modeling and analysis of hybrid
  systems: A case study in multi-robot coordination}.{\BBCQ}
\newblock
\BIn{} \APACrefbtitle {International Symposium on Formal Methods}
  {International symposium on formal methods}\ (\BPGS\ 212--232).
\PrintBackRefs{\CurrentBib}

\bibitem [\protect \citeauthoryear {%
Amodei%
\ \protect \BOthers {.}}{%
Amodei%
\ \protect \BOthers {.}}{%
{\protect \APACyear {2016}}%
}]{%
amodei2016deep}
\APACinsertmetastar {%
amodei2016deep}%
\begin{APACrefauthors}%
Amodei, D.%
, Anubhai, R.%
, Battenberg, E.%
, Case, C.%
, Casper, J.%
, Catanzaro, B.%
\BDBL {}others%
\end{APACrefauthors}%
\unskip\
\newblock
\APACrefYearMonthDay{2016}{}{}.
\newblock
{\BBOQ}\APACrefatitle {Deep Speech 2: End-to-end Speech Recognition in English
  and Mandarin} {Deep speech 2: End-to-end speech recognition in english and
  mandarin}.{\BBCQ}
\newblock
\BIn{} \APACrefbtitle {International Conference on International Conference on
  Machine Learning} {International conference on international conference on
  machine learning}\ (\BPGS\ 173--182).
\PrintBackRefs{\CurrentBib}

\bibitem [\protect \citeauthoryear {%
Balch%
\ \BBA {} Arkin%
}{%
Balch%
\ \BBA {} Arkin%
}{%
{\protect \APACyear {1998}}%
}]{%
balch1998behavior}
\APACinsertmetastar {%
balch1998behavior}%
\begin{APACrefauthors}%
Balch, T.%
\BCBT {}\ \BBA {} Arkin, R\BPBI C.%
\end{APACrefauthors}%
\unskip\
\newblock
\APACrefYearMonthDay{1998}{}{}.
\newblock
{\BBOQ}\APACrefatitle {Behavior-based formation control for multirobot teams}
  {Behavior-based formation control for multirobot teams}.{\BBCQ}
\newblock
\APACjournalVolNumPages{IEEE transactions on robotics and
  automation}{14}{6}{926--939}.
\PrintBackRefs{\CurrentBib}

\bibitem [\protect \citeauthoryear {%
Bareiss%
\ \BBA {} van~den Berg%
}{%
Bareiss%
\ \BBA {} van~den Berg%
}{%
{\protect \APACyear {2015}}%
}]{%
bareiss2015generalized}
\APACinsertmetastar {%
bareiss2015generalized}%
\begin{APACrefauthors}%
Bareiss, D.%
\BCBT {}\ \BBA {} van~den Berg, J.%
\end{APACrefauthors}%
\unskip\
\newblock
\APACrefYearMonthDay{2015}{}{}.
\newblock
{\BBOQ}\APACrefatitle {Generalized reciprocal collision avoidance} {Generalized
  reciprocal collision avoidance}.{\BBCQ}
\newblock
\APACjournalVolNumPages{The International Journal of Robotics
  Research}{34}{12}{1501--1514}.
\PrintBackRefs{\CurrentBib}

\bibitem [\protect \citeauthoryear {%
Barreto%
, Munos%
, Schaul%
\BCBL {}\ \BBA {} Silver%
}{%
Barreto%
\ \protect \BOthers {.}}{%
{\protect \APACyear {2017}}%
}]{%
Barreto:2017:SFT}
\APACinsertmetastar {%
Barreto:2017:SFT}%
\begin{APACrefauthors}%
Barreto, A.%
, Munos, R.%
, Schaul, T.%
\BCBL {}\ \BBA {} Silver, D.%
\end{APACrefauthors}%
\unskip\
\newblock
\APACrefYearMonthDay{2017}{}{}.
\newblock
{\BBOQ}\APACrefatitle {Successor Features for Transfer in Reinforcement
  Learning} {Successor features for transfer in reinforcement learning}.{\BBCQ}
\newblock
\BIn{} \APACrefbtitle {Neural Information Processing Systems (NIPS).} {Neural
  information processing systems (nips).}
\PrintBackRefs{\CurrentBib}

\bibitem [\protect \citeauthoryear {%
Bengio%
, Louradour%
, Collobert%
\BCBL {}\ \BBA {} Weston%
}{%
Bengio%
\ \protect \BOthers {.}}{%
{\protect \APACyear {2009}}%
}]{%
bengio2009curriculum}
\APACinsertmetastar {%
bengio2009curriculum}%
\begin{APACrefauthors}%
Bengio, Y.%
, Louradour, J.%
, Collobert, R.%
\BCBL {}\ \BBA {} Weston, J.%
\end{APACrefauthors}%
\unskip\
\newblock
\APACrefYearMonthDay{2009}{}{}.
\newblock
{\BBOQ}\APACrefatitle {Curriculum learning} {Curriculum learning}.{\BBCQ}
\newblock
\BIn{} \APACrefbtitle {International conference on machine learning}
  {International conference on machine learning}\ (\BPGS\ 41--48).
\PrintBackRefs{\CurrentBib}

\bibitem [\protect \citeauthoryear {%
J.~Chen%
\ \BBA {} Sun%
}{%
J.~Chen%
\ \BBA {} Sun%
}{%
{\protect \APACyear {2011}}%
}]{%
chen2011resource}
\APACinsertmetastar {%
chen2011resource}%
\begin{APACrefauthors}%
Chen, J.%
\BCBT {}\ \BBA {} Sun, D.%
\end{APACrefauthors}%
\unskip\
\newblock
\APACrefYearMonthDay{2011}{}{}.
\newblock
{\BBOQ}\APACrefatitle {Resource constrained multirobot task allocation based on
  leader--follower coalition methodology} {Resource constrained multirobot task
  allocation based on leader--follower coalition methodology}.{\BBCQ}
\newblock
\APACjournalVolNumPages{The International Journal of Robotics
  Research}{30}{12}{1423--1434}.
\PrintBackRefs{\CurrentBib}

\bibitem [\protect \citeauthoryear {%
J.~Chen%
, Sun%
, Yang%
\BCBL {}\ \BBA {} Chen%
}{%
J.~Chen%
\ \protect \BOthers {.}}{%
{\protect \APACyear {2010}}%
}]{%
chen2010leader}
\APACinsertmetastar {%
chen2010leader}%
\begin{APACrefauthors}%
Chen, J.%
, Sun, D.%
, Yang, J.%
\BCBL {}\ \BBA {} Chen, H.%
\end{APACrefauthors}%
\unskip\
\newblock
\APACrefYearMonthDay{2010}{}{}.
\newblock
{\BBOQ}\APACrefatitle {Leader-follower formation control of multiple
  non-holonomic mobile robots incorporating a receding-horizon scheme}
  {Leader-follower formation control of multiple non-holonomic mobile robots
  incorporating a receding-horizon scheme}.{\BBCQ}
\newblock
\APACjournalVolNumPages{The International Journal of Robotics
  Research}{29}{6}{727--747}.
\PrintBackRefs{\CurrentBib}

\bibitem [\protect \citeauthoryear {%
Y\BPBI F.~Chen%
, Everett%
, Liu%
\BCBL {}\ \BBA {} How%
}{%
Y\BPBI F.~Chen%
, Everett%
\BCBL {}\ \protect \BOthers {.}}{%
{\protect \APACyear {2017}}%
}]{%
chen2017socially}
\APACinsertmetastar {%
chen2017socially}%
\begin{APACrefauthors}%
Chen, Y\BPBI F.%
, Everett, M.%
, Liu, M.%
\BCBL {}\ \BBA {} How, J\BPBI P.%
\end{APACrefauthors}%
\unskip\
\newblock
\APACrefYearMonthDay{2017}{}{}.
\newblock
{\BBOQ}\APACrefatitle {Socially Aware Motion Planning with Deep Reinforcement
  Learning} {Socially aware motion planning with deep reinforcement
  learning}.{\BBCQ}
\newblock
\APACjournalVolNumPages{arXiv:1703.08862}{}{}{}.
\PrintBackRefs{\CurrentBib}

\bibitem [\protect \citeauthoryear {%
Y\BPBI F.~Chen%
, Liu%
, Everett%
\BCBL {}\ \BBA {} How%
}{%
Y\BPBI F.~Chen%
, Liu%
\BCBL {}\ \protect \BOthers {.}}{%
{\protect \APACyear {2017}}%
}]{%
chen2017decentralized}
\APACinsertmetastar {%
chen2017decentralized}%
\begin{APACrefauthors}%
Chen, Y\BPBI F.%
, Liu, M.%
, Everett, M.%
\BCBL {}\ \BBA {} How, J\BPBI P.%
\end{APACrefauthors}%
\unskip\
\newblock
\APACrefYearMonthDay{2017}{}{}.
\newblock
{\BBOQ}\APACrefatitle {Decentralized non-communicating multiagent collision
  avoidance with deep reinforcement learning} {Decentralized non-communicating
  multiagent collision avoidance with deep reinforcement learning}.{\BBCQ}
\newblock
\BIn{} \APACrefbtitle {International Conference on Robotics and Automation}
  {International conference on robotics and automation}\ (\BPGS\ 285--292).
\PrintBackRefs{\CurrentBib}

\bibitem [\protect \citeauthoryear {%
Claes%
, Hennes%
, Tuyls%
\BCBL {}\ \BBA {} Meeussen%
}{%
Claes%
\ \protect \BOthers {.}}{%
{\protect \APACyear {2012}}%
}]{%
claes2012collision}
\APACinsertmetastar {%
claes2012collision}%
\begin{APACrefauthors}%
Claes, D.%
, Hennes, D.%
, Tuyls, K.%
\BCBL {}\ \BBA {} Meeussen, W.%
\end{APACrefauthors}%
\unskip\
\newblock
\APACrefYearMonthDay{2012}{}{}.
\newblock
{\BBOQ}\APACrefatitle {Collision avoidance under bounded localization
  uncertainty} {Collision avoidance under bounded localization
  uncertainty}.{\BBCQ}
\newblock
\BIn{} \APACrefbtitle {Intelligent Robots and Systems (IROS), 2012 IEEE/RSJ
  International Conference on} {Intelligent robots and systems (iros), 2012
  ieee/rsj international conference on}\ (\BPGS\ 1192--1198).
\PrintBackRefs{\CurrentBib}

\bibitem [\protect \citeauthoryear {%
De~La~Cruz%
\ \BBA {} Carelli%
}{%
De~La~Cruz%
\ \BBA {} Carelli%
}{%
{\protect \APACyear {2008}}%
}]{%
de2008dynamic}
\APACinsertmetastar {%
de2008dynamic}%
\begin{APACrefauthors}%
De~La~Cruz, C.%
\BCBT {}\ \BBA {} Carelli, R.%
\end{APACrefauthors}%
\unskip\
\newblock
\APACrefYearMonthDay{2008}{}{}.
\newblock
{\BBOQ}\APACrefatitle {Dynamic model based formation control and obstacle
  avoidance of multi-robot systems} {Dynamic model based formation control and
  obstacle avoidance of multi-robot systems}.{\BBCQ}
\newblock
\APACjournalVolNumPages{Robotica}{26}{3}{345--356}.
\PrintBackRefs{\CurrentBib}

\bibitem [\protect \citeauthoryear {%
Egerstedt%
\ \BBA {} Hu%
}{%
Egerstedt%
\ \BBA {} Hu%
}{%
{\protect \APACyear {2002}}%
}]{%
egerstedt2002hybrid}
\APACinsertmetastar {%
egerstedt2002hybrid}%
\begin{APACrefauthors}%
Egerstedt, M.%
\BCBT {}\ \BBA {} Hu, X.%
\end{APACrefauthors}%
\unskip\
\newblock
\APACrefYearMonthDay{2002}{}{}.
\newblock
{\BBOQ}\APACrefatitle {A hybrid control approach to action coordination for
  mobile robots} {A hybrid control approach to action coordination for mobile
  robots}.{\BBCQ}
\newblock
\APACjournalVolNumPages{Automatica}{38}{1}{125--130}.
\PrintBackRefs{\CurrentBib}

\bibitem [\protect \citeauthoryear {%
Ess%
, Schindler%
, Leibe%
\BCBL {}\ \BBA {} Van~Gool%
}{%
Ess%
\ \protect \BOthers {.}}{%
{\protect \APACyear {2010}}%
}]{%
ess2010object}
\APACinsertmetastar {%
ess2010object}%
\begin{APACrefauthors}%
Ess, A.%
, Schindler, K.%
, Leibe, B.%
\BCBL {}\ \BBA {} Van~Gool, L.%
\end{APACrefauthors}%
\unskip\
\newblock
\APACrefYearMonthDay{2010}{}{}.
\newblock
{\BBOQ}\APACrefatitle {Object detection and tracking for autonomous navigation
  in dynamic environments} {Object detection and tracking for autonomous
  navigation in dynamic environments}.{\BBCQ}
\newblock
\APACjournalVolNumPages{The International Journal of Robotics
  Research}{29}{14}{1707--1725}.
\PrintBackRefs{\CurrentBib}

\bibitem [\protect \citeauthoryear {%
Everett%
, Chen%
\BCBL {}\ \BBA {} How%
}{%
Everett%
\ \protect \BOthers {.}}{%
{\protect \APACyear {2018}}%
}]{%
everett2018motion}
\APACinsertmetastar {%
everett2018motion}%
\begin{APACrefauthors}%
Everett, M.%
, Chen, Y\BPBI F.%
\BCBL {}\ \BBA {} How, J\BPBI P.%
\end{APACrefauthors}%
\unskip\
\newblock
\APACrefYearMonthDay{2018}{}{}.
\newblock
{\BBOQ}\APACrefatitle {Motion Planning Among Dynamic, Decision-Making Agents
  with Deep Reinforcement Learning} {Motion planning among dynamic,
  decision-making agents with deep reinforcement learning}.{\BBCQ}
\newblock
\APACjournalVolNumPages{arXiv preprint arXiv:1805.01956}{}{}{}.
\PrintBackRefs{\CurrentBib}

\bibitem [\protect \citeauthoryear {%
Flacco%
, Kröger%
, Luca%
\BCBL {}\ \BBA {} Khatib%
}{%
Flacco%
\ \protect \BOthers {.}}{%
{\protect \APACyear {2012}}%
}]{%
Flacco:2012:ICRA}
\APACinsertmetastar {%
Flacco:2012:ICRA}%
\begin{APACrefauthors}%
Flacco, F.%
, Kröger, T.%
, Luca, A\BPBI D.%
\BCBL {}\ \BBA {} Khatib, O.%
\end{APACrefauthors}%
\unskip\
\newblock
\APACrefYearMonthDay{2012}{}{}.
\newblock
{\BBOQ}\APACrefatitle {A depth space approach to human-robot collision
  avoidance} {A depth space approach to human-robot collision
  avoidance}.{\BBCQ}
\newblock
\BIn{} \APACrefbtitle {IEEE International Conference on Robotics and
  Automation} {Ieee international conference on robotics and automation}\
  (\BPG~338-345).
\PrintBackRefs{\CurrentBib}

\bibitem [\protect \citeauthoryear {%
Frans%
, Ho%
, Chen%
, Abbeel%
\BCBL {}\ \BBA {} Schulman%
}{%
Frans%
\ \protect \BOthers {.}}{%
{\protect \APACyear {2017}}%
}]{%
frans2017meta}
\APACinsertmetastar {%
frans2017meta}%
\begin{APACrefauthors}%
Frans, K.%
, Ho, J.%
, Chen, X.%
, Abbeel, P.%
\BCBL {}\ \BBA {} Schulman, J.%
\end{APACrefauthors}%
\unskip\
\newblock
\APACrefYearMonthDay{2017}{}{}.
\newblock
{\BBOQ}\APACrefatitle {Meta learning shared hierarchies} {Meta learning shared
  hierarchies}.{\BBCQ}
\newblock
\APACjournalVolNumPages{arXiv preprint arXiv:1710.09767}{}{}{}.
\PrintBackRefs{\CurrentBib}

\bibitem [\protect \citeauthoryear {%
Gillula%
, Hoffmann%
, Huang%
, Vitus%
\BCBL {}\ \BBA {} Tomlin%
}{%
Gillula%
\ \protect \BOthers {.}}{%
{\protect \APACyear {2011}}%
}]{%
gillula2011applications}
\APACinsertmetastar {%
gillula2011applications}%
\begin{APACrefauthors}%
Gillula, J\BPBI H.%
, Hoffmann, G\BPBI M.%
, Huang, H.%
, Vitus, M\BPBI P.%
\BCBL {}\ \BBA {} Tomlin, C\BPBI J.%
\end{APACrefauthors}%
\unskip\
\newblock
\APACrefYearMonthDay{2011}{}{}.
\newblock
{\BBOQ}\APACrefatitle {Applications of hybrid reachability analysis to robotic
  aerial vehicles} {Applications of hybrid reachability analysis to robotic
  aerial vehicles}.{\BBCQ}
\newblock
\APACjournalVolNumPages{The International Journal of Robotics
  Research}{30}{3}{335--354}.
\PrintBackRefs{\CurrentBib}

\bibitem [\protect \citeauthoryear {%
Godoy%
, Karamouzas%
, Guy%
\BCBL {}\ \BBA {} Gini%
}{%
Godoy%
\ \protect \BOthers {.}}{%
{\protect \APACyear {2016}}%
{\protect \APACexlab {{\protect \BCnt {1}}}}}]{%
godoy2016implicit}
\APACinsertmetastar {%
godoy2016implicit}%
\begin{APACrefauthors}%
Godoy, J.%
, Karamouzas, I.%
, Guy, S\BPBI J.%
\BCBL {}\ \BBA {} Gini, M.%
\end{APACrefauthors}%
\unskip\
\newblock
\APACrefYearMonthDay{2016{\protect \BCnt {1}}}{}{}.
\newblock
{\BBOQ}\APACrefatitle {Implicit Coordination in Crowded Multi-Agent Navigation}
  {Implicit coordination in crowded multi-agent navigation}.{\BBCQ}
\newblock
\BIn{} \APACrefbtitle {Thirtieth AAAI Conference on Artificial Intelligence.}
  {Thirtieth aaai conference on artificial intelligence.}
\PrintBackRefs{\CurrentBib}

\bibitem [\protect \citeauthoryear {%
Godoy%
, Karamouzas%
, Guy%
\BCBL {}\ \BBA {} Gini%
}{%
Godoy%
\ \protect \BOthers {.}}{%
{\protect \APACyear {2016}}%
{\protect \APACexlab {{\protect \BCnt {2}}}}}]{%
godoy2016moving}
\APACinsertmetastar {%
godoy2016moving}%
\begin{APACrefauthors}%
Godoy, J.%
, Karamouzas, I.%
, Guy, S\BPBI J.%
\BCBL {}\ \BBA {} Gini, M\BPBI L.%
\end{APACrefauthors}%
\unskip\
\newblock
\APACrefYearMonthDay{2016{\protect \BCnt {2}}}{}{}.
\newblock
{\BBOQ}\APACrefatitle {Moving in a Crowd: Safe and Efficient Navigation among
  Heterogeneous Agents.} {Moving in a crowd: Safe and efficient navigation
  among heterogeneous agents.}{\BBCQ}
\newblock
\BIn{} \APACrefbtitle {IJCAI} {Ijcai}\ (\BPGS\ 294--300).
\PrintBackRefs{\CurrentBib}

\bibitem [\protect \citeauthoryear {%
Graves%
, Mohamed%
\BCBL {}\ \BBA {} Hinton%
}{%
Graves%
\ \protect \BOthers {.}}{%
{\protect \APACyear {2013}}%
}]{%
graves2013speech}
\APACinsertmetastar {%
graves2013speech}%
\begin{APACrefauthors}%
Graves, A.%
, Mohamed, A\BHBI R.%
\BCBL {}\ \BBA {} Hinton, G.%
\end{APACrefauthors}%
\unskip\
\newblock
\APACrefYearMonthDay{2013}{}{}.
\newblock
{\BBOQ}\APACrefatitle {Speech recognition with deep recurrent neural networks}
  {Speech recognition with deep recurrent neural networks}.{\BBCQ}
\newblock
\BIn{} \APACrefbtitle {2013 IEEE international conference on acoustics, speech
  and signal processing} {2013 ieee international conference on acoustics,
  speech and signal processing}\ (\BPGS\ 6645--6649).
\PrintBackRefs{\CurrentBib}

\bibitem [\protect \citeauthoryear {%
Gupta%
, Johnson%
, Fei-Fei%
, Savarese%
\BCBL {}\ \BBA {} Alahi%
}{%
Gupta%
\ \protect \BOthers {.}}{%
{\protect \APACyear {2018}}%
}]{%
Gupta:2018:CVPR}
\APACinsertmetastar {%
Gupta:2018:CVPR}%
\begin{APACrefauthors}%
Gupta, A.%
, Johnson, J.%
, Fei-Fei, L.%
, Savarese, S.%
\BCBL {}\ \BBA {} Alahi, A.%
\end{APACrefauthors}%
\unskip\
\newblock
\APACrefYearMonthDay{2018}{}{}.
\newblock
{\BBOQ}\APACrefatitle {{Social GAN}: Socially Acceptable Trajectories With
  Generative Adversarial Networks} {{Social GAN}: Socially acceptable
  trajectories with generative adversarial networks}.{\BBCQ}
\newblock
\BIn{} \APACrefbtitle {IEEE Conference on Computer Vision and Pattern
  Recognition.} {Ieee conference on computer vision and pattern recognition.}
\PrintBackRefs{\CurrentBib}

\bibitem [\protect \citeauthoryear {%
He%
, Zhang%
, Ren%
\BCBL {}\ \BBA {} Sun%
}{%
He%
\ \protect \BOthers {.}}{%
{\protect \APACyear {2016}}%
}]{%
he2016deep}
\APACinsertmetastar {%
he2016deep}%
\begin{APACrefauthors}%
He, K.%
, Zhang, X.%
, Ren, S.%
\BCBL {}\ \BBA {} Sun, J.%
\end{APACrefauthors}%
\unskip\
\newblock
\APACrefYearMonthDay{2016}{}{}.
\newblock
{\BBOQ}\APACrefatitle {Deep Residual Learning for Image Recognition} {Deep
  residual learning for image recognition}.{\BBCQ}
\newblock
\BIn{} \APACrefbtitle {IEEE Conference on Computer Vision and Pattern
  Recognition.} {Ieee conference on computer vision and pattern recognition.}
\PrintBackRefs{\CurrentBib}

\bibitem [\protect \citeauthoryear {%
Heess%
\ \protect \BOthers {.}}{%
Heess%
\ \protect \BOthers {.}}{%
{\protect \APACyear {2017}}%
}]{%
heess2017emergence}
\APACinsertmetastar {%
heess2017emergence}%
\begin{APACrefauthors}%
Heess, N.%
, Sriram, S.%
, Lemmon, J.%
, Merel, J.%
, Wayne, G.%
, Tassa, Y.%
\BDBL {}others%
\end{APACrefauthors}%
\unskip\
\newblock
\APACrefYearMonthDay{2017}{}{}.
\newblock
{\BBOQ}\APACrefatitle {Emergence of Locomotion Behaviours in Rich Environments}
  {Emergence of locomotion behaviours in rich environments}.{\BBCQ}
\newblock
\APACjournalVolNumPages{arXiv:1707.02286}{}{}{}.
\PrintBackRefs{\CurrentBib}

\bibitem [\protect \citeauthoryear {%
Hennes%
, Claes%
, Meeussen%
\BCBL {}\ \BBA {} Tuyls%
}{%
Hennes%
\ \protect \BOthers {.}}{%
{\protect \APACyear {2012}}%
}]{%
hennes2012multi}
\APACinsertmetastar {%
hennes2012multi}%
\begin{APACrefauthors}%
Hennes, D.%
, Claes, D.%
, Meeussen, W.%
\BCBL {}\ \BBA {} Tuyls, K.%
\end{APACrefauthors}%
\unskip\
\newblock
\APACrefYearMonthDay{2012}{}{}.
\newblock
{\BBOQ}\APACrefatitle {Multi-robot collision avoidance with localization
  uncertainty} {Multi-robot collision avoidance with localization
  uncertainty}.{\BBCQ}
\newblock
\BIn{} \APACrefbtitle {International Conference on Autonomous Agents and
  Multiagent Systems-Volume 1} {International conference on autonomous agents
  and multiagent systems-volume 1}\ (\BPGS\ 147--154).
\PrintBackRefs{\CurrentBib}

\bibitem [\protect \citeauthoryear {%
Hu%
\ \BBA {} Sun%
}{%
Hu%
\ \BBA {} Sun%
}{%
{\protect \APACyear {2011}}%
}]{%
hu2011automatic}
\APACinsertmetastar {%
hu2011automatic}%
\begin{APACrefauthors}%
Hu, S.%
\BCBT {}\ \BBA {} Sun, D.%
\end{APACrefauthors}%
\unskip\
\newblock
\APACrefYearMonthDay{2011}{}{}.
\newblock
{\BBOQ}\APACrefatitle {Automatic transportation of biological cells with a
  robot-tweezer manipulation system} {Automatic transportation of biological
  cells with a robot-tweezer manipulation system}.{\BBCQ}
\newblock
\APACjournalVolNumPages{The International Journal of Robotics
  Research}{30}{14}{1681--1694}.
\PrintBackRefs{\CurrentBib}

\bibitem [\protect \citeauthoryear {%
Kahn%
, Villaflor%
, Pong%
, Abbeel%
\BCBL {}\ \BBA {} Levine%
}{%
Kahn%
\ \protect \BOthers {.}}{%
{\protect \APACyear {2017}}%
}]{%
kahn2017uncertainty}
\APACinsertmetastar {%
kahn2017uncertainty}%
\begin{APACrefauthors}%
Kahn, G.%
, Villaflor, A.%
, Pong, V.%
, Abbeel, P.%
\BCBL {}\ \BBA {} Levine, S.%
\end{APACrefauthors}%
\unskip\
\newblock
\APACrefYearMonthDay{2017}{}{}.
\newblock
{\BBOQ}\APACrefatitle {Uncertainty-Aware Reinforcement Learning for Collision
  Avoidance} {Uncertainty-aware reinforcement learning for collision
  avoidance}.{\BBCQ}
\newblock
\APACjournalVolNumPages{arXiv:1702.01182}{}{}{}.
\PrintBackRefs{\CurrentBib}

\bibitem [\protect \citeauthoryear {%
Kim%
\ \protect \BOthers {.}}{%
Kim%
\ \protect \BOthers {.}}{%
{\protect \APACyear {2015}}%
}]{%
Kim:2015:BRVO}
\APACinsertmetastar {%
Kim:2015:BRVO}%
\begin{APACrefauthors}%
Kim, S.%
, Guy, S\BPBI J.%
, Liu, W.%
, Wilkie, D.%
, Lau, R\BPBI W\BPBI H.%
, Lin, M\BPBI C.%
\BCBL {}\ \BBA {} Manocha, D.%
\end{APACrefauthors}%
\unskip\
\newblock
\APACrefYearMonthDay{2015}{}{}.
\newblock
{\BBOQ}\APACrefatitle {{BRVO:} Predicting pedestrian trajectories using
  velocity-space reasoning} {{BRVO:} predicting pedestrian trajectories using
  velocity-space reasoning}.{\BBCQ}
\newblock
\APACjournalVolNumPages{International Journal of Robotics
  Research}{34}{2}{201--217}.
\PrintBackRefs{\CurrentBib}

\bibitem [\protect \citeauthoryear {%
Kingma%
\ \BBA {} Ba%
}{%
Kingma%
\ \BBA {} Ba%
}{%
{\protect \APACyear {2014}}%
}]{%
kingma2014adam}
\APACinsertmetastar {%
kingma2014adam}%
\begin{APACrefauthors}%
Kingma, D.%
\BCBT {}\ \BBA {} Ba, J.%
\end{APACrefauthors}%
\unskip\
\newblock
\APACrefYearMonthDay{2014}{}{}.
\newblock
{\BBOQ}\APACrefatitle {Adam: A method for stochastic optimization} {Adam: A
  method for stochastic optimization}.{\BBCQ}
\newblock
\APACjournalVolNumPages{arXiv:1412.6980}{}{}{}.
\PrintBackRefs{\CurrentBib}

\bibitem [\protect \citeauthoryear {%
Krizhevsky%
, Sutskever%
\BCBL {}\ \BBA {} Hinton%
}{%
Krizhevsky%
\ \protect \BOthers {.}}{%
{\protect \APACyear {2012}}%
}]{%
krizhevsky2012imagenet}
\APACinsertmetastar {%
krizhevsky2012imagenet}%
\begin{APACrefauthors}%
Krizhevsky, A.%
, Sutskever, I.%
\BCBL {}\ \BBA {} Hinton, G\BPBI E.%
\end{APACrefauthors}%
\unskip\
\newblock
\APACrefYearMonthDay{2012}{}{}.
\newblock
{\BBOQ}\APACrefatitle {Imagenet classification with deep convolutional neural
  networks} {Imagenet classification with deep convolutional neural
  networks}.{\BBCQ}
\newblock
\BIn{} \APACrefbtitle {Advances in Neural Information Processing Systems}
  {Advances in neural information processing systems}\ (\BPGS\ 1097--1105).
\PrintBackRefs{\CurrentBib}

\bibitem [\protect \citeauthoryear {%
Lenz%
, Knepper%
\BCBL {}\ \BBA {} Saxena%
}{%
Lenz%
\ \protect \BOthers {.}}{%
{\protect \APACyear {2015}}%
}]{%
lenzdeepMPC2015}
\APACinsertmetastar {%
lenzdeepMPC2015}%
\begin{APACrefauthors}%
Lenz, I.%
, Knepper, R.%
\BCBL {}\ \BBA {} Saxena, A.%
\end{APACrefauthors}%
\unskip\
\newblock
\APACrefYearMonthDay{2015}{}{}.
\newblock
{\BBOQ}\APACrefatitle {DeepMPC: Learning Deep Latent Features for Model
  Predictive Control} {Deepmpc: Learning deep latent features for model
  predictive control}.{\BBCQ}
\newblock
\BIn{} \APACrefbtitle {RSS.} {Rss.}
\PrintBackRefs{\CurrentBib}

\bibitem [\protect \citeauthoryear {%
Levine%
, Finn%
, Darrell%
\BCBL {}\ \BBA {} Abbeel%
}{%
Levine%
\ \protect \BOthers {.}}{%
{\protect \APACyear {2016}}%
}]{%
levine2016end}
\APACinsertmetastar {%
levine2016end}%
\begin{APACrefauthors}%
Levine, S.%
, Finn, C.%
, Darrell, T.%
\BCBL {}\ \BBA {} Abbeel, P.%
\end{APACrefauthors}%
\unskip\
\newblock
\APACrefYearMonthDay{2016}{}{}.
\newblock
{\BBOQ}\APACrefatitle {End-to-end training of deep visuomotor policies}
  {End-to-end training of deep visuomotor policies}.{\BBCQ}
\newblock
\APACjournalVolNumPages{Journal of Machine Learning Research}{17}{39}{1--40}.
\PrintBackRefs{\CurrentBib}

\bibitem [\protect \citeauthoryear {%
Long%
, Fan%
\BCBL {}\ \protect \BOthers {.}}{%
Long%
, Fan%
\BCBL {}\ \protect \BOthers {.}}{%
{\protect \APACyear {2017}}%
}]{%
long2017towards}
\APACinsertmetastar {%
long2017towards}%
\begin{APACrefauthors}%
Long, P.%
, Fan, T.%
, Liao, X.%
, Liu, W.%
, Zhang, H.%
\BCBL {}\ \BBA {} Pan, J.%
\end{APACrefauthors}%
\unskip\
\newblock
\APACrefYearMonthDay{2017}{}{}.
\newblock
{\BBOQ}\APACrefatitle {Towards Optimally Decentralized Multi-Robot Collision
  Avoidance via Deep Reinforcement Learning} {Towards optimally decentralized
  multi-robot collision avoidance via deep reinforcement learning}.{\BBCQ}
\newblock
\BIn{} \APACrefbtitle {IEEE International Conference on Robotics and
  Automation.} {Ieee international conference on robotics and automation.}
\PrintBackRefs{\CurrentBib}

\bibitem [\protect \citeauthoryear {%
Long%
, Liu%
\BCBL {}\ \BBA {} Pan%
}{%
Long%
, Liu%
\BCBL {}\ \BBA {} Pan%
}{%
{\protect \APACyear {2017}}%
}]{%
long2017deep}
\APACinsertmetastar {%
long2017deep}%
\begin{APACrefauthors}%
Long, P.%
, Liu, W.%
\BCBL {}\ \BBA {} Pan, J.%
\end{APACrefauthors}%
\unskip\
\newblock
\APACrefYearMonthDay{2017}{}{}.
\newblock
{\BBOQ}\APACrefatitle {Deep-Learned Collision Avoidance Policy for Distributed
  Multiagent Navigation} {Deep-learned collision avoidance policy for
  distributed multiagent navigation}.{\BBCQ}
\newblock
\APACjournalVolNumPages{Robotics and Automation Letters}{2}{2}{656--663}.
\PrintBackRefs{\CurrentBib}

\bibitem [\protect \citeauthoryear {%
Luna%
\ \BBA {} Bekris%
}{%
Luna%
\ \BBA {} Bekris%
}{%
{\protect \APACyear {2011}}%
}]{%
luna2011efficient}
\APACinsertmetastar {%
luna2011efficient}%
\begin{APACrefauthors}%
Luna, R.%
\BCBT {}\ \BBA {} Bekris, K\BPBI E.%
\end{APACrefauthors}%
\unskip\
\newblock
\APACrefYearMonthDay{2011}{}{}.
\newblock
{\BBOQ}\APACrefatitle {Efficient and complete centralized multi-robot path
  planning} {Efficient and complete centralized multi-robot path
  planning}.{\BBCQ}
\newblock
\BIn{} \APACrefbtitle {Intelligent Robots and Systems (IROS), 2011 IEEE/RSJ
  International Conference on} {Intelligent robots and systems (iros), 2011
  ieee/rsj international conference on}\ (\BPGS\ 3268--3275).
\PrintBackRefs{\CurrentBib}

\bibitem [\protect \citeauthoryear {%
Michael%
, Fink%
\BCBL {}\ \BBA {} Kumar%
}{%
Michael%
\ \protect \BOthers {.}}{%
{\protect \APACyear {2011}}%
}]{%
michael2011cooperative}
\APACinsertmetastar {%
michael2011cooperative}%
\begin{APACrefauthors}%
Michael, N.%
, Fink, J.%
\BCBL {}\ \BBA {} Kumar, V.%
\end{APACrefauthors}%
\unskip\
\newblock
\APACrefYearMonthDay{2011}{}{}.
\newblock
{\BBOQ}\APACrefatitle {Cooperative manipulation and transportation with aerial
  robots} {Cooperative manipulation and transportation with aerial
  robots}.{\BBCQ}
\newblock
\APACjournalVolNumPages{Autonomous Robots}{30}{1}{73--86}.
\PrintBackRefs{\CurrentBib}

\bibitem [\protect \citeauthoryear {%
Muller%
, Ben%
, Cosatto%
, Flepp%
\BCBL {}\ \BBA {} Cun%
}{%
Muller%
\ \protect \BOthers {.}}{%
{\protect \APACyear {2006}}%
}]{%
muller2006off}
\APACinsertmetastar {%
muller2006off}%
\begin{APACrefauthors}%
Muller, U.%
, Ben, J.%
, Cosatto, E.%
, Flepp, B.%
\BCBL {}\ \BBA {} Cun, Y\BPBI L.%
\end{APACrefauthors}%
\unskip\
\newblock
\APACrefYearMonthDay{2006}{}{}.
\newblock
{\BBOQ}\APACrefatitle {Off-road obstacle avoidance through end-to-end learning}
  {Off-road obstacle avoidance through end-to-end learning}.{\BBCQ}
\newblock
\BIn{} \APACrefbtitle {Advances in neural information processing systems}
  {Advances in neural information processing systems}\ (\BPGS\ 739--746).
\PrintBackRefs{\CurrentBib}

\bibitem [\protect \citeauthoryear {%
Nair%
\ \BBA {} Hinton%
}{%
Nair%
\ \BBA {} Hinton%
}{%
{\protect \APACyear {2010}}%
}]{%
nair2010rectified}
\APACinsertmetastar {%
nair2010rectified}%
\begin{APACrefauthors}%
Nair, V.%
\BCBT {}\ \BBA {} Hinton, G\BPBI E.%
\end{APACrefauthors}%
\unskip\
\newblock
\APACrefYearMonthDay{2010}{}{}.
\newblock
{\BBOQ}\APACrefatitle {Rectified linear units improve restricted boltzmann
  machines} {Rectified linear units improve restricted boltzmann
  machines}.{\BBCQ}
\newblock
\BIn{} \APACrefbtitle {International conference on machine learning}
  {International conference on machine learning}\ (\BPGS\ 807--814).
\PrintBackRefs{\CurrentBib}

\bibitem [\protect \citeauthoryear {%
Ondruska%
, Dequaire%
, Zeng~Wang%
\BCBL {}\ \BBA {} Posner%
}{%
Ondruska%
\ \protect \BOthers {.}}{%
{\protect \APACyear {2016}}%
}]{%
OndruskaRSS2016}
\APACinsertmetastar {%
OndruskaRSS2016}%
\begin{APACrefauthors}%
Ondruska, P.%
, Dequaire, J.%
, Zeng~Wang, D.%
\BCBL {}\ \BBA {} Posner, I.%
\end{APACrefauthors}%
\unskip\
\newblock
\APACrefYearMonthDay{2016}{}{}.
\newblock
{\BBOQ}\APACrefatitle {{End-to-End Tracking and Semantic Segmentation Using
  Recurrent Neural Networks}} {{End-to-End Tracking and Semantic Segmentation
  Using Recurrent Neural Networks}}.{\BBCQ}
\newblock
\BIn{} \APACrefbtitle {Robotics: Science and Systems, Workshop on Limits and
  Potentials of Deep Learning in Robotics.} {Robotics: Science and systems,
  workshop on limits and potentials of deep learning in robotics.}
\PrintBackRefs{\CurrentBib}

\bibitem [\protect \citeauthoryear {%
Ondruska%
\ \BBA {} Posner%
}{%
Ondruska%
\ \BBA {} Posner%
}{%
{\protect \APACyear {2016}}%
}]{%
OndruskaAAAI2016}
\APACinsertmetastar {%
OndruskaAAAI2016}%
\begin{APACrefauthors}%
Ondruska, P.%
\BCBT {}\ \BBA {} Posner, I.%
\end{APACrefauthors}%
\unskip\
\newblock
\APACrefYearMonthDay{2016}{}{}.
\newblock
{\BBOQ}\APACrefatitle {Deep Tracking: Seeing Beyond Seeing Using Recurrent
  Neural Networks} {Deep tracking: Seeing beyond seeing using recurrent neural
  networks}.{\BBCQ}
\newblock
\BIn{} \APACrefbtitle {AAAI Conference on Artificial Intelligence.} {Aaai
  conference on artificial intelligence.}
\newblock
\APACaddressPublisher{Phoenix, Arizona USA}{}.
\PrintBackRefs{\CurrentBib}

\bibitem [\protect \citeauthoryear {%
Peng%
, Andrychowicz%
, Zaremba%
\BCBL {}\ \BBA {} Abbeel%
}{%
Peng%
\ \protect \BOthers {.}}{%
{\protect \APACyear {2017}}%
}]{%
peng2017sim}
\APACinsertmetastar {%
peng2017sim}%
\begin{APACrefauthors}%
Peng, X\BPBI B.%
, Andrychowicz, M.%
, Zaremba, W.%
\BCBL {}\ \BBA {} Abbeel, P.%
\end{APACrefauthors}%
\unskip\
\newblock
\APACrefYearMonthDay{2017}{}{}.
\newblock
{\BBOQ}\APACrefatitle {Sim-to-real transfer of robotic control with dynamics
  randomization} {Sim-to-real transfer of robotic control with dynamics
  randomization}.{\BBCQ}
\newblock
\APACjournalVolNumPages{arXiv:1710.06537}{}{}{}.
\PrintBackRefs{\CurrentBib}

\bibitem [\protect \citeauthoryear {%
Pfeiffer%
, Schaeuble%
, Nieto%
, Siegwart%
\BCBL {}\ \BBA {} Cadena%
}{%
Pfeiffer%
\ \protect \BOthers {.}}{%
{\protect \APACyear {2017}}%
}]{%
pfeiffer2017perception}
\APACinsertmetastar {%
pfeiffer2017perception}%
\begin{APACrefauthors}%
Pfeiffer, M.%
, Schaeuble, M.%
, Nieto, J.%
, Siegwart, R.%
\BCBL {}\ \BBA {} Cadena, C.%
\end{APACrefauthors}%
\unskip\
\newblock
\APACrefYearMonthDay{2017}{}{}.
\newblock
{\BBOQ}\APACrefatitle {From perception to decision: A data-driven approach to
  end-to-end motion planning for autonomous ground robots} {From perception to
  decision: A data-driven approach to end-to-end motion planning for autonomous
  ground robots}.{\BBCQ}
\newblock
\BIn{} \APACrefbtitle {International Conference on Robotics and Automation}
  {International conference on robotics and automation}\ (\BPGS\ 1527--1533).
\PrintBackRefs{\CurrentBib}

\bibitem [\protect \citeauthoryear {%
Quottrup%
, Bak%
\BCBL {}\ \BBA {} Zamanabadi%
}{%
Quottrup%
\ \protect \BOthers {.}}{%
{\protect \APACyear {2004}}%
}]{%
quottrup2004multi}
\APACinsertmetastar {%
quottrup2004multi}%
\begin{APACrefauthors}%
Quottrup, M\BPBI M.%
, Bak, T.%
\BCBL {}\ \BBA {} Zamanabadi, R.%
\end{APACrefauthors}%
\unskip\
\newblock
\APACrefYearMonthDay{2004}{}{}.
\newblock
{\BBOQ}\APACrefatitle {Multi-robot planning: A timed automata approach}
  {Multi-robot planning: A timed automata approach}.{\BBCQ}
\newblock
\BIn{} \APACrefbtitle {Robotics and Automation, 2004. Proceedings. ICRA'04.
  2004 IEEE International Conference on} {Robotics and automation, 2004.
  proceedings. icra'04. 2004 ieee international conference on}\ (\BVOL~5,
  \BPGS\ 4417--4422).
\PrintBackRefs{\CurrentBib}

\bibitem [\protect \citeauthoryear {%
Ross%
\ \protect \BOthers {.}}{%
Ross%
\ \protect \BOthers {.}}{%
{\protect \APACyear {2013}}%
}]{%
ross2013learning}
\APACinsertmetastar {%
ross2013learning}%
\begin{APACrefauthors}%
Ross, S.%
, Melik-Barkhudarov, N.%
, Shankar, K\BPBI S.%
, Wendel, A.%
, Dey, D.%
, Bagnell, J\BPBI A.%
\BCBL {}\ \BBA {} Hebert, M.%
\end{APACrefauthors}%
\unskip\
\newblock
\APACrefYearMonthDay{2013}{}{}.
\newblock
{\BBOQ}\APACrefatitle {Learning monocular reactive uav control in cluttered
  natural environments} {Learning monocular reactive uav control in cluttered
  natural environments}.{\BBCQ}
\newblock
\BIn{} \APACrefbtitle {International Conference on Robotics and Automation}
  {International conference on robotics and automation}\ (\BPGS\ 1765--1772).
\PrintBackRefs{\CurrentBib}

\bibitem [\protect \citeauthoryear {%
Rusu%
\ \protect \BOthers {.}}{%
Rusu%
\ \protect \BOthers {.}}{%
{\protect \APACyear {2016}}%
}]{%
rusu2016sim}
\APACinsertmetastar {%
rusu2016sim}%
\begin{APACrefauthors}%
Rusu, A\BPBI A.%
, Vecerik, M.%
, Roth{\"o}rl, T.%
, Heess, N.%
, Pascanu, R.%
\BCBL {}\ \BBA {} Hadsell, R.%
\end{APACrefauthors}%
\unskip\
\newblock
\APACrefYearMonthDay{2016}{}{}.
\newblock
{\BBOQ}\APACrefatitle {Sim-to-real robot learning from pixels with progressive
  nets} {Sim-to-real robot learning from pixels with progressive nets}.{\BBCQ}
\newblock
\BIn{} \APACrefbtitle {Conference on Robot Learning.} {Conference on robot
  learning.}
\PrintBackRefs{\CurrentBib}

\bibitem [\protect \citeauthoryear {%
Sadeghi%
\ \BBA {} Levine%
}{%
Sadeghi%
\ \BBA {} Levine%
}{%
{\protect \APACyear {2017}}%
}]{%
sadeghi2016cad2rl}
\APACinsertmetastar {%
sadeghi2016cad2rl}%
\begin{APACrefauthors}%
Sadeghi, F.%
\BCBT {}\ \BBA {} Levine, S.%
\end{APACrefauthors}%
\unskip\
\newblock
\APACrefYearMonthDay{2017}{}{}.
\newblock
{\BBOQ}\APACrefatitle {Cad2rl: Real single-image flight without a single real
  image} {Cad2rl: Real single-image flight without a single real image}.{\BBCQ}
\newblock
\BIn{} \APACrefbtitle {Robotics: Science and System.} {Robotics: Science and
  system.}
\PrintBackRefs{\CurrentBib}

\bibitem [\protect \citeauthoryear {%
Schulman%
, Levine%
, Abbeel%
, Jordan%
\BCBL {}\ \BBA {} Moritz%
}{%
Schulman%
, Levine%
\BCBL {}\ \protect \BOthers {.}}{%
{\protect \APACyear {2015}}%
}]{%
trpo}
\APACinsertmetastar {%
trpo}%
\begin{APACrefauthors}%
Schulman, J.%
, Levine, S.%
, Abbeel, P.%
, Jordan, M.%
\BCBL {}\ \BBA {} Moritz, P.%
\end{APACrefauthors}%
\unskip\
\newblock
\APACrefYearMonthDay{2015}{}{}.
\newblock
{\BBOQ}\APACrefatitle {Trust region policy optimization} {Trust region policy
  optimization}.{\BBCQ}
\newblock
\BIn{} \APACrefbtitle {International Conference on Machine Learning}
  {International conference on machine learning}\ (\BPGS\ 1889--1897).
\PrintBackRefs{\CurrentBib}

\bibitem [\protect \citeauthoryear {%
Schulman%
, Moritz%
, Levine%
, Jordan%
\BCBL {}\ \BBA {} Abbeel%
}{%
Schulman%
, Moritz%
\BCBL {}\ \protect \BOthers {.}}{%
{\protect \APACyear {2015}}%
}]{%
schulman2015high}
\APACinsertmetastar {%
schulman2015high}%
\begin{APACrefauthors}%
Schulman, J.%
, Moritz, P.%
, Levine, S.%
, Jordan, M.%
\BCBL {}\ \BBA {} Abbeel, P.%
\end{APACrefauthors}%
\unskip\
\newblock
\APACrefYearMonthDay{2015}{}{}.
\newblock
{\BBOQ}\APACrefatitle {High-dimensional continuous control using generalized
  advantage estimation} {High-dimensional continuous control using generalized
  advantage estimation}.{\BBCQ}
\newblock
\APACjournalVolNumPages{arXiv:1506.02438}{}{}{}.
\PrintBackRefs{\CurrentBib}

\bibitem [\protect \citeauthoryear {%
Schulman%
, Wolski%
, Dhariwal%
, Radford%
\BCBL {}\ \BBA {} Klimov%
}{%
Schulman%
\ \protect \BOthers {.}}{%
{\protect \APACyear {2017}}%
}]{%
schulman2017proximal}
\APACinsertmetastar {%
schulman2017proximal}%
\begin{APACrefauthors}%
Schulman, J.%
, Wolski, F.%
, Dhariwal, P.%
, Radford, A.%
\BCBL {}\ \BBA {} Klimov, O.%
\end{APACrefauthors}%
\unskip\
\newblock
\APACrefYearMonthDay{2017}{}{}.
\newblock
{\BBOQ}\APACrefatitle {Proximal Policy Optimization Algorithms} {Proximal
  policy optimization algorithms}.{\BBCQ}
\newblock
\APACjournalVolNumPages{arXiv:1707.06347}{}{}{}.
\PrintBackRefs{\CurrentBib}

\bibitem [\protect \citeauthoryear {%
Schwartz%
\ \BBA {} Sharir%
}{%
Schwartz%
\ \BBA {} Sharir%
}{%
{\protect \APACyear {1983}}%
}]{%
schwartz1983piano}
\APACinsertmetastar {%
schwartz1983piano}%
\begin{APACrefauthors}%
Schwartz, J\BPBI T.%
\BCBT {}\ \BBA {} Sharir, M.%
\end{APACrefauthors}%
\unskip\
\newblock
\APACrefYearMonthDay{1983}{}{}.
\newblock
{\BBOQ}\APACrefatitle {On the piano movers' problem: III. Coordinating the
  motion of several independent bodies: The special case of circular bodies
  moving amidst polygonal barriers} {On the piano movers' problem: Iii.
  coordinating the motion of several independent bodies: The special case of
  circular bodies moving amidst polygonal barriers}.{\BBCQ}
\newblock
\APACjournalVolNumPages{The International Journal of Robotics
  Research}{2}{3}{46--75}.
\PrintBackRefs{\CurrentBib}

\bibitem [\protect \citeauthoryear {%
Sergeant%
, S{\"u}nderhauf%
, Milford%
\BCBL {}\ \BBA {} Upcroft%
}{%
Sergeant%
\ \protect \BOthers {.}}{%
{\protect \APACyear {2015}}%
}]{%
sergeant2015multimodal}
\APACinsertmetastar {%
sergeant2015multimodal}%
\begin{APACrefauthors}%
Sergeant, J.%
, S{\"u}nderhauf, N.%
, Milford, M.%
\BCBL {}\ \BBA {} Upcroft, B.%
\end{APACrefauthors}%
\unskip\
\newblock
\APACrefYearMonthDay{2015}{}{}.
\newblock
{\BBOQ}\APACrefatitle {Multimodal deep autoencoders for control of a mobile
  robot} {Multimodal deep autoencoders for control of a mobile robot}.{\BBCQ}
\newblock
\BIn{} \APACrefbtitle {Australasian Conference on Robotics and Automation.}
  {Australasian conference on robotics and automation.}
\PrintBackRefs{\CurrentBib}

\bibitem [\protect \citeauthoryear {%
Sharon%
, Stern%
, Felner%
\BCBL {}\ \BBA {} Sturtevant%
}{%
Sharon%
\ \protect \BOthers {.}}{%
{\protect \APACyear {2015}}%
}]{%
sharon2015conflict}
\APACinsertmetastar {%
sharon2015conflict}%
\begin{APACrefauthors}%
Sharon, G.%
, Stern, R.%
, Felner, A.%
\BCBL {}\ \BBA {} Sturtevant, N\BPBI R.%
\end{APACrefauthors}%
\unskip\
\newblock
\APACrefYearMonthDay{2015}{}{}.
\newblock
{\BBOQ}\APACrefatitle {Conflict-based search for optimal multi-agent
  pathfinding} {Conflict-based search for optimal multi-agent
  pathfinding}.{\BBCQ}
\newblock
\APACjournalVolNumPages{Artificial Intelligence}{219}{}{40--66}.
\PrintBackRefs{\CurrentBib}

\bibitem [\protect \citeauthoryear {%
Shucker%
, Murphey%
\BCBL {}\ \BBA {} Bennett%
}{%
Shucker%
\ \protect \BOthers {.}}{%
{\protect \APACyear {2007}}%
}]{%
shucker2007switching}
\APACinsertmetastar {%
shucker2007switching}%
\begin{APACrefauthors}%
Shucker, B.%
, Murphey, T.%
\BCBL {}\ \BBA {} Bennett, J\BPBI K.%
\end{APACrefauthors}%
\unskip\
\newblock
\APACrefYearMonthDay{2007}{}{}.
\newblock
{\BBOQ}\APACrefatitle {Switching rules for decentralized control with simple
  control laws} {Switching rules for decentralized control with simple control
  laws}.{\BBCQ}
\newblock
\BIn{} \APACrefbtitle {American Control Conference, 2007. ACC'07} {American
  control conference, 2007. acc'07}\ (\BPGS\ 1485--1492).
\PrintBackRefs{\CurrentBib}

\bibitem [\protect \citeauthoryear {%
Snape%
, Van Den~Berg%
, Guy%
\BCBL {}\ \BBA {} Manocha%
}{%
Snape%
\ \protect \BOthers {.}}{%
{\protect \APACyear {2010}}%
}]{%
snape2010smooth}
\APACinsertmetastar {%
snape2010smooth}%
\begin{APACrefauthors}%
Snape, J.%
, Van Den~Berg, J.%
, Guy, S\BPBI J.%
\BCBL {}\ \BBA {} Manocha, D.%
\end{APACrefauthors}%
\unskip\
\newblock
\APACrefYearMonthDay{2010}{}{}.
\newblock
{\BBOQ}\APACrefatitle {Smooth and collision-free navigation for multiple robots
  under differential-drive constraints} {Smooth and collision-free navigation
  for multiple robots under differential-drive constraints}.{\BBCQ}
\newblock
\BIn{} \APACrefbtitle {International Conference on Intelligent Robots and
  Systems} {International conference on intelligent robots and systems}\
  (\BPGS\ 4584--4589).
\PrintBackRefs{\CurrentBib}

\bibitem [\protect \citeauthoryear {%
Snape%
, van~den Berg%
, Guy%
\BCBL {}\ \BBA {} Manocha%
}{%
Snape%
\ \protect \BOthers {.}}{%
{\protect \APACyear {2011}}%
}]{%
snape2011hybrid}
\APACinsertmetastar {%
snape2011hybrid}%
\begin{APACrefauthors}%
Snape, J.%
, van~den Berg, J.%
, Guy, S\BPBI J.%
\BCBL {}\ \BBA {} Manocha, D.%
\end{APACrefauthors}%
\unskip\
\newblock
\APACrefYearMonthDay{2011}{}{}.
\newblock
{\BBOQ}\APACrefatitle {The hybrid reciprocal velocity obstacle} {The hybrid
  reciprocal velocity obstacle}.{\BBCQ}
\newblock
\APACjournalVolNumPages{Transactions on Robotics}{27}{4}{696--706}.
\PrintBackRefs{\CurrentBib}

\bibitem [\protect \citeauthoryear {%
Stone%
\ \BBA {} Veloso%
}{%
Stone%
\ \BBA {} Veloso%
}{%
{\protect \APACyear {1999}}%
}]{%
stone1999task}
\APACinsertmetastar {%
stone1999task}%
\begin{APACrefauthors}%
Stone, P.%
\BCBT {}\ \BBA {} Veloso, M.%
\end{APACrefauthors}%
\unskip\
\newblock
\APACrefYearMonthDay{1999}{}{}.
\newblock
{\BBOQ}\APACrefatitle {Task decomposition, dynamic role assignment, and
  low-bandwidth communication for real-time strategic teamwork} {Task
  decomposition, dynamic role assignment, and low-bandwidth communication for
  real-time strategic teamwork}.{\BBCQ}
\newblock
\APACjournalVolNumPages{Artificial Intelligence}{110}{2}{241--273}.
\PrintBackRefs{\CurrentBib}

\bibitem [\protect \citeauthoryear {%
Sun%
, Wang%
, Shang%
\BCBL {}\ \BBA {} Feng%
}{%
Sun%
\ \protect \BOthers {.}}{%
{\protect \APACyear {2009}}%
}]{%
sun2009synchronization}
\APACinsertmetastar {%
sun2009synchronization}%
\begin{APACrefauthors}%
Sun, D.%
, Wang, C.%
, Shang, W.%
\BCBL {}\ \BBA {} Feng, G.%
\end{APACrefauthors}%
\unskip\
\newblock
\APACrefYearMonthDay{2009}{}{}.
\newblock
{\BBOQ}\APACrefatitle {A synchronization approach to trajectory tracking of
  multiple mobile robots while maintaining time-varying formations} {A
  synchronization approach to trajectory tracking of multiple mobile robots
  while maintaining time-varying formations}.{\BBCQ}
\newblock
\APACjournalVolNumPages{IEEE Transactions on Robotics}{25}{5}{1074--1086}.
\PrintBackRefs{\CurrentBib}

\bibitem [\protect \citeauthoryear {%
Tai%
, Paolo%
\BCBL {}\ \BBA {} Liu%
}{%
Tai%
\ \protect \BOthers {.}}{%
{\protect \APACyear {2017}}%
}]{%
tai2017virtual}
\APACinsertmetastar {%
tai2017virtual}%
\begin{APACrefauthors}%
Tai, L.%
, Paolo, G.%
\BCBL {}\ \BBA {} Liu, M.%
\end{APACrefauthors}%
\unskip\
\newblock
\APACrefYearMonthDay{2017}{}{}.
\newblock
{\BBOQ}\APACrefatitle {Virtual-to-real Deep Reinforcement Learning: Continuous
  Control of Mobile Robots for Mapless Navigation} {Virtual-to-real deep
  reinforcement learning: Continuous control of mobile robots for mapless
  navigation}.{\BBCQ}
\newblock
\BIn{} \APACrefbtitle {International Conference on Intelligent Robots and
  Systems.} {International conference on intelligent robots and systems.}
\PrintBackRefs{\CurrentBib}

\bibitem [\protect \citeauthoryear {%
Tang%
, Thomas%
\BCBL {}\ \BBA {} Kumar%
}{%
Tang%
\ \protect \BOthers {.}}{%
{\protect \APACyear {2018}}%
}]{%
tang2018hold}
\APACinsertmetastar {%
tang2018hold}%
\begin{APACrefauthors}%
Tang, S.%
, Thomas, J.%
\BCBL {}\ \BBA {} Kumar, V.%
\end{APACrefauthors}%
\unskip\
\newblock
\APACrefYearMonthDay{2018}{}{}.
\newblock
{\BBOQ}\APACrefatitle {Hold Or take Optimal Plan (HOOP): A quadratic
  programming approach to multi-robot trajectory generation} {Hold or take
  optimal plan (hoop): A quadratic programming approach to multi-robot
  trajectory generation}.{\BBCQ}
\newblock
\APACjournalVolNumPages{The International Journal of Robotics
  Research}{}{}{0278364917741532}.
\PrintBackRefs{\CurrentBib}

\bibitem [\protect \citeauthoryear {%
Tobin%
\ \protect \BOthers {.}}{%
Tobin%
\ \protect \BOthers {.}}{%
{\protect \APACyear {2017}}%
}]{%
tobin2017domain}
\APACinsertmetastar {%
tobin2017domain}%
\begin{APACrefauthors}%
Tobin, J.%
, Fong, R.%
, Ray, A.%
, Schneider, J.%
, Zaremba, W.%
\BCBL {}\ \BBA {} Abbeel, P.%
\end{APACrefauthors}%
\unskip\
\newblock
\APACrefYearMonthDay{2017}{}{}.
\newblock
{\BBOQ}\APACrefatitle {Domain randomization for transferring deep neural
  networks from simulation to the real world} {Domain randomization for
  transferring deep neural networks from simulation to the real world}.{\BBCQ}
\newblock
\BIn{} \APACrefbtitle {Intelligent Robots and Systems (IROS), 2017 IEEE/RSJ
  International Conference on} {Intelligent robots and systems (iros), 2017
  ieee/rsj international conference on}\ (\BPGS\ 23--30).
\PrintBackRefs{\CurrentBib}

\bibitem [\protect \citeauthoryear {%
Turpin%
, Michael%
\BCBL {}\ \BBA {} Kumar%
}{%
Turpin%
\ \protect \BOthers {.}}{%
{\protect \APACyear {2014}}%
}]{%
turpin2014capt}
\APACinsertmetastar {%
turpin2014capt}%
\begin{APACrefauthors}%
Turpin, M.%
, Michael, N.%
\BCBL {}\ \BBA {} Kumar, V.%
\end{APACrefauthors}%
\unskip\
\newblock
\APACrefYearMonthDay{2014}{}{}.
\newblock
{\BBOQ}\APACrefatitle {Capt: Concurrent assignment and planning of trajectories
  for multiple robots} {Capt: Concurrent assignment and planning of
  trajectories for multiple robots}.{\BBCQ}
\newblock
\APACjournalVolNumPages{The International Journal of Robotics
  Research}{33}{1}{98--112}.
\PrintBackRefs{\CurrentBib}

\bibitem [\protect \citeauthoryear {%
van~den Berg%
, Guy%
, Lin%
\BCBL {}\ \BBA {} Manocha%
}{%
van~den Berg%
\ \protect \BOthers {.}}{%
{\protect \APACyear {2011}}%
{\protect \APACexlab {{\protect \BCnt {1}}}}}]{%
Berg:ORCA:2011}
\APACinsertmetastar {%
Berg:ORCA:2011}%
\begin{APACrefauthors}%
van~den Berg, J.%
, Guy, S\BPBI J.%
, Lin, M.%
\BCBL {}\ \BBA {} Manocha, D.%
\end{APACrefauthors}%
\unskip\
\newblock
\APACrefYearMonthDay{2011{\protect \BCnt {1}}}{}{}.
\newblock
{\BBOQ}\APACrefatitle {International Symposium on Robotics Research}
  {International symposium on robotics research}.{\BBCQ}
\newblock
\BIn{} C.~Pradalier, R.~Siegwart\BCBL {}\ \BBA {} G.~Hirzinger\ (\BEDS),
  (\BPGS\ 3--19).
\newblock
\APACaddressPublisher{Berlin, Heidelberg}{Springer Berlin Heidelberg}.
\PrintBackRefs{\CurrentBib}

\bibitem [\protect \citeauthoryear {%
van~den Berg%
, Guy%
, Lin%
\BCBL {}\ \BBA {} Manocha%
}{%
van~den Berg%
\ \protect \BOthers {.}}{%
{\protect \APACyear {2011}}%
{\protect \APACexlab {{\protect \BCnt {2}}}}}]{%
van2011reciprocal}
\APACinsertmetastar {%
van2011reciprocal}%
\begin{APACrefauthors}%
van~den Berg, J.%
, Guy, S\BPBI J.%
, Lin, M.%
\BCBL {}\ \BBA {} Manocha, D.%
\end{APACrefauthors}%
\unskip\
\newblock
\APACrefYearMonthDay{2011{\protect \BCnt {2}}}{}{}.
\newblock
{\BBOQ}\APACrefatitle {Reciprocal n-body collision avoidance} {Reciprocal
  n-body collision avoidance}.{\BBCQ}
\newblock
\BIn{} \APACrefbtitle {Robotics research} {Robotics research}\ (\BPGS\ 3--19).
\newblock
\APACaddressPublisher{}{Springer}.
\PrintBackRefs{\CurrentBib}

\bibitem [\protect \citeauthoryear {%
van~den Berg%
, Lin%
\BCBL {}\ \BBA {} Manocha%
}{%
van~den Berg%
\ \protect \BOthers {.}}{%
{\protect \APACyear {2008}}%
}]{%
van2008reciprocal}
\APACinsertmetastar {%
van2008reciprocal}%
\begin{APACrefauthors}%
van~den Berg, J.%
, Lin, M.%
\BCBL {}\ \BBA {} Manocha, D.%
\end{APACrefauthors}%
\unskip\
\newblock
\APACrefYearMonthDay{2008}{}{}.
\newblock
{\BBOQ}\APACrefatitle {Reciprocal velocity obstacles for real-time multi-agent
  navigation} {Reciprocal velocity obstacles for real-time multi-agent
  navigation}.{\BBCQ}
\newblock
\BIn{} \APACrefbtitle {International Conference on Robotics and Automation}
  {International conference on robotics and automation}\ (\BPGS\ 1928--1935).
\PrintBackRefs{\CurrentBib}

\bibitem [\protect \citeauthoryear {%
Yi%
, Li%
\BCBL {}\ \BBA {} Wang%
}{%
Yi%
\ \protect \BOthers {.}}{%
{\protect \APACyear {2016}}%
}]{%
Yi:2016:ECCV}
\APACinsertmetastar {%
Yi:2016:ECCV}%
\begin{APACrefauthors}%
Yi, S.%
, Li, H.%
\BCBL {}\ \BBA {} Wang, X.%
\end{APACrefauthors}%
\unskip\
\newblock
\APACrefYearMonthDay{2016}{}{}.
\newblock
{\BBOQ}\APACrefatitle {Pedestrian Behavior Understanding and Prediction with
  Deep Neural Networks} {Pedestrian behavior understanding and prediction with
  deep neural networks}.{\BBCQ}
\newblock
\BIn{} \APACrefbtitle {European Conference on Computer Vision} {European
  conference on computer vision}\ (\BPGS\ 263--279).
\PrintBackRefs{\CurrentBib}

\bibitem [\protect \citeauthoryear {%
Yu%
\ \BBA {} LaValle%
}{%
Yu%
\ \BBA {} LaValle%
}{%
{\protect \APACyear {2016}}%
}]{%
yu2016optimal}
\APACinsertmetastar {%
yu2016optimal}%
\begin{APACrefauthors}%
Yu, J.%
\BCBT {}\ \BBA {} LaValle, S\BPBI M.%
\end{APACrefauthors}%
\unskip\
\newblock
\APACrefYearMonthDay{2016}{}{}.
\newblock
{\BBOQ}\APACrefatitle {Optimal multirobot path planning on graphs: Complete
  algorithms and effective heuristics} {Optimal multirobot path planning on
  graphs: Complete algorithms and effective heuristics}.{\BBCQ}
\newblock
\APACjournalVolNumPages{IEEE Transactions on Robotics}{32}{5}{1163--1177}.
\PrintBackRefs{\CurrentBib}

\bibitem [\protect \citeauthoryear {%
J.~Zhang%
, Springenberg%
, Boedecker%
\BCBL {}\ \BBA {} Burgard%
}{%
J.~Zhang%
\ \protect \BOthers {.}}{%
{\protect \APACyear {2016}}%
}]{%
zhang2016deep}
\APACinsertmetastar {%
zhang2016deep}%
\begin{APACrefauthors}%
Zhang, J.%
, Springenberg, J\BPBI T.%
, Boedecker, J.%
\BCBL {}\ \BBA {} Burgard, W.%
\end{APACrefauthors}%
\unskip\
\newblock
\APACrefYearMonthDay{2016}{}{}.
\newblock
{\BBOQ}\APACrefatitle {Deep reinforcement learning with successor features for
  navigation across similar environments} {Deep reinforcement learning with
  successor features for navigation across similar environments}.{\BBCQ}
\newblock
\APACjournalVolNumPages{arXiv:1612.05533}{}{}{}.
\PrintBackRefs{\CurrentBib}

\bibitem [\protect \citeauthoryear {%
T.~Zhang%
, Kahn%
, Levine%
\BCBL {}\ \BBA {} Abbeel%
}{%
T.~Zhang%
\ \protect \BOthers {.}}{%
{\protect \APACyear {2016}}%
}]{%
zhang:2016:LDC}
\APACinsertmetastar {%
zhang:2016:LDC}%
\begin{APACrefauthors}%
Zhang, T.%
, Kahn, G.%
, Levine, S.%
\BCBL {}\ \BBA {} Abbeel, P.%
\end{APACrefauthors}%
\unskip\
\newblock
\APACrefYearMonthDay{2016}{}{}.
\newblock
{\BBOQ}\APACrefatitle {Learning deep control policies for autonomous aerial
  vehicles with MPC-guided policy search} {Learning deep control policies for
  autonomous aerial vehicles with mpc-guided policy search}.{\BBCQ}
\newblock
\BIn{} \APACrefbtitle {IEEE International Conference on Robotics and
  Automation} {Ieee international conference on robotics and automation}\
  (\BPG~528-535).
\PrintBackRefs{\CurrentBib}

\bibitem [\protect \citeauthoryear {%
Zhou%
, Wang%
, Bandyopadhyay%
\BCBL {}\ \BBA {} Schwager%
}{%
Zhou%
\ \protect \BOthers {.}}{%
{\protect \APACyear {2017}}%
}]{%
zhou2017fast}
\APACinsertmetastar {%
zhou2017fast}%
\begin{APACrefauthors}%
Zhou, D.%
, Wang, Z.%
, Bandyopadhyay, S.%
\BCBL {}\ \BBA {} Schwager, M.%
\end{APACrefauthors}%
\unskip\
\newblock
\APACrefYearMonthDay{2017}{}{}.
\newblock
{\BBOQ}\APACrefatitle {Fast, on-line collision avoidance for dynamic vehicles
  using buffered voronoi cells} {Fast, on-line collision avoidance for dynamic
  vehicles using buffered voronoi cells}.{\BBCQ}
\newblock
\APACjournalVolNumPages{IEEE Robotics and Automation
  Letters}{2}{2}{1047--1054}.
\PrintBackRefs{\CurrentBib}

\bibitem [\protect \citeauthoryear {%
Zhu%
\ \protect \BOthers {.}}{%
Zhu%
\ \protect \BOthers {.}}{%
{\protect \APACyear {2017}}%
}]{%
zhu2017target}
\APACinsertmetastar {%
zhu2017target}%
\begin{APACrefauthors}%
Zhu, Y.%
, Mottaghi, R.%
, Kolve, E.%
, Lim, J\BPBI J.%
, Gupta, A.%
, Fei-Fei, L.%
\BCBL {}\ \BBA {} Farhadi, A.%
\end{APACrefauthors}%
\unskip\
\newblock
\APACrefYearMonthDay{2017}{}{}.
\newblock
{\BBOQ}\APACrefatitle {Target-driven visual navigation in indoor scenes using
  deep reinforcement learning} {Target-driven visual navigation in indoor
  scenes using deep reinforcement learning}.{\BBCQ}
\newblock
\BIn{} \APACrefbtitle {International Conference on Robotics and Automation}
  {International conference on robotics and automation}\ (\BPGS\ 3357--3364).
\PrintBackRefs{\CurrentBib}

\end{thebibliography}
}

\end{document}